\documentclass[10pt,twocolumn,letterpaper]{article}

\usepackage{cvpr}              %

\usepackage{kotex}

\usepackage{cuted} %

\usepackage{currfile} %

\usepackage{caption} %
\newcommand{\MV}{\mathbf{M}_V} %
\newcommand{\ML}{\mathbf{M}_L} %
\newcommand{\NL}{N_L} %

\newcommand{\W}{\mathbf{W}}
\newcommand{\G}{\mathbf{G}}
\newcommand{\Sm}{\mathbf{S}}
\newcommand{\I}{\mathbf{I}}
\newcommand{\Lh}{\hat{\mathbf{L}}}
\newcommand{\B}{\mathbf{B}}
\newcommand{\x}{\mathbf{x}}

\newcommand{\p}{\mathbf{p}}

\newcommand{\Loss}{\mathcal{L}}

\definecolor{cvprblue}{rgb}{0.21,0.49,0.74}
\usepackage[pagebackref,breaklinks,colorlinks,allcolors=cvprblue]{hyperref}

\def\confName{CVPR}
\def\confYear{2027}

\title{Integrating Learned Priors through Prior-Space Reparameterization \\for Physically Based Inverse Rendering}

\author{
Hun Hwangbo$^1$ \quad Minjun Park$^1$ \quad
Yongsung Kim$^1$ \quad
Wooseok Song$^2$ \\
Jungmin Ko$^2$ \quad
Junghyun Hwang$^3$ \quad
Sungroh Yoon$^{1,2,3}$
\\[0.5em]
$^1$IPAI, $^2$ECE, $^3$FAM, \,
Seoul National University\\
 \url{https://github.com/hunhwb/PRep-IR}
}

\begin{document}
\maketitle
\begin{strip}
\centering\setlength{\tabcolsep}{.6pt}\renewcommand{\arraystretch}{0}
\begin{tabular}{c cccc}
&\parbox[b]{0.238\textwidth}{\centering\fontsize{9}{10}\selectfont MaterialFusion (125.3 min)}&\parbox[b]{0.238\textwidth}{\centering\fontsize{9}{10}\selectfont DiffReg-PBIR (25.0 min)}&\parbox[b]{0.238\textwidth}{\centering\fontsize{9}{10}\selectfont Ours (6.2 min)}&\parbox[b]{0.238\textwidth}{\centering\fontsize{9}{10}\selectfont Ground truth}\\[3pt]
\rotatebox{90}{\makebox[0.094\textwidth][c]{\fontsize{9}{10}\selectfont Novel view}}&\begin{minipage}[t]{0.238\textwidth}\includegraphics[width=\linewidth]{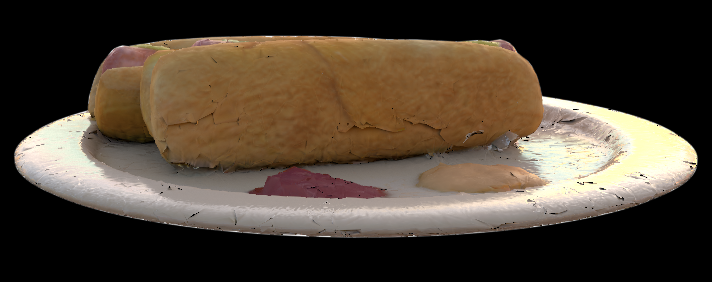}\llap{\raisebox{2pt}{\textcolor{white}{\fontsize{9}{10}\selectfont 30.09}}\hspace{2.5pt}}\end{minipage}&\begin{minipage}[t]{0.238\textwidth}\includegraphics[width=\linewidth]{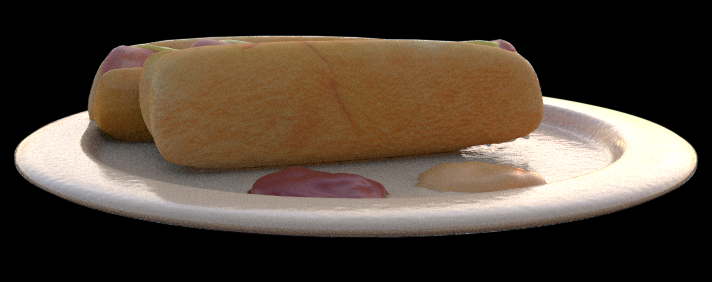}\llap{\raisebox{2pt}{\textcolor{white}{\fontsize{9}{10}\selectfont 33.00}}\hspace{2.5pt}}\end{minipage}&\begin{minipage}[t]{0.238\textwidth}\includegraphics[width=\linewidth]{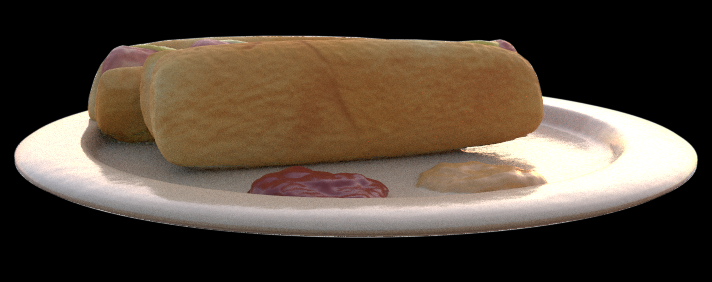}\llap{\raisebox{2pt}{\textcolor{white}{\fontsize{9}{10}\selectfont \textbf{35.62}}}\hspace{2.5pt}}\end{minipage}&\begin{minipage}[t]{0.238\textwidth}\includegraphics[width=\linewidth]{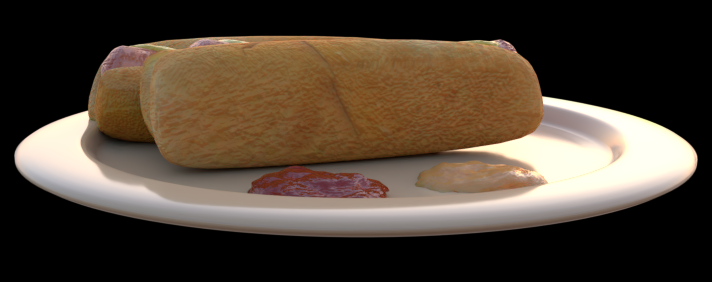}\end{minipage}\\[1.2pt]
\rotatebox{90}{\makebox[0.094\textwidth][c]{\fontsize{9}{10}\selectfont Albedo}}&\begin{minipage}[t]{0.238\textwidth}\includegraphics[width=\linewidth]{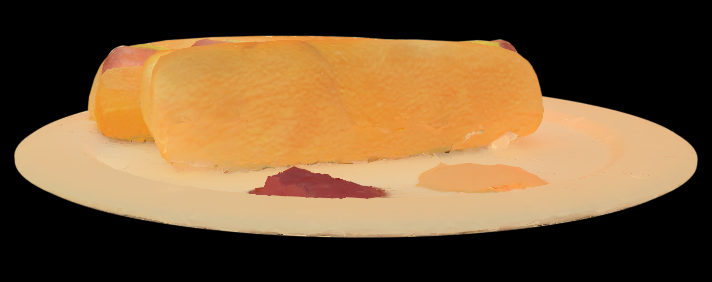}\llap{\raisebox{2pt}{\textcolor{white}{\fontsize{9}{10}\selectfont 30.89}}\hspace{2.5pt}}\end{minipage}&\begin{minipage}[t]{0.238\textwidth}\includegraphics[width=\linewidth]{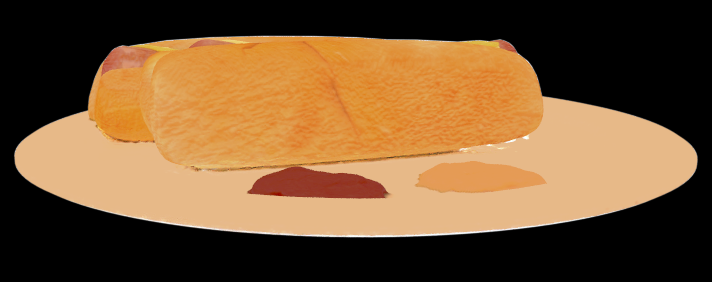}\llap{\raisebox{2pt}{\textcolor{white}{\fontsize{9}{10}\selectfont 33.49}}\hspace{2.5pt}}\end{minipage}&\begin{minipage}[t]{0.238\textwidth}\includegraphics[width=\linewidth]{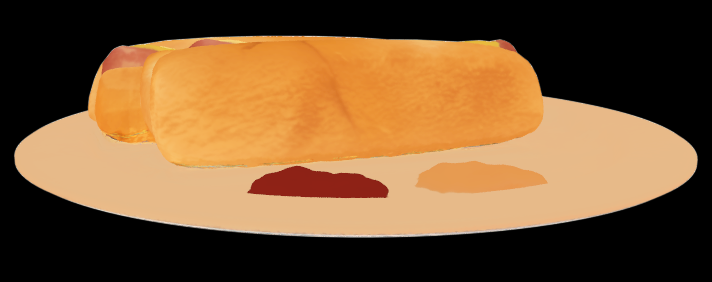}\llap{\raisebox{2pt}{\textcolor{white}{\fontsize{9}{10}\selectfont \textbf{37.80}}}\hspace{2.5pt}}\end{minipage}&\begin{minipage}[t]{0.238\textwidth}\includegraphics[width=\linewidth]{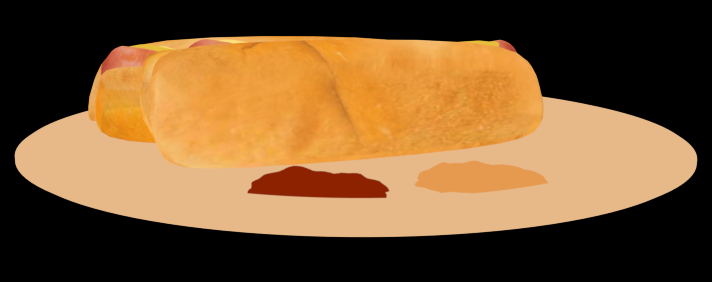}\end{minipage}\\[1.2pt]
\rotatebox{90}{\makebox[0.094\textwidth][c]{\fontsize{9}{10}\selectfont Relighting}}&\begin{minipage}[t]{0.238\textwidth}\includegraphics[width=\linewidth]{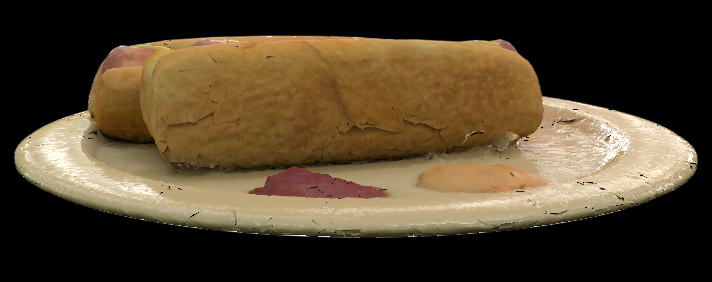}\llap{\raisebox{2pt}{\textcolor{white}{\fontsize{9}{10}\selectfont 28.73}}\hspace{2.5pt}}\end{minipage}&\begin{minipage}[t]{0.238\textwidth}\includegraphics[width=\linewidth]{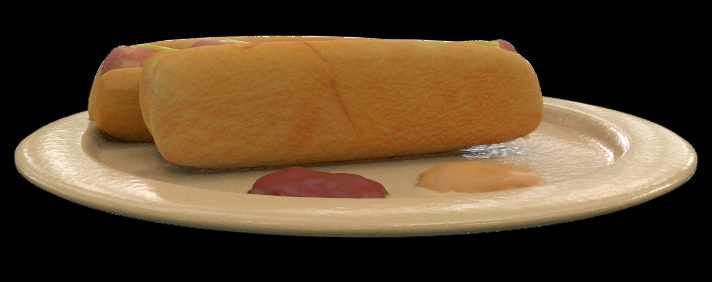}\llap{\raisebox{2pt}{\textcolor{white}{\fontsize{9}{10}\selectfont 31.90}}\hspace{2.5pt}}\end{minipage}&\begin{minipage}[t]{0.238\textwidth}\includegraphics[width=\linewidth]{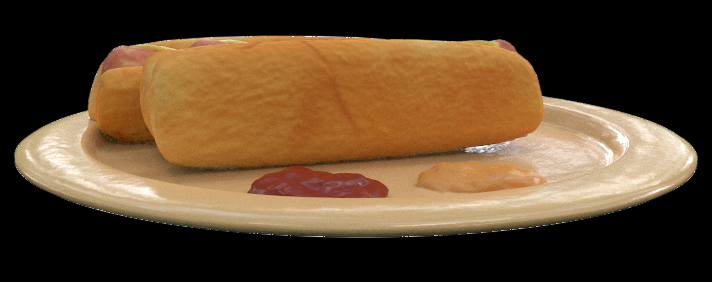}\llap{\raisebox{2pt}{\textcolor{white}{\fontsize{9}{10}\selectfont \textbf{32.43}}}\hspace{2.5pt}}\end{minipage}&\begin{minipage}[t]{0.238\textwidth}\includegraphics[width=\linewidth]{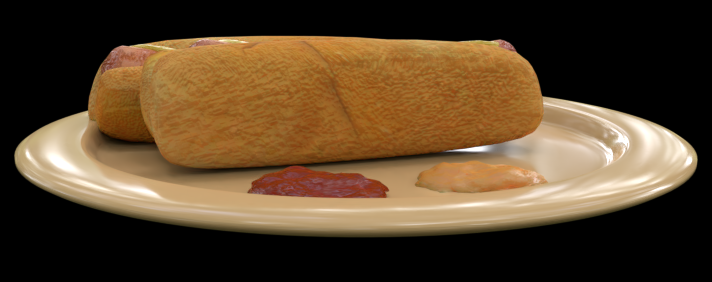}\end{minipage}\\[1.2pt]
\end{tabular}
\vspace{2pt}\captionsetup{font=small,labelfont=bf,labelsep=period,skip=3pt}
\captionsetup{hypcap=false}
\captionof{figure}{\textbf{Learned priors as a material reparameterization.} Novel view, albedo and relighting of TensoIR-Synthetic Hotdog. Ours and DiffReg-PBIR use DiffusionRenderer as the prior. MaterialFusion uses its native learned-prior formulation. Numbers are full-image PSNR (dB).
Times denote the optimization time of each run.}
\captionsetup{hypcap=true}
\label{fig:teaser}
\end{strip}

\begin{abstract}
Physically based inverse rendering is an ambiguous problem, allowing independently optimized material variables to absorb appearance variations caused by illumination. 
Learned material predictions provide useful cues about which surface points have similar materials.
Existing approaches often incorporate these cues through auxiliary regularization losses.
Instead, we propose a novel prior-guided reparameterization of the material optimization space.
We fuse multi-view material predictions onto a mesh and construct a permutohedral lattice in the resulting prior feature space.
Surface points with similar predictions share lattice variables, restricting the material field to a prior-defined subspace that is optimized jointly with geometry and illumination using a purely photometric objective.
Experiments on three synthetic benchmarks demonstrate improved material decomposition and relighting, with results on par with the strongest baseline on Stanford-ORB.
As a complete system, our method improves relighting PSNR by 1.4--1.5\,dB over the strongest evaluated mesh-based baseline on Synthetic4Relight and TensoIR-Synthetic.
Analysis of the optimization shows that shared variables pool per-vertex photometric gradients into updates that are more consistent across sampled views.
In a controlled comparison within the same pipeline, the restricted subspace, despite its reduced degrees of freedom, yields more accurate material decomposition than both unrestricted per-vertex optimization and a regularization loss built from the same prior.
\end{abstract}

\section{Introduction}
\label{sec:intro}

Physically based inverse rendering recovers the geometry, materials and illumination of a scene from photographs, enabling relighting, editing and novel-view rendering.
The task is inherently ambiguous in that different combinations of materials and illumination can explain the same observations.
When materials are optimized independently at each surface point, this ambiguity can cause shadows and highlights to be absorbed into the recovered reflectance.

Learned material predictors provide useful cues for resolving this ambiguity.
Although their predictions may contain errors, they often reliably reveal which surfaces have similar materials~\cite{zeng2024rgb,kocsis2024iid,liang2025diffusionrenderer,wu2026mvinverse}.
Recent inverse rendering methods incorporate these cues through auxiliary losses that anchor materials to predicted values~\cite{zhu2026ptir}, distill a material diffusion model~\cite{litman2025materialfusion}, or encourage points with similar predictions to have similar materials~\cite{diffusion_material_regularization_eccv}.
Such regularization losses can help guide the decomposition, but they also alter the objective and shift the resulting solution.
This motivates an alternative: using learned prior predictions to reduce the degrees of freedom and restrict the optimization space. 
Specifically, we let surface points with similar predicted materials share material variables, leaving the optimization objective unchanged.

In this work, we propose a prior-guided material reparameterization for physically based inverse rendering.
We fuse multi-view material predictions of learned models onto a mesh and construct a permutohedral lattice~\cite{adams2010fast} in the resulting prior feature space.
Surface materials are represented by shared lattice variables, coupling points with similar predictions even when they are spatially distant.
We optimize the lattice values jointly with geometry and illumination using a purely photometric objective.

Experiments on Synthetic4Relight~\cite{zhang2022modeling}, TensoIR-Synthetic~\cite{tensoir}, DTC-Synthetic~\cite{diffusion_material_regularization_eccv} and Stanford-ORB~\cite{orb} demonstrate the effectiveness of this approach with DiffusionRenderer priors.
Our method improves relighting PSNR by 1.4--1.5\,dB over the strongest evaluated mesh-based baseline on Synthetic4Relight and TensoIR-Synthetic, and achieves competitive results on real captures.

We further analyze how restricting the optimization space helps.
Independent per-vertex updates are dominated by variation across sampled views and path-tracing noise.
Sharing lattice variables pools the photometric gradients of vertices with similar prior predictions, making the updates far more consistent across views.
Furthermore, our prior-space reparameterization uses two orders of magnitude fewer variables than the mesh has vertices, while it sacrifices little representational capacity.
Optimization within this subspace recovers more accurate albedo than both unrestricted per-vertex optimization and a regularization loss built from the same prior.

Our contributions are:
\begin{itemize}
\item A material reparameterization that incorporates learned prior predictions through a prior-space lattice representation without changing the optimization objective.
\item State-of-the-art relighting among mesh-based inverse rendering methods on synthetic benchmarks, with effective transfer to real captures.
\item An analysis showing that shared prior-space variables pool photometric gradients into more consistent updates, yielding more accurate materials than unrestricted per-vertex optimization.
\end{itemize}

\section{Related Work}
\label{sec:related_work}

\subsection{Inverse Rendering}
\label{sec:inverse_rendering}
Inverse rendering separates the intrinsic properties of an object from its illumination, 
with common approaches falling into two families.
Optimization-based methods treat materials and lighting as free variables and fit them to the observations, 
on meshes~\cite{Munkberg_2022_CVPR,hasselgren2022nvdiffrecmc,neuralpbir2023,diffusion_material_regularization_eccv}, 
on Gaussian primitives~\cite{irgs,svgir,gsir,han2026radiogs,fan2025rng,zhu2026ptir}, or on hybrids of the two~\cite{geosplatting,lin2026rotatedlightsconsistentefficient}.
They explicitly fit the input images, but the many material degrees of freedom can allow shadows and highlights to be absorbed into the material.
Learning-based methods train a network on large, mostly synthetic datasets to predict materials directly, 
usually either with image~\cite{zeng2024rgb,ke2025marigold,pavlovic2026marigold,neural_lightrig,kocsis2024iid} or video diffusion models~\cite{liang2025diffusionrenderer,unirelight,munkberg2025videomat,zeng2026rgbxnextrealisticgenerativerendering,fang2025vrgbxvideoeditingaccurate}.
There are also approaches that use multi-view transformers such as MVInverse~\cite{wu2026mvinverse}, MVID~\cite{du2026mvidfeedforwardmultiviewintrinsic}, LIRM~\cite{li2025lirm} or LSRM~\cite{lsrm}.
They are fast and capture the overall structure, but are not guaranteed to be consistent across views or to faithfully reproduce the observed images.

\subsection{Integrating Priors in Inverse Rendering}
Hand-crafted priors, such as total variation, smoothness of the material maps,
 or a white-light assumption, have long been added to optimization-based inverse rendering pipelines as regularization losses~\cite{neuralpbir2023,irgs,han2026radiogs,gsir,Munkberg_2022_CVPR,hasselgren2022nvdiffrecmc}.
Recently, learned priors have increasingly been injected the same way.
DiffReg-PBIR~\cite{diffusion_material_regularization_eccv} uses only the structure of the G-buffers predicted by DiffusionRenderer~\cite{liang2025diffusionrenderer}:
a joint bilateral regularizer on a permutohedral lattice encourages points with similar predicted materials to have similar optimized materials.
PTIR-GS~\cite{zhu2026ptir} regularizes the optimized material properties by anchoring the optimization to predictions from RGB$\leftrightarrow$X~\cite{zeng2024rgb}.
MaterialFusion~\cite{litman2025materialfusion} distills a material diffusion model into the optimized albedo and roughness through a score-distillation loss. 
All of these methods add an auxiliary term to the photometric objective.
MaterialClusterGS~\cite{zhang2026materialclustergs} instead restricts the degrees of freedom,
exploiting the fact that a real scene contains few distinct materials, and softly assigns primitives to a fixed-size material palette.
Like DiffReg-PBIR, we use similarities in predicted materials.
In our method, these similarities define shared variables rather than being incorporated as an auxiliary loss.

\subsection{Preconditioning and Reparameterization}
\label{sec:rw_precond}
Poor conditioning of the loss landscape and noisy gradient estimates can slow the convergence of gradient-based optimization.
Adaptive coordinate scaling, as in Adam~\cite{adam}, and spatial smoothing, as with Sobolev gradients~\cite{neuberger1997sobolev}, modify the updates to improve optimization.
A linear change of variables can also couple updates without adding a loss~\cite{neuberger1997sobolev,nicolet2021large}.
When the map spans the original variable space, it preserves the set of attainable solutions; a lower-rank map additionally constrains the representation.

In inverse rendering, Large Steps~\cite{nicolet2021large} uses a mesh-based linear reparameterization to couple geometry updates and permit larger steps.
Chang~\etal~\cite{chang2024spatiotemporal} further improve this smoothing through a spatiotemporal bilateral filter that combines cross-bilateral spatial filtering with Adam's temporal averaging to suppress noise while preserving sharp features.

In computational photography, bilateral-space methods lift pixels into a grid indexed by position and color,
operate on its few vertices, and slice the result back to the pixels.
Grouping pixels by position and color supports edge-aware processing, while the compact grid makes filtering and global optimization efficient~\cite{chen2007real,adams2010fast,krahenbuhl2011efficient,barron2016fast}.
For example, Barron~\etal~\cite{barron2015fast} optimize stereo for synthetic defocus using this idea: 
disparity maps are smooth within regions of similar color and sharp across their edges, 
and the optimization can run on a smaller subspace, with far fewer variables with bilateral grids than with pixels.
Our method follows the same principle. We use learned material-prior features to define the optimization space, representing surface materials through shared lattice variables in this space.

\section{Preliminaries}
\label{sec:preliminaries}

\subsection{Linear Reparameterization}
\label{sec:reparam}
Let $\x\in\mathbb R^{n}$ be the variables of an objective $\Loss(\x)$, \eg, per-vertex materials.
Instead of $\x$, one can optimize latent variables $\mathbf w\in\mathbb R^{k}$ through a fixed linear map,
\begin{equation}
  \x=\x_0+\mathbf P\,\mathbf w ,
  \label{eq:reparam}
\end{equation}
with $\mathbf P\in\mathbb R^{n\times k}$ and a constant $\x_0$.
The chain rule gives $\partial\Loss/\partial\mathbf w=\mathbf P^{\top}\partial\Loss/\partial\x$, so a gradient-descent step of size $\eta$
on $\mathbf w$ moves $\x$ by
\begin{equation}
  \Delta\x=-\eta\,\mathbf P\mathbf P^{\top}\frac{\partial\Loss}{\partial\x}.
  \label{eq:reparam_update}
\end{equation}
The map fixes both the attainable values of $\x$ (the affine subspace $\x_0+\operatorname{range}(\mathbf P)$) and how each step moves them from the gradients on $\x$.
A full-row-rank map preserves the attainable solutions, whereas a lower-rank map restricts optimization to this affine subspace.
Thus, a reparameterization can constrain the solution space and reshape gradient steps without adding a loss term.

\subsection{The Permutohedral Lattice}
\label{sec:permutohedral_lattice}
The permutohedral lattice~\cite{adams2010fast} is a data structure designed for fast high-dimensional Gaussian filtering of images.
It tessellates feature space $\mathbb{R}^d$ with simplices of $d{+}1$ points, instead of the $2^d$ corners of a hypercubic bilateral grid.
Each input sample (\eg, a pixel in image filtering) with feature $\p\in\mathbb{R}^d$ is assigned to a simplex and represented by barycentric weights on its $d{+}1$ points, which sum to one.
Collecting these weights for all $n$ samples gives a sparse matrix $\Sm\in\mathbb{R}^{\NL\times n}$ over the $\NL$ occupied lattice points.

\emph{Splatting} maps a per-sample signal $\mathbf{g}\in\mathbb{R}^n$
onto the lattice via $\Sm\mathbf{g}$, accumulating barycentrically weighted sample values at each lattice point.
\emph{Slicing} maps a lattice signal $\mathbf{c}\in\mathbb{R}^{\NL}$
back to the samples via $\Sm^\top\mathbf{c}$ using barycentric interpolation.

Together, these operations allow per-sample signals to be efficiently processed in the lattice domain and interpolated back to the original points.
Samples with nearby features can share lattice points, allowing information to be pooled and redistributed according to feature similarity.
We use this representation to aggregate material gradients and compute coupled updates in the lattice domain.

\section{Method}
\label{sec:method}
\begin{figure*}[t]
\centering
\includegraphics[width=\textwidth]{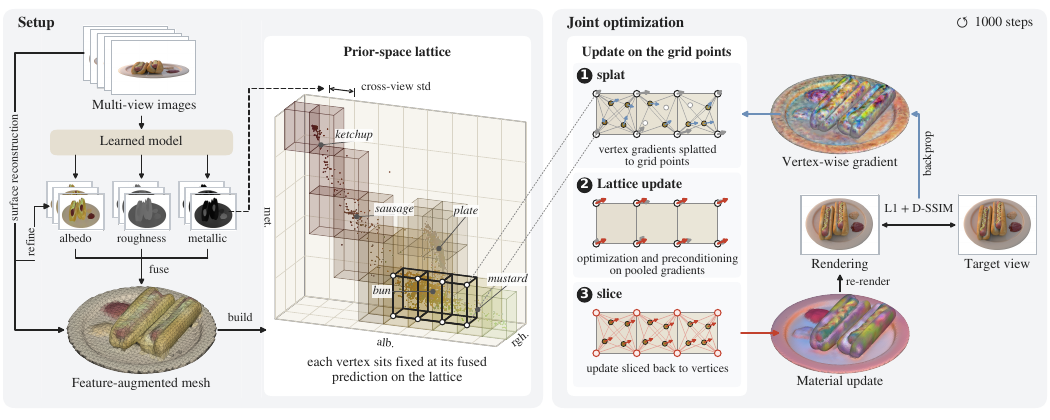}
\vspace{-18pt}
\caption{\textbf{Pipeline overview and prior-space material updates.}
Multi-view images supervise surface reconstruction and provide material predictions.
Input RGB refines predicted albedo before fusion onto the initial mesh.
The fused prior defines a fixed permutohedral lattice over albedo RGB, roughness and metallic.
Per-vertex photometric gradients are splatted onto the lattice, and the resulting updates are sliced back onto the mesh vertices.
For visualization, the lattice is drawn as a regular 3-D grid. In the 5-D permutohedral lattice, each vertex uses the six corners of its enclosing simplex.
}
\label{fig:schematic}
\vspace{-6pt}
\end{figure*}

Our method recovers mesh geometry, materials, and illumination from multi-view images captured under fixed lighting.
Its central idea is to use learned material predictions to define a restricted space of material fields for optimization, rather than assigning an independent material variable to every mesh vertex.
We first describe this material reparameterization (\cref{sec:materials}) and its update process (\cref{sec:update}), then the full pipeline (\cref{sec:setup}).

\subsection{Problem Formulation}
\label{sec:model_objective}
We represent a scene by mesh vertex positions $\mathbf V$, per-vertex material logits $\MV$, and an environment map $\mathbf E$.
The material logits $\MV \in\mathbb R^{N\times 5}$ encode albedo RGB, roughness and metallic, which are mapped through a sigmoid to bounded values for rendering.
We jointly optimize geometry, materials, and illumination using a photometric loss $\Loss$ that compares rendered and observed images.

We aim to solve the following optimization problem:
\begin{equation}
  \min_{\mathbf V,\mathbf M_V,\mathbf E}
  \Loss(\mathbf V,\mathbf M_V,\mathbf E).
  \label{eq:optimization_objective}
\end{equation}

With independent per-vertex materials, vertices predicted to have identical or similar features are optimized freely.
Our goal is to use the prior's similarity structure to tie such degrees of freedom together while retaining the variation needed to faithfully explain the images.
We do so by changing the material representation, without adding a prior-induced regularization loss to the objective of~\cref{eq:optimization_objective}.

\subsection{Prior-Guided Material Reparameterization}
\label{sec:materials}
Assume that each mesh vertex has a fixed prior feature $\mathbf p_i\in[0,1]^5$, containing predicted albedo RGB, roughness, and metallic.
We use these features as coordinates in a permutohedral lattice (\cref{sec:permutohedral_lattice}).
Each mesh vertex is associated with the lattice points of its enclosing simplex through barycentric weights.
Collecting these weights gives the sparse splat matrix $\Sm\in\mathbb R^{\NL\times N}$, where $N$ is the number of mesh vertices and $\NL$ is the number of occupied lattice points, with $\NL \ll N$ in our setting.
Thus, vertices with similar prior features are represented by similar lattice variables (middle of \cref{fig:schematic}).
Because the features contain no position, this sharing follows predicted material similarity rather than proximity on the mesh.

Instead of directly optimizing independent per-vertex logits $\MV$ to solve~\cref{eq:optimization_objective}, we do it in the lattice domain.
To this end, we track material-logit offsets $\ML\in\mathbb R^{\NL\times 5}$ at the lattice points.
Then we can interpolate the lattice point offsets onto the mesh by slicing:
\begin{equation}
  \MV=\mathbf{M}_0+\Sm^\top\ML,
  \label{eq:change_variable}
\end{equation}
where $\mathbf{M}_0$ is the logit initialization value.
This is the linear reparameterization of \cref{sec:reparam} with $\mathbf P=\Sm^\top$.

For the material variables, this reparameterization turns the original optimization problem into:
\begin{equation}
  \min_{\mathbf M_V\in\mathcal M}
  \Loss(\mathbf V,\mathbf M_V,\mathbf E)
  =
  \min_{\mathbf M_L}
  \Loss\!\left(\mathbf V,\mathbf M_0+\Sm^\top\mathbf M_L,\mathbf E\right),
  \label{eq:restricted_optimization}
\end{equation}
where $\mathcal M=\mathbf M_0+\operatorname{range}(\Sm^\top)$ is the affine subspace of material logits defined by the lattice representation.
A change to one lattice variable affects every mesh vertex that uses it, so these vertices can no longer vary independently.
Since vertices with similar prior features receive similar offsets,
the prior determines which mesh vertices share optimization variables.
As long as the prior predictions capture the overall structure of the materials, \ie, which regions are alike and which differ, sharing can eliminate unnecessary variation while preserving the representational capacity needed to explain the observed images.

\subsection{Optimization through Lattice Variables}
\label{sec:update}
We next describe how material updates are computed in the joint optimization stage using this material reparameterization (right of~\cref{fig:schematic}).
\paragraph{Splatting.}
Differentiable rendering~\cite{jakob2022dr,zhang2023projective} gives per-vertex gradients $\partial\Loss/\partial\MV$ at every optimization step.
By the chain rule, the gradients on the lattice points can be obtained using the splat matrix:
\begin{equation}
  \frac{\partial\Loss}{\partial\ML}=\Sm\frac{\partial\Loss}{\partial\MV}.
  \label{eq:splat}
\end{equation}
Each lattice point thus receives a barycentrically weighted sum of gradients from the mesh vertices that use it.
Photometric evidence from vertices with similar prior predictions is pooled into these points, even when the vertices are distant on the surface.

\paragraph{Update on the lattice.}
We then apply the Adam~\cite{adam} optimizer and Large Steps~\cite{nicolet2021large} preconditioning on the pooled lattice point gradients to obtain the final lattice variable update for the step. 
Adam works as a temporal smoothing filter, and Large Steps helps smooth updates between neighboring lattice points. 

\paragraph{Slicing.}
The updated lattice offsets are sliced back to the mesh to recompute the per-vertex material logits using \cref{eq:change_variable}.
These logits are then decoded into material values for rendering in the next optimization step.

Splatting and slicing remain sparse operations, with only six weights per mesh vertex for our five-dimensional features.
Implementation details on the lattice update are provided in \cref{sec:material_ls}.

\subsection{Scene Initialization and Joint Optimization}
\label{sec:setup}
\label{sec:lattice}
\paragraph{Mesh and prior prediction.}
Before joint optimization, we reconstruct the initial mesh using the surface initialization of DiffReg-PBIR~\cite{diffusion_material_regularization_eccv}.
For the prior model that predicts albedo, roughness, and metallic for each training view,
we use DiffusionRenderer~\cite{liang2025diffusionrenderer}, though any per-pixel material predictor could be used. 
Additional results with MVInverse~\cite{wu2026mvinverse} priors are provided in \cref{sec:alternative_priors}.

\paragraph{Refinement and fusion of predicted materials.}
\label{sec:prior_refinement}
We fuse per-view albedo, roughness and metallic predictions onto the initial mesh to obtain the fixed per-vertex prior features used in \cref{sec:materials}.
Since blurred predictions can incorrectly couple distinct materials through shared lattice variables, we refine albedo before fusion.
The refinement uses a joint trilateral filter guided by input RGB, predicted albedo and spatial proximity~\cite{kopf2007jbu,matsuo2016depthmap}.
Details are provided in \cref{sec:pipeline_setup}.

\paragraph{Metallic.}
Under a photometric objective, metallic is ambiguous with base color, and when optimized it tends to settle at intermediate values instead of the near-binary values expected of real materials.
Thus, we fix metallic to zero in our main experiments; however, we also report results for learned metallic and explore a dielectric--metallic dual hypothesis that returns binary metallic in~\cref{sec:metallic}.

\paragraph{Lattice spacing.}
Lattice spacing controls how broadly vertices share variables in prior feature space.
We set it from the median cross-view standard deviation of prior features at vertices observed in multiple training views.

\paragraph{Joint scene optimization.}
\label{sec:geometry_illumination}
We optimize lattice materials jointly with mesh geometry and illumination using gradients from the same renderings for 1000 steps.
The optimization loss $\Loss$ combines L1 and D-SSIM~\cite{wang2004image} on the object and L1 on the observed background~\cite{neuralpbir2023,diffusion_material_regularization_eccv}.
Geometry optimization uses mesh-based Large Steps~\cite{nicolet2021large}, while illumination follows Neural-PBIR~\cite{neuralpbir2023}, starting with spherical Gaussians and then switching to an environment texture.
The full optimization settings are given in \cref{sec:optimization_setup}.

\begin{table*}[t]
\centering
\small
\setlength{\tabcolsep}{5.0pt}
\caption{\textbf{Synthetic benchmarks.}
Means over four scenes each.
Opt.\ denotes optimization time. Total additionally includes setup, prior inference and mesh initialization.
+\,learned metallic additionally optimizes metallic instead of fixing it to zero.
}
\label{tab:main}
\begin{tabular}{l cccccccc cc}
\toprule
& \multicolumn{2}{c}{Albedo} & \multicolumn{2}{c}{Relighting} & \multicolumn{2}{c}{NVS} & Roughness & Normal & \multicolumn{2}{c}{Time (min)} \\
\cmidrule(lr){2-3}\cmidrule(lr){4-5}\cmidrule(lr){6-7}\cmidrule(lr){8-8}\cmidrule(lr){9-9}\cmidrule(lr){10-11}
Method & PSNR$\uparrow$ & SSIM$\uparrow$ & PSNR$\uparrow$ & SSIM$\uparrow$ & PSNR$\uparrow$ & SSIM$\uparrow$ & MAE$\downarrow$ & MAE ($^\circ$)$\downarrow$ & Opt.$\downarrow$ & Total$\downarrow$ \\
\midrule
\multicolumn{11}{c}{\textbf{Synthetic4Relight}~\cite{zhang2022modeling} } \\
\midrule
Neural-PBIR~\cite{neuralpbir2023} & 29.54 & .933 & 30.76 & .958 & 32.06 & .940 & .133 & -- & 7.2 & \textbf{18} \\
MaterialFusion~\cite{litman2025materialfusion} & 28.31 & .941 & 29.83 & .949 & 32.66 & \textbf{.958} & .110 & -- & 122 & 122 \\
DiffReg-PBIR~\cite{diffusion_material_regularization_eccv} & 31.31 & .958 & 29.94 & .958 & 32.54 & .949 & .098 & -- & 22.5 & 128 \\
\cmidrule{1-11}
Ours & \textbf{33.08} & \textbf{.966} & \textbf{32.26} & \textbf{.965} & \textbf{34.19} & \underline{.953} & \underline{.074} & -- & \textbf{6.1} & \underline{111} \\
\quad +\,learned metallic & \underline{32.44} & \underline{.964} & \underline{32.04} & \underline{.964} & \underline{34.14} & \underline{.953} & \textbf{.065} & -- & \underline{6.3} & \underline{111} \\
\specialrule{\lightrulewidth}{\aboverulesep}{1.5pt}\specialrule{\lightrulewidth}{0pt}{\belowrulesep}
\multicolumn{11}{c}{\textbf{TensoIR-Synthetic}~\cite{tensoir} } \\
\midrule
Neural-PBIR~\cite{neuralpbir2023} & 24.21 & .894 & 24.72 & .913 & 27.40 & .920 & .229 & 5.05 & \underline{6.5} & \textbf{17} \\
MaterialFusion~\cite{litman2025materialfusion} & 25.46 & .928 & 25.93 & .920 & 28.72 & .937 & \textbf{.208} & 5.12 & 123.5 & 124 \\
DiffReg-PBIR~\cite{diffusion_material_regularization_eccv} & 27.98 & \underline{.941} & 27.01 & .933 & 29.20 & .933 & .268 & \underline{4.10} & 22.2 & 124 \\
\cmidrule{1-11}
Ours & \underline{28.72} & \textbf{.950} & \underline{28.41} & \underline{.944} & \textbf{31.40} & \underline{.946} & .234 & \textbf{3.35} & \textbf{5.9} & \underline{107} \\
\quad +\,learned metallic & \textbf{29.27} & \textbf{.950} & \textbf{28.68} & \textbf{.947} & \underline{31.35} & \textbf{.947} & \underline{.221} & \textbf{3.35} & 6.6 & \underline{107} \\
\bottomrule
\end{tabular}
\end{table*}

\section{Experiments}
\label{sec:experiments}
\begin{figure}[t]
\centering
\setlength{\tabcolsep}{.4pt}
\renewcommand{\arraystretch}{0}
\begin{tabular}{c cccc}
&\parbox[b]{0.238\linewidth}{\centering\fontsize{7}{8}\selectfont MaterialFusion~\cite{litman2025materialfusion}}&\parbox[b]{0.238\linewidth}{\centering\fontsize{7}{8}\selectfont DiffReg-PBIR~\cite{diffusion_material_regularization_eccv}}&\parbox[b]{0.238\linewidth}{\centering\fontsize{7}{8}\selectfont Ours}&\parbox[b]{0.238\linewidth}{\centering\fontsize{7}{8}\selectfont Ground truth}\\[2pt]
\rotatebox{90}{\parbox{0.305\linewidth}{\centering\fontsize{6}{7}\selectfont Novel view}}&\begin{minipage}[t]{0.238\linewidth}\includegraphics[width=\linewidth]{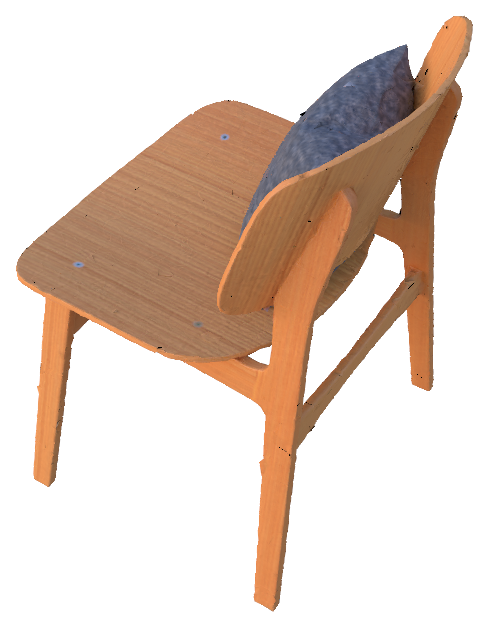}\llap{\raisebox{1.5pt}{\textcolor{black}{\fontsize{7}{8}\selectfont 34.65}}\hspace{1.5pt}}\end{minipage}&\begin{minipage}[t]{0.238\linewidth}\includegraphics[width=\linewidth]{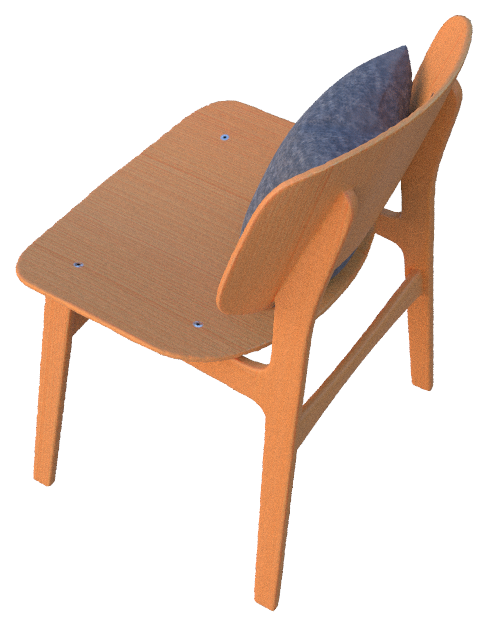}\llap{\raisebox{1.5pt}{\textcolor{black}{\fontsize{7}{8}\selectfont 34.23}}\hspace{1.5pt}}\end{minipage}&\begin{minipage}[t]{0.238\linewidth}\includegraphics[width=\linewidth]{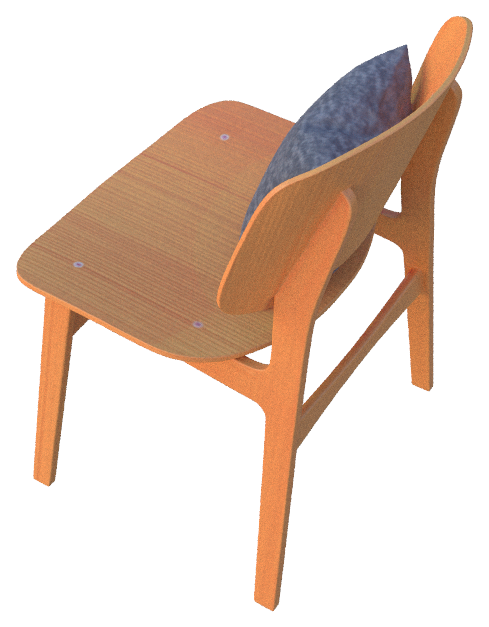}\llap{\raisebox{1.5pt}{\textcolor{black}{\fontsize{7}{8}\selectfont \textbf{35.72}}}\hspace{1.5pt}}\end{minipage}&\begin{minipage}[t]{0.238\linewidth}\includegraphics[width=\linewidth]{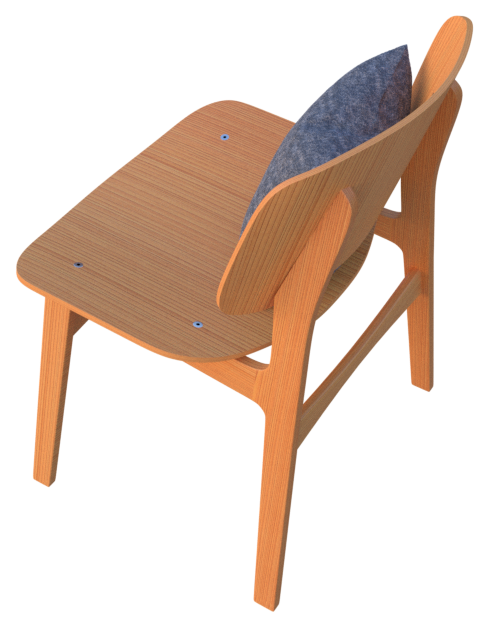}\end{minipage}\\[.8pt]
\rotatebox{90}{\parbox{0.305\linewidth}{\centering\fontsize{6}{7}\selectfont Relighting}}&\begin{minipage}[t]{0.238\linewidth}\includegraphics[width=\linewidth]{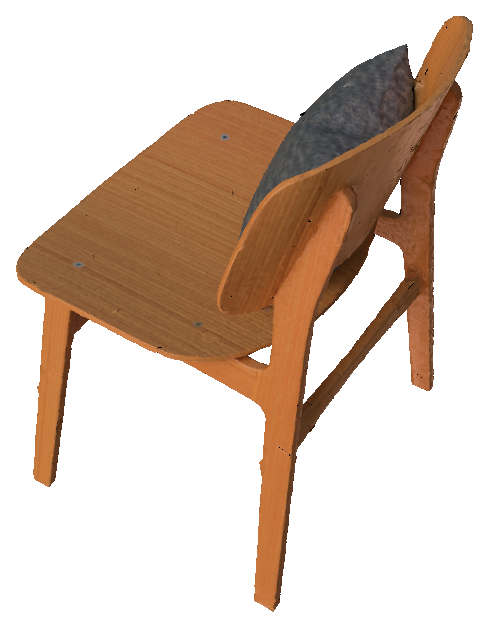}\llap{\raisebox{1.5pt}{\textcolor{black}{\fontsize{7}{8}\selectfont 31.67}}\hspace{1.5pt}}\end{minipage}&\begin{minipage}[t]{0.238\linewidth}\includegraphics[width=\linewidth]{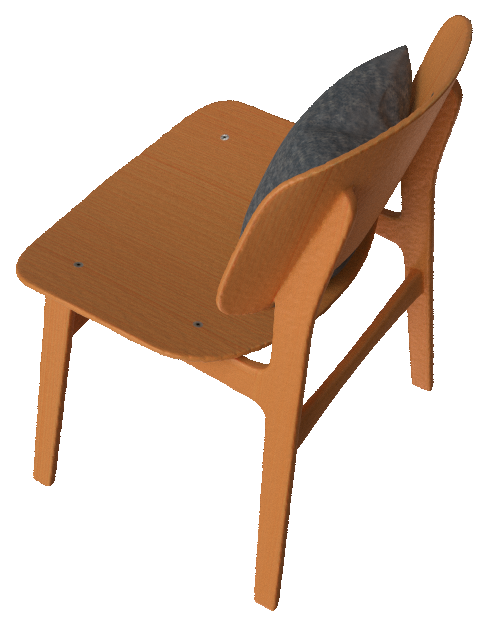}\llap{\raisebox{1.5pt}{\textcolor{black}{\fontsize{7}{8}\selectfont 31.69}}\hspace{1.5pt}}\end{minipage}&\begin{minipage}[t]{0.238\linewidth}\includegraphics[width=\linewidth]{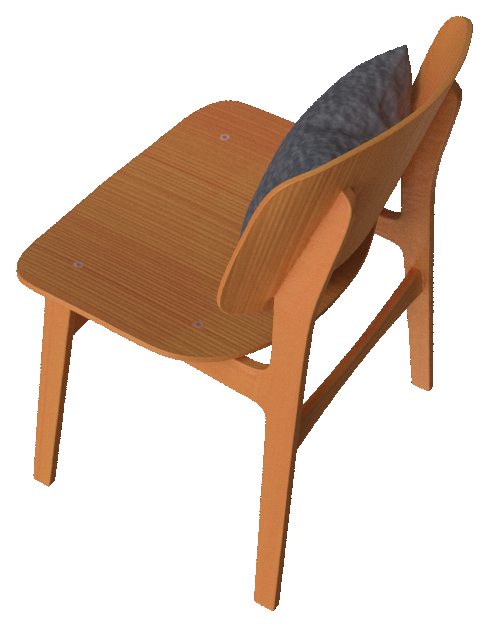}\llap{\raisebox{1.5pt}{\textcolor{black}{\fontsize{7}{8}\selectfont \textbf{34.02}}}\hspace{1.5pt}}\end{minipage}&\begin{minipage}[t]{0.238\linewidth}\includegraphics[width=\linewidth]{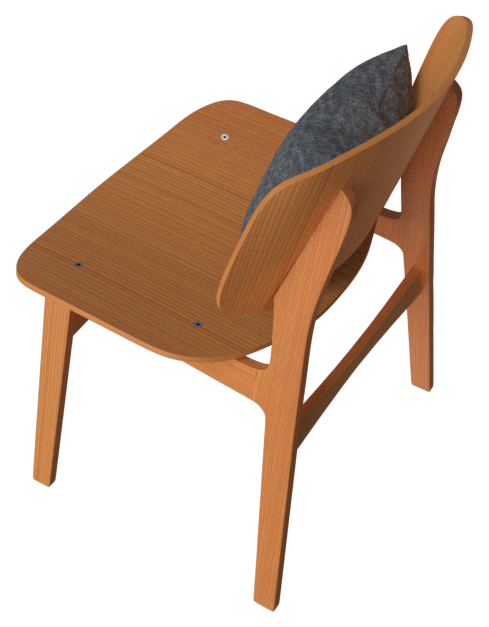}\end{minipage}\\[.8pt]
\multicolumn{5}{c}{}\\[3pt]
\rotatebox{90}{\parbox{0.143\linewidth}{\centering\fontsize{6}{7}\selectfont Novel view}}&\begin{minipage}[t]{0.238\linewidth}\includegraphics[width=\linewidth]{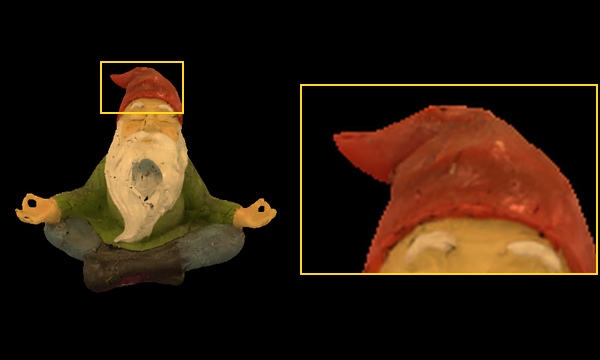}\llap{\raisebox{1.5pt}{\textcolor{white}{\fontsize{7}{8}\selectfont 30.85}}\hspace{1.5pt}}\end{minipage}&\begin{minipage}[t]{0.238\linewidth}\includegraphics[width=\linewidth]{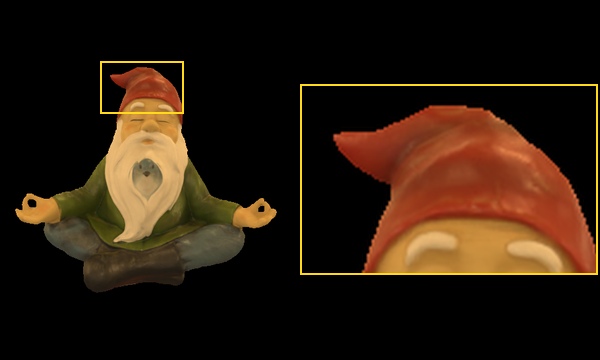}\llap{\raisebox{1.5pt}{\textcolor{white}{\fontsize{7}{8}\selectfont 34.18}}\hspace{1.5pt}}\end{minipage}&\begin{minipage}[t]{0.238\linewidth}\includegraphics[width=\linewidth]{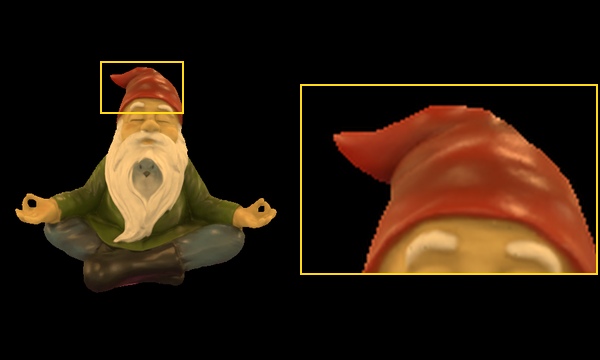}\llap{\raisebox{1.5pt}{\textcolor{white}{\fontsize{7}{8}\selectfont \textbf{36.25}}}\hspace{1.5pt}}\end{minipage}&\begin{minipage}[t]{0.238\linewidth}\includegraphics[width=\linewidth]{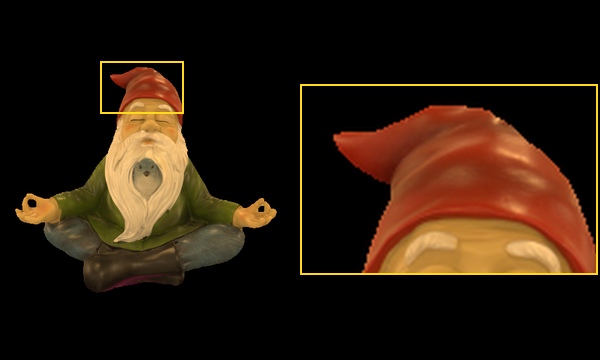}\end{minipage}\\[.8pt]
\rotatebox{90}{\parbox{0.143\linewidth}{\centering\fontsize{6}{7}\selectfont Relighting}}&\begin{minipage}[t]{0.238\linewidth}\includegraphics[width=\linewidth]{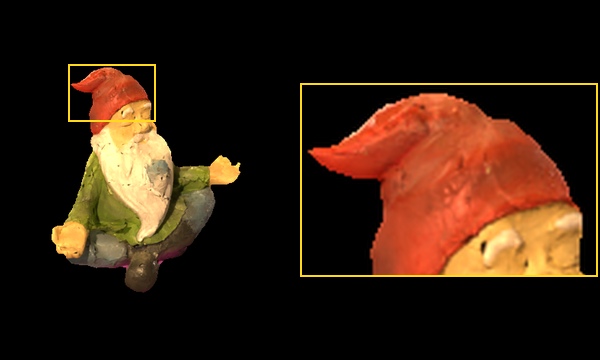}\llap{\raisebox{1.5pt}{\textcolor{white}{\fontsize{7}{8}\selectfont 24.65}}\hspace{1.5pt}}\end{minipage}&\begin{minipage}[t]{0.238\linewidth}\includegraphics[width=\linewidth]{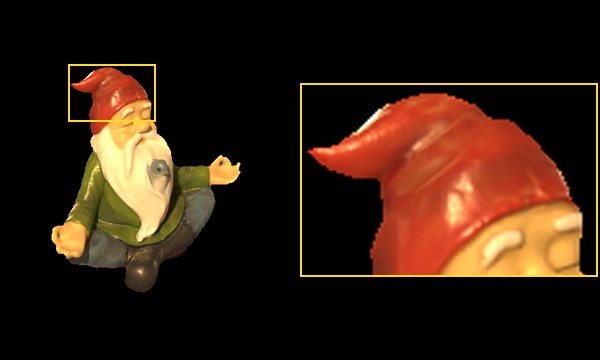}\llap{\raisebox{1.5pt}{\textcolor{white}{\fontsize{7}{8}\selectfont 27.34}}\hspace{1.5pt}}\end{minipage}&\begin{minipage}[t]{0.238\linewidth}\includegraphics[width=\linewidth]{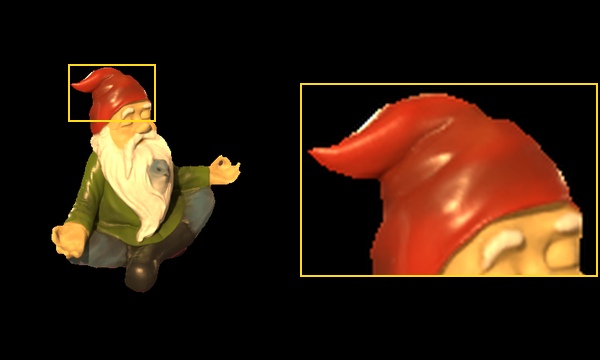}\llap{\raisebox{1.5pt}{\textcolor{white}{\fontsize{7}{8}\selectfont \textbf{28.78}}}\hspace{1.5pt}}\end{minipage}&\begin{minipage}[t]{0.238\linewidth}\includegraphics[width=\linewidth]{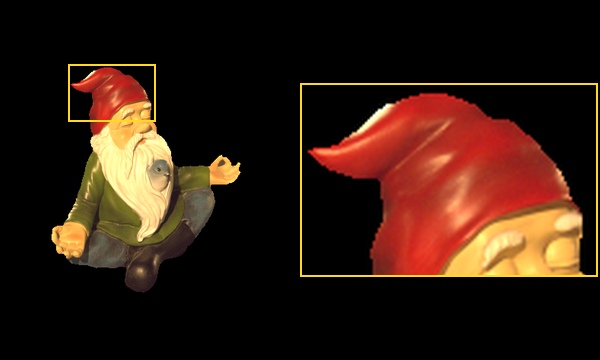}\end{minipage}\\[.8pt]
\multicolumn{5}{c}{}\\[3pt]
&\fontsize{7}{8}\selectfont Neural-PBIR~\cite{neuralpbir2023}&\fontsize{7}{8}\selectfont DiffReg-PBIR&\fontsize{7}{8}\selectfont Ours&\fontsize{7}{8}\selectfont Ground truth\\[1.5pt]
\rotatebox{90}{\parbox{0.143\linewidth}{\centering\fontsize{6}{7}\selectfont Relighting}}&\begin{minipage}[t]{0.238\linewidth}\includegraphics[width=\linewidth]{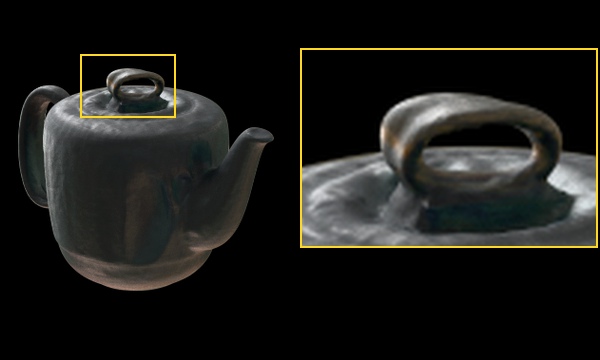}\llap{\raisebox{1.5pt}{\textcolor{white}{\fontsize{7}{8}\selectfont 32.44}}\hspace{1.5pt}}\end{minipage}&\begin{minipage}[t]{0.238\linewidth}\includegraphics[width=\linewidth]{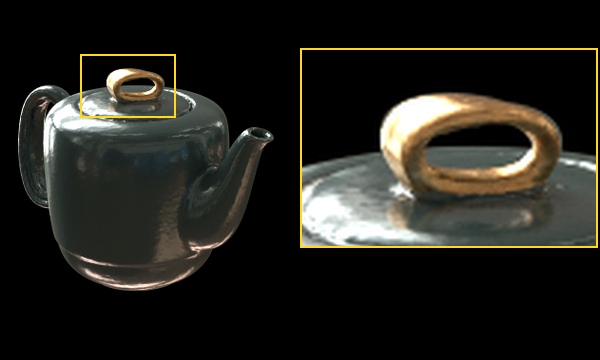}\llap{\raisebox{1.5pt}{\textcolor{white}{\fontsize{7}{8}\selectfont 37.37}}\hspace{1.5pt}}\end{minipage}&\begin{minipage}[t]{0.238\linewidth}\includegraphics[width=\linewidth]{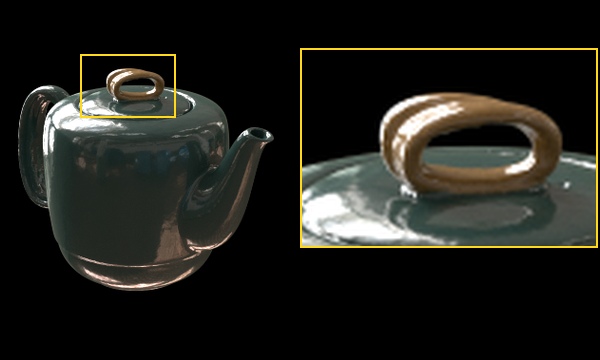}\llap{\raisebox{1.5pt}{\textcolor{white}{\fontsize{7}{8}\selectfont \textbf{41.17}}}\hspace{1.5pt}}\end{minipage}&\begin{minipage}[t]{0.238\linewidth}\includegraphics[width=\linewidth]{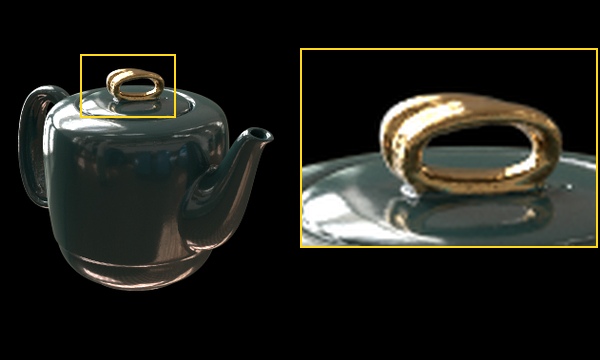}\end{minipage}\\[.8pt]
\end{tabular}
\vspace{-4pt}
\caption{\textbf{Qualitative comparison.} Novel views and relighting on Synthetic4Relight Chair and Stanford-ORB Gnome, and relighting on DTC-Synthetic Teapot. Boxes mark enlarged regions. Numbers are full-image PSNR (PSNR-H on Stanford-ORB).}
\label{fig:qual}
\end{figure}

\subsection{Experimental Details}
\label{sec:impl}

\paragraph{Datasets.}We evaluate our method on three synthetic datasets, Synthetic4Relight~\cite{zhang2022modeling}, TensoIR-Synthetic~\cite{tensoir} and DTC-Synthetic~\cite{diffusion_material_regularization_eccv}, and a real-world benchmark, Stanford-ORB~\cite{orb}.
We report the seven-scene \texttt{light} subset defined in the official Stanford-ORB benchmark code, and the full version in \cref{sec:orb_full}.

\paragraph{Baselines.}
We primarily compare our approach with mesh-based inverse rendering baselines: Neural-PBIR~\cite{neuralpbir2023},
 which does not employ learned priors,
  MaterialFusion~\cite{litman2025materialfusion}, which incorporates learned priors via score distillation,
 and DiffReg-PBIR~\cite{diffusion_material_regularization_eccv}, which uses learned priors through a joint bilateral regularization loss.
For baselines, we follow their released implementations and, where available, their released settings (\cref{sec:supp_baseline_settings}).

\paragraph{Metrics.}
We evaluate novel-view synthesis (NVS) under the captured illumination, relighting under novel illumination, and albedo, roughness and normals where ground truth is available.
We also report total pipeline and optimization time.
Unless stated otherwise, all quantitative results are means of three runs with different random seeds.
Standard deviations across seeds are at most 0.1\,dB in PSNR on the synthetic benchmarks (\cref{tab:prior_models_synthetic}).
All reported runtimes are measured on a single NVIDIA L40S GPU.

\subsection{Results on Synthetic Datasets}
\begin{table}[t]
\centering
\footnotesize
\setlength{\tabcolsep}{1.6pt}
\caption{\textbf{Stanford-ORB}~\cite{orb}. The seven scenes of the light subset, scored by the official benchmark  (PSNR-H on HDR images, PSNR-L on tone-mapped images).
+\,learned metallic additionally optimizes metallic instead of fixing it to zero.
}
\label{tab:orb_compact}
\begin{tabular}{l ccc ccc}
\toprule
& \multicolumn{3}{c}{Relighting} & \multicolumn{3}{c}{NVS} \\
\cmidrule(lr){2-4}\cmidrule(lr){5-7}
Method & PSNR-H & PSNR-L & SSIM & PSNR-H & PSNR-L & SSIM \\
\midrule
Neural-PBIR & 26.61 & 34.07 & .9822 & 30.45 & 37.49 & .9882 \\
MaterialFusion & 25.76 & 33.45 & .9761 & 28.42 & 36.07 & .9820 \\
DiffReg-PBIR & 27.80 & \textbf{36.16} & .9843 & 30.45 & \textbf{38.92} & .9887 \\
\cmidrule{1-7}
Ours & \textbf{27.93} & \underline{35.86} & \textbf{.9851} & \textbf{30.79} & \underline{38.70} & \textbf{.9903} \\
\quad +\,learned metallic & \underline{27.86} & 35.79 & \underline{.9847} & \underline{30.69} & 38.55 & \underline{.9901} \\
\bottomrule
\end{tabular}
\end{table}

\begin{table*}[t]
\centering
\small
\setlength{\tabcolsep}{4.2pt}
\caption{\textbf{How the prior enters material optimization.} All three variants share refined prior predictions, scene initialization, illumination parameterization and the 1000-step joint optimization schedule. Means over four scenes per synthetic dataset and seven Stanford-ORB light scenes.}
\label{tab:prior_use}
\begin{tabular}{l cccc cccc cc}
\toprule
& \multicolumn{4}{c}{Synthetic4Relight} & \multicolumn{4}{c}{TensoIR-Synthetic} & \multicolumn{2}{c}{ORB light} \\
\cmidrule(lr){2-5}\cmidrule(lr){6-9}\cmidrule(lr){10-11}
Variant & Albedo$\uparrow$ & Relit$\uparrow$ & NVS$\uparrow$ & Rough.$\downarrow$ & Albedo$\uparrow$ & Relit$\uparrow$ & NVS$\uparrow$ & Rough.$\downarrow$ & NL$\uparrow$ & NV$\uparrow$ \\
\midrule
Per-vertex, no prior & 29.91 & 31.37 & \underline{34.14} & .128 & 26.37 & 27.60 & \textbf{31.67} & \underline{.274} & 26.92 & \textbf{30.89} \\
Per-vertex, prior as a loss & \underline{31.20} & \underline{31.92} & 34.00 & \underline{.096} & \underline{27.55} & \underline{28.17} & \underline{31.44} & .292 & \textbf{27.94} & 30.60 \\
Prior-space reparameterization (ours) & \textbf{33.08} & \textbf{32.26} & \textbf{34.19} & \textbf{.074} & \textbf{28.72} & \textbf{28.41} & 31.40 & \textbf{.234} & \underline{27.93} & \underline{30.79} \\
\bottomrule
\end{tabular}
\vspace{-6pt}
\end{table*}

\begin{table}[t]
\centering
\small
\setlength{\tabcolsep}{5pt}
\caption{\textbf{DTC-Synthetic.} Relighting metrics using the released evaluation protocol: seven-scene means with three novel illuminations and 20 views per illumination.}
\label{tab:dtc}
\begin{tabular}{l ccc}
\toprule
Method & PSNR$\uparrow$ & SSIM$\uparrow$ & LPIPS$\downarrow$ \\
\midrule
Neural-PBIR~\cite{neuralpbir2023} & 38.88 & .9925 & .0100 \\
MaterialFusion~\cite{litman2025materialfusion} & 29.26 & .9739 & .0255 \\
DiffReg-PBIR~\cite{diffusion_material_regularization_eccv} & \underline{42.87} & \underline{.9961} & \underline{.0063} \\
Ours & \textbf{44.49} & \textbf{.9969} & \textbf{.0054} \\
\bottomrule

\end{tabular}
\vspace{-6pt}
\end{table}

\Cref{tab:main} summarizes two synthetic benchmarks.
Our method improves relighting PSNR over the strongest evaluated baseline by 1.50\,dB on Synthetic4Relight and 1.40\,dB on TensoIR-Synthetic.
The corresponding albedo PSNR gains are 1.77\,dB and 0.74\,dB, respectively.
NVS improves over DiffReg-PBIR by 1.65 and 2.20\,dB on the two benchmarks, respectively.
Learning metallic instead of fixing it to zero changes relighting PSNR by less than 0.3\,dB on either benchmark, raising albedo PSNR on TensoIR-Synthetic but lowering it on Synthetic4Relight (\cref{sec:metallic}).
Our method also outperforms all three baselines on all three DTC-Synthetic relighting metrics (\cref{tab:dtc}).
These are comparisons between complete systems, which differ from ours in more than the use of the prior, including geometry optimization, the material parameterization, the optimization schedule and the treatment of metallic.
In particular, the released Neural-PBIR and DiffReg-PBIR pipelines keep the reconstructed mesh fixed, whereas ours optimizes vertex positions jointly.
With the mesh fixed, our method remains above all baselines on Synthetic4Relight but can fall behind DiffReg-PBIR on TensoIR-Synthetic (\cref{sec:geometry_ablation}).
\Cref{sec:effect_preconditioning} isolates the contribution of the reparameterization within our pipeline.
\Cref{fig:qual} qualitatively compares novel-view synthesis and relighting, with additional qualitative examples in~\cref{sec:supp_qual}.

Joint optimization takes about 6 minutes on average, compared with 6.5--7.2 for Neural-PBIR, about 22 for DiffReg-PBIR and two hours for MaterialFusion.
\Cref{tab:main} also reports total pipeline time.
Prior inference accounts for much of the total pipeline time due to DiffusionRenderer's~\cite{liang2025diffusionrenderer} per-image diffusion sampling of the G-buffers.

\subsection{Results on Real-World Datasets}
\Cref{tab:orb_compact} reports the seven scenes of Stanford-ORB's official light subset.
Our method achieves higher novel-view HDR PSNR and SSIM than every baseline.
It also exceeds DiffReg-PBIR in relighting HDR PSNR and SSIM.
DiffReg-PBIR retains a higher PSNR on tone-mapped images in both tasks.
\Cref{sec:prior_quality} relates this smaller margin to the quality of the learned prior.

\section{Analysis}
\label{sec:analysis}

\subsection{Controlled Comparison on Prior Integration}
\label{sec:effect_preconditioning}
To isolate how the prior enters material optimization, we compare three variants with the same refined prior prediction inputs and joint geometry--illumination setup (\cref{tab:prior_use}): free per-vertex materials without a prior, per-vertex materials with DiffReg-PBIR's~\cite{diffusion_material_regularization_eccv} joint bilateral prior loss, and our prior-space reparameterization.
Both per-vertex variants use independent material variables; our method shares variables through the prior-space lattice while retaining a purely photometric objective.
Relative to the no-prior control, our method changes relighting PSNR by $+0.89$ and $+0.81$\,dB and albedo PSNR by $+3.17$ and $+2.35$\,dB on Synthetic4Relight and TensoIR-Synthetic, respectively.
The corresponding Stanford-ORB relighting HDR PSNR difference is $+1.01$\,dB.
Relative to the prior-as-a-loss control, our method changes relighting PSNR by $+0.34$ and $+0.24$\,dB and albedo PSNR by $+1.88$ and $+1.17$\,dB on Synthetic4Relight and TensoIR-Synthetic, respectively.
The corresponding Stanford-ORB relighting HDR PSNR difference is $-0.01$\,dB.
Additional comparisons with alternative material smoothing controls using Large Steps~\cite{nicolet2021large} and STAdam~\cite{chang2024spatiotemporal} are provided in~\cref{sec:smoothing_baselines}.

\subsection{Effect on Update Consistency}
\label{sec:update_consistency}
We then compare free per-vertex materials, per-vertex materials with a prior loss, and our reparameterization using the same refined prior predictions and optimization schedule.

During optimization, per-vertex updates can vary substantially with the sampled training views and noise induced by the stochastic nature of path tracing used in the differentiable renderer.
To measure this variability, we fix each variant's step-100 optimization state and compute the update for each of 16 randomly sampled pairs of training views.
We split these updates into a consistent component, which captures the direction that persists across sampled views, and noise, which captures the variation around that direction.
For the prior-loss variant, we further distinguish the photometric and prior-loss contributions to the consistent component.
Implementation details are provided in \cref{sec:update_consistency_details}.

As \cref{fig:step_split} shows, noise accounts for 92\% and 98\% of the update without a prior for albedo and roughness, respectively.   
Our reparameterization changes the consistent share from 8\% to 61\% for albedo and from 2\% to 20\% for roughness.
The prior loss contributes 40\% and 74\% of the total update for albedo and roughness, respectively.
By sharing variables across vertices with similar prior predictions, our reparameterization pools their photometric gradients and increases consistency across sampled views.

The two uses of the prior affect updates differently.
The auxiliary loss adds a gradient contribution that encourages agreement with the prior's similarity structure.
The prior-space reparameterization instead couples photometric updates through shared variables.

\subsection{Effect of Subspace Restriction}
\label{sec:subspace_restriction}
\begin{table}[t]
\centering
\small
\setlength{\tabcolsep}{1.2pt}
\caption{\textbf{Representational capacity of the lattice reparameterization.} Four-scene albedo PSNR means for Synthetic4Relight and TensoIR-Synthetic.  Oracle is from a single seed. Parentheses indicate the lattice's difference (dB) from the per-vertex variant.}
\label{tab:lattice_oracle}
\begin{tabular}{l rr cc cc}
\toprule
& \multicolumn{2}{c}{Point counts} & \multicolumn{2}{c}{Per-vertex} & \multicolumn{2}{c}{Prior lattice} \\
\cmidrule(lr){2-3}\cmidrule(lr){4-5}\cmidrule(lr){6-7}
Dataset & Mesh & Lattice & Oracle & Opt. & Oracle & Opt. \\
\midrule
S4R & 287.5K & 1.7K &\textbf{36.38}& 29.91 & 35.12 {\scriptsize($-$1.26)} & \textbf{33.08} {\scriptsize($+$3.17)} \\
TIR & 317.1K & 1.8K & \textbf{33.87} & 26.37 & 32.86 {\scriptsize($-$1.01)} & \textbf{28.72} {\scriptsize($+$2.35)} \\
\bottomrule
\end{tabular}
\vspace{-6pt}
\end{table}

Sharing lattice variables restricts the material fields that our method can represent.
To quantify the potential loss of albedo detail due to this restriction, we first fuse ground-truth albedo onto the final mesh to obtain a per-vertex oracle.
Using our reparameterization, we then solve for the albedo offsets at the prior-space lattice points by least-squares fitting to the oracle vertex albedos.
We evaluate both the per-vertex oracle and its lattice approximation using the same albedo protocol as the optimized results.
Details and additional results are provided in \cref{sec:lattice_oracle_details}.

\Cref{tab:lattice_oracle} shows that the prior lattice oracle is 1.26 and 1.01\,dB below the per-vertex oracle on Synthetic4Relight and TensoIR-Synthetic, respectively.
Yet independent per-vertex optimization falls 6.47 and 7.50\,dB short of its per-vertex oracle, whereas ours falls 2.04 and 4.14\,dB short of the lattice oracle and ends 3.17 and 2.35\,dB higher.
Albedo recovery in the unrestricted space is thus limited by what photometric optimization can reach, not by what the parameterization can represent.
Restricting the space gives up about 1\,dB of representable accuracy and recovers more than twice that amount in optimization, in line with the more consistent updates measured in \cref{sec:update_consistency}.
In \cref{sec:prior_quality}, we show that most of the remaining gap on Synthetic4Relight stems from the quality of the prior.

\subsection{Effect of Prior Quality}
\label{sec:prior_quality}
\definecolor{splitData}{HTML}{6D8FB8}%
\definecolor{splitPrior}{HTML}{C2412D}%
\definecolor{splitNoise}{HTML}{F7F5EF}%
\definecolor{splitEdge}{HTML}{8A8A8A}%
\providecommand{\splitswatch}[1]{{\setlength{\fboxsep}{0pt}\setlength{\fboxrule}{.4pt}\fcolorbox{#1}{#1}{\rule{0pt}{.65em}\rule{.65em}{0pt}}}}%
\begin{figure}[t]
\centering
\includegraphics[width=\columnwidth]{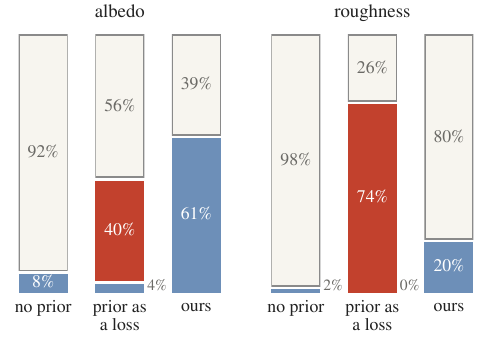}%
\vspace{-14pt}
\caption{\textbf{Update consistency across sampled views.}
Material-logit update energy at step 100 for the three controlled variants of~\cref{sec:effect_preconditioning}.
Their energy is split into noise
({\setlength{\fboxsep}{0pt}\setlength{\fboxrule}{.4pt}\fcolorbox{splitEdge}{splitNoise}{\rule{0pt}{.65em}\rule{.65em}{0pt}}})
and a consistent component attributed to the photometric loss (\splitswatch{splitData})
and the prior loss (\splitswatch{splitPrior}).
Mean over 15 scenes.
}
\label{fig:step_split}
\vspace{-6pt}
\end{figure}

\begin{figure}[t]
\centering
\includegraphics[width=\columnwidth]{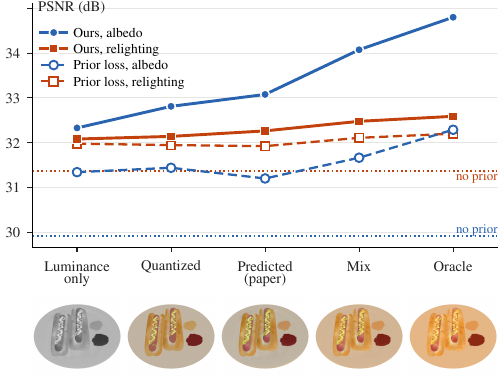}%
\vspace{-12pt}
\caption{\textbf{Effect of prior quality.}
Mean albedo and relighting PSNR on Synthetic4Relight of ours and the prior-loss control with increasing prior quality, and of the no-prior variant.
The two lowest levels apply material confusion to the predicted prior.
Thumbnails show the fused prior albedo on Hotdog used for that quality level.
}
\label{fig:prior_quality}
\vspace{-6pt}
\end{figure}

Our reparameterization relies on the prior to decide which vertices share variables, so its benefit depends on how well the prior separates distinct materials.
To quantify this dependence, we vary only the fused prior albedo on Synthetic4Relight and keep all other optimization settings fixed.
Starting from the refined DiffusionRenderer prior, we degrade it at a mild and a strong level, replace it with ground-truth albedo fused by the same procedure (oracle), or average the predicted and oracle priors (mix).
In \cref{fig:prior_quality}, the degradation is material confusion: we quantize the prior albedo to eight colors (mild) or keep only its luminance (strong), so that the prior distinguishes progressively fewer materials.
We run our method and the prior-loss control of \cref{sec:effect_preconditioning} with each prior.
Implementation details are provided in \cref{sec:prior_quality_details}.

As \cref{fig:prior_quality} shows, our albedo PSNR rises from 32.33 to 33.08 to 34.80\,dB from the luminance-only to the predicted to the oracle prior, and relighting PSNR from 32.08 to 32.26 to 32.59\,dB.
A better prior alone closes most of the gap left in \cref{sec:subspace_restriction}.
With the oracle prior, albedo comes close to the 35.12\,dB lattice oracle of \cref{tab:lattice_oracle}.
Even the luminance-only prior improves on optimization without a prior, and by more than the prior-loss control does.

The prior-loss control benefits far less from a better prior.
From the luminance-only to the oracle prior, its albedo PSNR rises by 0.95\,dB, compared with 2.47\,dB for ours.
Strengthening the loss does not close this gap: a $10\times$ prior-loss weight makes the control no more sensitive to the prior and lowers its albedo PSNR below that of optimization without a prior (\cref{fig:prior_loss_weight}).
TensoIR-Synthetic and two other degradations, blur and shadow baking, show similar trends (\cref{sec:prior_quality_details}).
Our advantage over the control thus grows with prior quality and narrows as the prior degrades, with the relighting margin falling from $+0.39$\,dB with the oracle prior to $+0.10$\,dB with the luminance-only prior.
This offers an explanation for Stanford-ORB, where the two differ by only 0.01\,dB in relighting (\cref{tab:prior_use}) and the per-view predictions already lose fine printed detail (\cref{sec:prior_detail}).
Our method should therefore benefit directly from more accurate predictors.

\paragraph{Limitations.}
Our method inherits the structure of the prior.
Materials that the prior does not separate share variables and cannot be told apart by the photometric loss, which removes fine detail missing from the predictions.

\section{Conclusion}
\label{sec:conclusion}
We have presented a reparameterization that integrates learned material priors into physically based inverse rendering without an auxiliary loss.
A fixed prior-space lattice defines shared material variables, which are optimized jointly with geometry and illumination under a purely photometric objective.
Within the same pipeline, this recovers more accurate albedo than a regularization loss built from the same prior and improves relighting on synthetic benchmarks, while remaining on par with the strongest baseline on Stanford-ORB.
Our analysis shows that sharing variables makes material updates more consistent across sampled views, that the lattice subspace costs little representational capacity, and that the benefit grows with the quality of the prior.
More accurate material predictors therefore translate directly into better reconstructions, and we view this as a step toward learned inverse rendering models that achieve such consistency in their own predictions.

\section*{Acknowledgements}
This work was supported by Institute of Information \& communications Technology Planning \& Evaluation (IITP) grant funded by the Korea government (MSIT) [NO.RS-2021-II211343, Artificial Intelligence Graduate School Program (Seoul National University)], 
the National Research Foundation of Korea (NRF) grant funded by the Korea government (MSIT) (No.  2022R1A3B1077720)
and the BK21 FOUR program of the Education and Research Program for Future ICT Pioneers, Seoul National University in 2026.

\section*{LLM Usage Statement}
In this work, we used generative AI, including Claude Code and OpenAI Codex, to assist with implementing the methods, executing experiments, and preparing figures and tables presenting the results, under continuous author supervision and interaction. 
Additionally, we used generative AI to refine the writing of the paper. 
We have reviewed all AI-generated material and take responsibility for the final content of this work.

\par
{
    \small
    \bibliographystyle{ieeenat_fullname}
    \bibliography{main}
}

\clearpage
\ifdefined\nolinenumbers\nolinenumbers\fi
\setcounter{page}{1}
\renewcommand{\thepage}{S\arabic{page}}
\setcounter{section}{0}
\renewcommand{\thesection}{\Alph{section}}
\renewcommand{\theHsection}{supp.\arabic{section}}
\setcounter{table}{0}
\renewcommand{\thetable}{S\arabic{table}}
\renewcommand{\theHtable}{supp.\arabic{table}}
\setcounter{figure}{0}
\renewcommand{\thefigure}{S\arabic{figure}}
\renewcommand{\theHfigure}{supp.\arabic{figure}}
\setcounter{equation}{0}
\renewcommand{\theequation}{S\arabic{equation}}
\renewcommand{\theHequation}{supp.\arabic{equation}}
\maketitlesupplementary

\section{Results with MVInverse Priors}
\label{sec:alternative_priors}
The main paper reports our method with DiffusionRenderer~\cite{liang2025diffusionrenderer} priors.
This section repeats the main experiments and analyses with MVInverse~\cite{wu2026mvinverse} priors.
In each configuration, the same model also provides the normal maps that guide surface reconstruction. All other settings are shared.
The following sections report both configurations where available and denote them DR and MVI.

\subsection{Benchmark Results}
\Cref{tab:prior_models_synthetic,tab:prior_models_orb} report our method with MVInverse priors on the synthetic benchmarks and Stanford-ORB, respectively.
Both configurations use joint geometry, material and illumination optimization, and the DiffusionRenderer rows are the same as those in the main tables.
The two priors perform similarly: Ours (MVI) is within 0.21\,dB of Ours (DR) in every PSNR on the synthetic benchmarks and within 0.11\,dB on Stanford-ORB; on DTC-Synthetic its relighting PSNR is 0.22\,dB lower (\cref{tab:metallic_dtc}).

\begin{table*}[t]
\centering
\small
\setlength{\tabcolsep}{3.2pt}
\caption{\textbf{Synthetic benchmarks with alternative prior models.}
Means over four scenes each, with sample standard deviations over three seeds.
SSIM standard deviations are at most .001.
Opt.\ denotes joint optimization time. Total additionally includes setup, prior inference and mesh initialization.
+\,learned metallic additionally optimizes metallic instead of fixing it to zero.
}
\label{tab:prior_models_synthetic}
\begin{tabular}{l cccccccc cc}
\toprule
& \multicolumn{2}{c}{Albedo} & \multicolumn{2}{c}{Relighting} & \multicolumn{2}{c}{Novel view synthesis} & Roughness & Normal & \multicolumn{2}{c}{Time (min)} \\
\cmidrule(lr){2-3}\cmidrule(lr){4-5}\cmidrule(lr){6-7}\cmidrule(lr){8-8}\cmidrule(lr){9-9}\cmidrule(lr){10-11}
Method & PSNR$\uparrow$ & SSIM$\uparrow$ & PSNR$\uparrow$ & SSIM$\uparrow$ & PSNR$\uparrow$ & SSIM$\uparrow$ & MAE$\downarrow$ & MAE ($^\circ$)$\downarrow$ & Opt.$\downarrow$ & Total$\downarrow$ \\
\midrule
\multicolumn{11}{c}{\textbf{Synthetic4Relight}~\cite{zhang2022modeling} } \\
\midrule
Neural-PBIR~\cite{neuralpbir2023} & 29.54{\tiny$\pm$.03} & .933 & 30.76{\tiny$\pm$.01} & .958 & 32.06{\tiny$\pm$.01} & .940 & .133{\tiny$\pm$.002} & -- & 7.2 & \textbf{18} \\
MaterialFusion~\cite{litman2025materialfusion} & 28.31{\tiny$\pm$.02} & .941 & 29.83{\tiny$\pm$.02} & .949 & 32.66{\tiny$\pm$.06} & \textbf{.958} & .110{\tiny$\pm$.001} & -- & 122 & 122 \\
DiffReg-PBIR~\cite{diffusion_material_regularization_eccv} & 31.31{\tiny$\pm$.05} & .958 & 29.94{\tiny$\pm$.02} & .958 & 32.54{\tiny$\pm$.00} & .949 & .098{\tiny$\pm$.002} & -- & 22.5 & 128 \\
\cmidrule{1-11}
Ours (MVI) & \textbf{33.29}{\tiny$\pm$.09} & \textbf{.968} & \underline{32.23}{\tiny$\pm$.02} & \textbf{.966} & \underline{34.26}{\tiny$\pm$.02} & .954 & .073{\tiny$\pm$.003} & -- & 6.4 & \underline{29} \\
\quad +\,learned metallic & 31.89{\tiny$\pm$.04} & .964 & 32.03{\tiny$\pm$.02} & \underline{.965} & \textbf{34.29}{\tiny$\pm$.03} & \underline{.955} & \underline{.067}{\tiny$\pm$.003} & -- & 6.7 & \underline{29} \\
Ours (DR) & \underline{33.08}{\tiny$\pm$.03} & \underline{.966} & \textbf{32.26}{\tiny$\pm$.02} & \underline{.965} & 34.19{\tiny$\pm$.07} & .953 & .074{\tiny$\pm$.002} & -- & \textbf{6.1} & 111 \\
\quad +\,learned metallic & 32.44{\tiny$\pm$.05} & .964 & 32.04{\tiny$\pm$.02} & .964 & 34.14{\tiny$\pm$.07} & .953 & \textbf{.065}{\tiny$\pm$.002} & -- & \underline{6.3} & 111 \\
\specialrule{\lightrulewidth}{\aboverulesep}{1.5pt}\specialrule{\lightrulewidth}{0pt}{\belowrulesep}
\multicolumn{11}{c}{\textbf{TensoIR-Synthetic}~\cite{tensoir} } \\
\midrule
Neural-PBIR~\cite{neuralpbir2023} & 24.21{\tiny$\pm$.05} & .894 & 24.72{\tiny$\pm$.02} & .913 & 27.40{\tiny$\pm$.00} & .920 & .229{\tiny$\pm$.003} & 5.05{\tiny$\pm$.00} & 6.5 & \textbf{17} \\
MaterialFusion~\cite{litman2025materialfusion} & 25.46{\tiny$\pm$.04} & .928 & 25.93{\tiny$\pm$.04} & .920 & 28.72{\tiny$\pm$.09} & .937 & \textbf{.208}{\tiny$\pm$.001} & 5.12{\tiny$\pm$.00} & 123.5 & 124 \\
DiffReg-PBIR~\cite{diffusion_material_regularization_eccv} & 27.98{\tiny$\pm$.10} & .941 & 27.01{\tiny$\pm$.02} & .933 & 29.20{\tiny$\pm$.01} & .933 & .268{\tiny$\pm$.004} & 4.10{\tiny$\pm$.00} & 22.2 & 124 \\
\cmidrule{1-11}
Ours (MVI) & 28.51{\tiny$\pm$.02} & \underline{.947} & 28.21{\tiny$\pm$.04} & .942 & 31.29{\tiny$\pm$.03} & \underline{.946} & .233{\tiny$\pm$.004} & 3.52{\tiny$\pm$.01} & \underline{6.4} & \underline{27} \\
\quad +\,learned metallic & 28.20{\tiny$\pm$.01} & .945 & \underline{28.56}{\tiny$\pm$.03} & \underline{.945} & 31.27{\tiny$\pm$.07} & \textbf{.947} & .231{\tiny$\pm$.001} & \underline{3.51}{\tiny$\pm$.01} & 7.0 & 28 \\
Ours (DR) & \underline{28.72}{\tiny$\pm$.02} & \textbf{.950} & 28.41{\tiny$\pm$.02} & .944 & \textbf{31.40}{\tiny$\pm$.06} & \underline{.946} & .234{\tiny$\pm$.004} & \textbf{3.35}{\tiny$\pm$.01} & \textbf{5.9} & 107 \\
\quad +\,learned metallic & \textbf{29.27}{\tiny$\pm$.05} & \textbf{.950} & \textbf{28.68}{\tiny$\pm$.01} & \textbf{.947} & \underline{31.35}{\tiny$\pm$.07} & \textbf{.947} & \underline{.221}{\tiny$\pm$.002} & \textbf{3.35}{\tiny$\pm$.01} & 6.6 & 107 \\
\bottomrule
\end{tabular}
\end{table*}

\begin{table}[t]
\centering
\footnotesize
\setlength{\tabcolsep}{2.6pt}
\caption{\textbf{Stanford-ORB with alternative prior models}~\cite{orb}. The seven scenes of the light subset, scored by the official benchmark  (PSNR-H on HDR images, PSNR-L on tone-mapped images). MVI: MVInverse; DR: DiffusionRenderer.
}
\label{tab:prior_models_orb}
\begin{tabular}{l ccc ccc}
\toprule
& \multicolumn{3}{c}{Relighting} & \multicolumn{3}{c}{Novel view synthesis} \\
\cmidrule(lr){2-4}\cmidrule(lr){5-7}
Method & PSNR-H & PSNR-L & SSIM & PSNR-H & PSNR-L & SSIM \\
\midrule
Neural-PBIR & 26.61 & 34.07 & .9822 & 30.45 & 37.49 & .9882 \\
MaterialFusion & 25.76 & 33.45 & .9761 & 28.42 & 36.07 & .9820 \\
DiffReg-PBIR & 27.80 & \textbf{36.16} & .9843 & 30.45 & \textbf{38.92} & .9887 \\
\cmidrule{1-7}
Ours (MVI) & \underline{27.82} & 35.85 & \underline{.9849} & \underline{30.72} & 38.63 & \underline{.9902} \\
Ours (DR) & \textbf{27.93} & \underline{35.86} & \textbf{.9851} & \textbf{30.79} & \underline{38.70} & \textbf{.9903} \\
\bottomrule
\end{tabular}
\end{table}

\subsection{Controlled Comparison with MVInverse}
\label{sec:prior_integration_mvi}
\begin{table*}[t]
\centering
\small
\setlength{\tabcolsep}{4.2pt}
\caption{\textbf{How the prior enters material optimization.} The variants share refined MVInverse predictions, scene initialization, illumination parameterization and the joint optimization schedule.
ORB light reports HDR PSNR.
}
\label{tab:prior_use_mvi}
\begin{tabular}{l cccc cccc cc}
\toprule
& \multicolumn{4}{c}{Synthetic4Relight} & \multicolumn{4}{c}{TensoIR-Synthetic} & \multicolumn{2}{c}{ORB light} \\
\cmidrule(lr){2-5}\cmidrule(lr){6-9}\cmidrule(lr){10-11}
Variant & Albedo$\uparrow$ & Relit$\uparrow$ & NVS$\uparrow$ & Rough.$\downarrow$ & Albedo$\uparrow$ & Relit$\uparrow$ & NVS$\uparrow$ & Rough.$\downarrow$ & NL$\uparrow$ & NV$\uparrow$ \\
\midrule
Per-vertex, no prior & 29.73 & 31.35 & \underline{34.10} & .129 & 26.23 & 27.46 & \textbf{31.50} & \underline{.276} & 26.87 & \textbf{30.85} \\
Per-vertex, prior as a loss & \underline{31.48} & \underline{31.95} & 33.84 & \underline{.094} & \underline{27.43} & \underline{27.93} & 31.19 & .288 & \underline{27.80} & 30.28 \\
Prior-space reparameterization (ours) & \textbf{33.29} & \textbf{32.23} & \textbf{34.26} & \textbf{.073} & \textbf{28.51} & \textbf{28.21} & \underline{31.29} & \textbf{.233} & \textbf{27.82} & \underline{30.72} \\
\bottomrule
\end{tabular}
\end{table*}

We do the same controlled comparison using MVInverse priors as we did in~\cref{sec:effect_preconditioning} (\cref{tab:prior_use_mvi}): free per-vertex materials without a prior, per-vertex materials with DiffReg-PBIR's~\cite{diffusion_material_regularization_eccv} joint bilateral loss, and our prior-space reparameterization.
Compared to the prior-as-a-loss variant, our reparameterization shows better quality on every metric and dataset.
These results favor our prior-guided reparameterization, which restricts the material search space while retaining a purely photometric objective, over the prior-loss formulation.

\subsection{Update Consistency with MVInverse}
We do the same update consistency analysis using MVInverse priors as we did in~\cref{sec:update_consistency}.  
Implementation details are provided in \cref{sec:update_consistency_details}.

The same trends hold for MVInverse priors. 
As \cref{fig:step_split_mvi} shows, noise accounts for more than 90\% of the update energy without a prior.
Our reparameterization increases the consistent share several-fold, e.g., from 8\% to 54\% for albedo, whereas the prior-loss variant increases consistency primarily from the added prior term. 

\definecolor{splitData}{HTML}{6D8FB8}%
\definecolor{splitPrior}{HTML}{C2412D}%
\definecolor{splitNoise}{HTML}{F7F5EF}%
\definecolor{splitEdge}{HTML}{8A8A8A}%
\providecommand{\splitswatch}[1]{{\setlength{\fboxsep}{0pt}\setlength{\fboxrule}{.4pt}\fcolorbox{#1}{#1}{\rule{0pt}{.65em}\rule{.65em}{0pt}}}}%
\begin{figure}[t]
\centering
\includegraphics[width=\columnwidth]{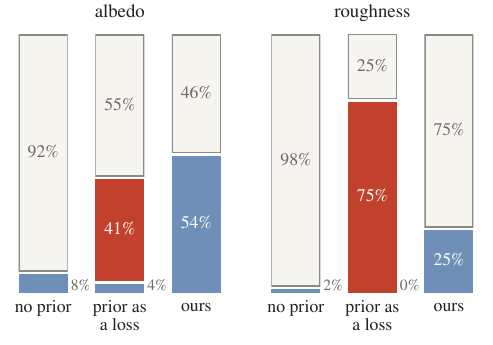}%
\vspace{-14pt}
\caption{\textbf{Update consistency across sampled views.}
Material-logit update energy at step 100 for the three variants of~\cref{tab:prior_use_mvi}.
Their energy is split into noise
({\setlength{\fboxsep}{0pt}\setlength{\fboxrule}{.4pt}\fcolorbox{splitEdge}{splitNoise}{\rule{0pt}{.65em}\rule{.65em}{0pt}}})
and a consistent component attributed to the photometric loss (\splitswatch{splitData})
and the prior loss (\splitswatch{splitPrior}), as defined in \cref{sec:update_consistency_details}.
MVInverse priors; mean over 15 scenes.
}
\label{fig:step_split_mvi}
\vspace{-6pt}
\end{figure}

\section{Metallic Estimation}
\label{sec:metallic}
The experiments of our main paper fix metallic to zero.
This section examines how metallic can instead be jointly optimized using our prior-space lattice.

\paragraph{Ambiguity with base color.}
In the metallic workflow~\cite{burley2012physically}, base color acts as diffuse albedo at $m=0$ and as specular reflectance at $m=1$, and intermediate $m$ blends the two.
Different combinations of metallic, base color and roughness can therefore explain the same training images, so the photometric loss does not uniquely determine $m$.
A surface point is nevertheless either a dielectric or a conductor, and metallic is expected to be near binary.

We first show that a directly learned metallic settles at intermediate values, and that a binarization loss removes these values without resolving the underlying ambiguity.
We then describe a dielectric--metallic dual hypothesis that uses the binary nature of metallic more directly.
All variants use the same refined priors, scene initialization, lattice and 1000-step schedule as the fixed-zero setting.

\begin{figure*}[t]
\centering
\includegraphics[width=.8\linewidth]{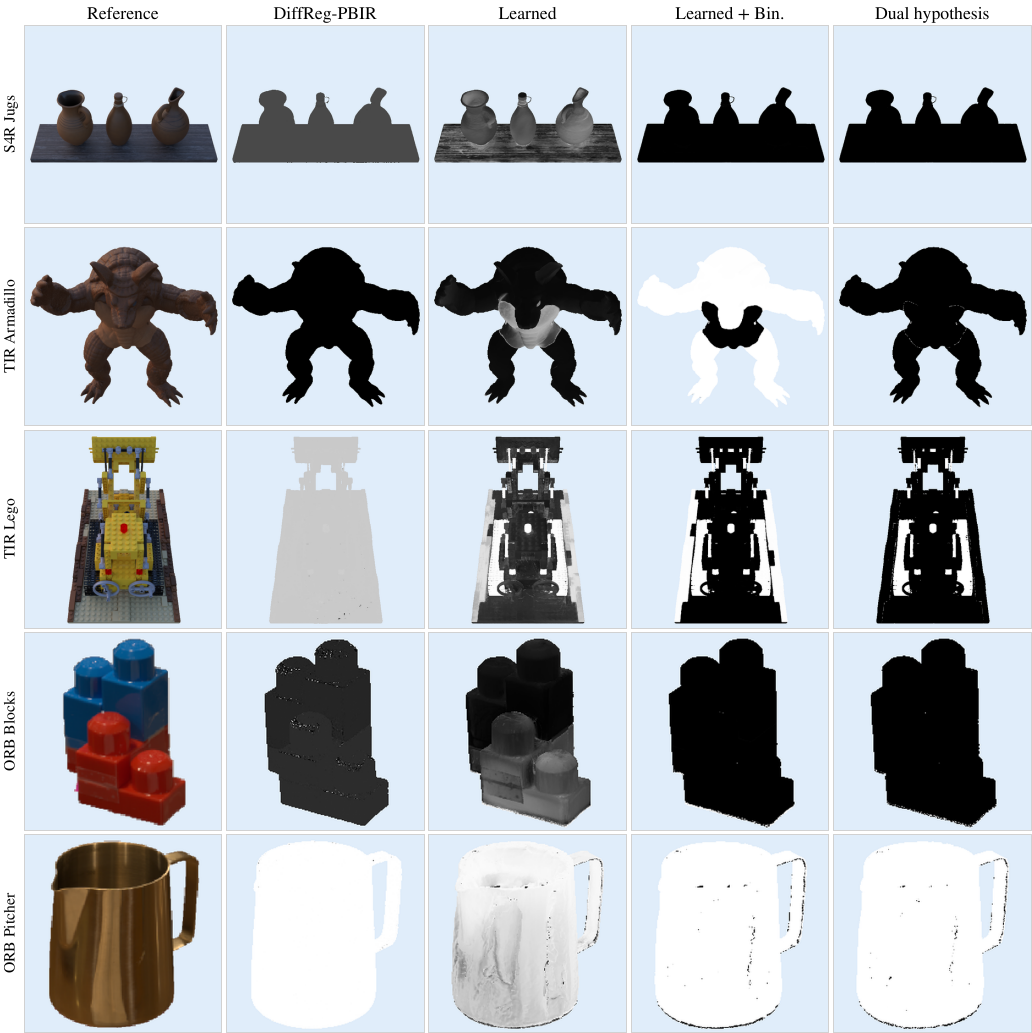}
\caption{\textbf{Estimated metallic.} Scalar metallic from one test view, shown linearly from black (0) to white (1).
DiffReg-PBIR and our learned-metallic variant return large regions of intermediate values.
The binarization loss removes these values but can push whole regions to the wrong side (Armadillo).
The dual hypothesis returns a binary map.
Our variants use the MVInverse prior.
Learned + Bin.\ uses weight $0.01$.}
\label{fig:metallic_maps}
\end{figure*}

\begin{table*}[t]
\centering
\small
\setlength{\tabcolsep}{3.6pt}
\caption{\textbf{Metallic estimation on the synthetic benchmarks.} PSNR (dB) for albedo, relighting and novel views, and roughness MAE.
All rows are three-seed means.}
\label{tab:metallic_synthetic}
\begin{tabular}{ll cccc cccc}
\toprule
& & \multicolumn{4}{c}{Synthetic4Relight (4)} & \multicolumn{4}{c}{TensoIR-Synthetic (4)} \\
\cmidrule(lr){3-6}\cmidrule(lr){7-10}
Init. & Metallic & Albedo$\uparrow$ & Relit$\uparrow$ & NVS$\uparrow$ & Rough.$\downarrow$ & Albedo$\uparrow$ & Relit$\uparrow$ & NVS$\uparrow$ & Rough.$\downarrow$ \\
\midrule
MVI & Fixed 0 & \textbf{33.29} & \underline{32.23} & 34.26 & .073 & \underline{28.51} & 28.21 & \underline{31.29} & .233 \\
MVI & Learned & 31.89 & 32.03 & \underline{34.29} & \textbf{.067} & 28.20 & \textbf{28.56} & 31.27 & .231 \\
MVI & Learned + Bin.\ ($0.01$) & 31.25 & 32.22 & 34.28 & .070 & 26.87 & 26.93 & 31.08 & \textbf{.222} \\
MVI & Learned + Bin.\ ($0.1$) & 31.26 & 32.21 & 34.25 & \underline{.069} & 26.90 & 26.94 & 31.08 & \underline{.223} \\
MVI & Dual hypothesis & \underline{33.04} & \textbf{32.24} & \textbf{34.30} & .073 & \textbf{29.32} & \underline{28.24} & \textbf{31.30} & \textbf{.222} \\
\midrule
DR & Fixed 0 & \textbf{33.08} & \textbf{32.26} & \textbf{34.19} & \underline{.074} & 28.72 & 28.41 & \underline{31.40} & .234 \\
DR & Learned & 32.44 & \underline{32.04} & \underline{34.14} & \textbf{.065} & \underline{29.27} & \textbf{28.68} & 31.35 & \textbf{.221} \\
DR & Learned + Bin.\ ($0.01$) & 31.14 & 30.88 & 33.93 & .112 & 28.64 & 27.15 & 31.22 & \underline{.226} \\
DR & Learned + Bin.\ ($0.1$) & 30.95 & 30.84 & 33.88 & .114 & 28.11 & 27.08 & 31.14 & .228 \\
DR & Dual hypothesis & \underline{33.07} & \textbf{32.26} & \textbf{34.19} & \underline{.074} & \textbf{29.33} & \underline{28.46} & \textbf{31.41} & .232 \\
\bottomrule
\end{tabular}
\end{table*}

\begin{table}[t]
\centering
\small
\setlength{\tabcolsep}{4pt}
\caption{\textbf{Metallic estimation on the Stanford-ORB light subset.} Official HDR PSNR (dB) over seven scenes.
All rows are three-seed means.}
\label{tab:metallic_orb}
\begin{tabular}{ll cc}
\toprule
Init. & Metallic & NL$\uparrow$ & NV$\uparrow$ \\
\midrule
MVI & Fixed 0 & \textbf{27.82} & \underline{30.72} \\
MVI & Learned & 27.67 & \textbf{30.74} \\
MVI & Learned + Bin.\ ($0.01$) & \underline{27.73} & 30.63 \\
MVI & Learned + Bin.\ ($0.1$) & \underline{27.73} & 30.54 \\
MVI & Dual hypothesis & 27.70 & \underline{30.72} \\
\midrule
DR & Fixed 0 & \underline{27.93} & \textbf{30.79} \\
DR & Learned & 27.86 & 30.69 \\
DR & Learned + Bin.\ ($0.01$) & \textbf{27.98} & 30.60 \\
DR & Learned + Bin.\ ($0.1$) & 27.89 & 30.45 \\
DR & Dual hypothesis & 27.92 & \underline{30.70} \\
\bottomrule
\end{tabular}
\end{table}

\begin{table}[t]
\centering
\small
\setlength{\tabcolsep}{5pt}
\caption{\textbf{Learned metallic on DTC-Synthetic.} Seven-scene means over three optimization seeds under the same released relighting protocol as \cref{tab:dtc}. Each pair differs only in whether metallic is fixed to zero or learned from a uniform value of 0.2.}
\label{tab:metallic_dtc}
\begin{tabular}{ll ccc}
\toprule
Initialization & Metallic & PSNR$\uparrow$ & SSIM$\uparrow$ & LPIPS$\downarrow$ \\
\midrule
Ours (MVI) & Fixed 0 & \underline{44.27} & \underline{.9962} & \underline{.0060} \\
Ours (MVI) & Learned & \textbf{44.88} & \textbf{.9964} & \textbf{.0057} \\
\midrule
Ours (DR) & Fixed 0 & \underline{44.49} & \underline{.9969} & \underline{.0054} \\
Ours (DR) & Learned & \textbf{45.36} & \textbf{.9972} & \textbf{.0051} \\
\bottomrule
\end{tabular}
\end{table}

\subsection{Learned Metallic}
\label{sec:learned_metallic}
We first treat metallic $m$ as a third material field optimized by the photometric loss alongside base color and roughness.
We initialize metallic at a uniform value of 0.2 and give it the same lattice, preconditioning and learning rate as roughness.

\paragraph{Convergence to intermediate values.}
The na\"ive optimization does not recover the expected near-binary metallic values.
With learned metallic, 44--73\% of the vertices end with $m\in(0.1,0.9)$ across the three benchmarks.
This behavior is not specific to our parameterization.
DiffReg-PBIR~\cite{diffusion_material_regularization_eccv}, which also optimizes a continuous metallic, returns intermediate values over more than 97\% of the visible surface in five of the 15 scenes, in the test view of \cref{fig:metallic_maps}.

\paragraph{Effect on accuracy.}
Relative to fixed-zero metallic, learned metallic changes relighting PSNR by at most 0.35\,dB on Synthetic4Relight, TensoIR-Synthetic and Stanford-ORB, and lowers Synthetic4Relight albedo PSNR by up to 1.4\,dB.
It clearly helps only on DTC-Synthetic, raising relighting PSNR by 0.6--0.9\,dB (\cref{tab:metallic_dtc}).
The added freedom therefore helps some benchmarks without consistently improving material recovery or relighting.

\subsection{Binarization Loss}
\label{sec:metallic_binarization}
A simple way to impose the binary prior on learned metallic is to penalize intermediate values,
\begin{equation}
  \Loss_{\mathrm{bin}}=\frac{1}{N}\sum_{i=1}^{N} m_i\,(1-m_i),
  \label{eq:metallic_binarization}
\end{equation}
over the $N$ vertices.
We add $w_{\mathrm{bin}}\Loss_{\mathrm{bin}}$ to the objective with $w_{\mathrm{bin}}\in\{0.01,0.1\}$, ramped in linearly between steps 300 and 600 so that the early fit is unconstrained.

The loss makes metallic nearly binary: on average at most 2.3\% of the vertices of a dataset remain at intermediate values.
However, it pushes each region toward whichever endpoint its ambiguous value happened to be closer to.
On TensoIR-Synthetic Armadillo, 82\%/96\% of the vertices become metal for MVI/DR (\cref{fig:metallic_maps}), and relighting PSNR drops by 5.6\,dB with MVI.
Relative to learned metallic, the loss lowers albedo PSNR by 0.6--1.5\,dB in every synthetic configuration and relighting PSNR by 1.2--1.6\,dB in all but one (MVI on Synthetic4Relight; \cref{tab:metallic_synthetic}).
Sharpening a continuous metallic thus fixes the outcome of the ambiguity instead of resolving it.

\subsection{Dielectric--Metallic Dual Hypothesis}
\label{sec:dual_hypothesis}
Instead of relaxing the binary choice into a continuous variable, we introduce a novel method that keeps both alternatives explicit and lets the photometric residual choose between them.

\paragraph{Hypotheses.}
Each vertex $i$ carries a dielectric hypothesis with base color and roughness $(\mathbf a_i^{D},r_i^{D})$, rendered with $m=0$, and a metal hypothesis $(\mathbf a_i^{M},r_i^{M})$, rendered with $m=1$.
All four fields use the prior-space parameterization of \cref{eq:change_variable} on the same lattice, with the learning rates of albedo and roughness.
Both base colors start at 0.3 and both roughness values at 0.6.
A binary label $z_i\in\{0,1\}$ selects the hypothesis that represents the vertex, with all vertices starting as dielectric.
Since neither hypothesis is ever rendered with intermediate metallic, base color has a single role within each of them.

\paragraph{Optimization.}
For every training view of a step, we render the scene twice.
The first render uses the selected hypothesis at each vertex, and its photometric loss updates geometry, illumination and the selected material fields.
The second render uses the unselected hypothesis at each vertex, with geometry and illumination detached, and updates only the unselected fields.
The unselected hypothesis thus remains a fitted alternative under the current geometry and illumination.
For efficiency, after step 400, the second render is evaluated every fourth step.

\paragraph{Evidence.}
For hypothesis $h\in\{D,M\}$, let $\rho^{h}(p)$ be the mean absolute sRGB error at pixel $p$ of the render in which the visible surface uses hypothesis $h$.
We distribute this error to the three vertices of the primary-hit triangle with barycentric weights $b_i(p)$ and accumulate
\begin{equation}
  R_i^{h}\leftarrow\gamma R_i^{h}+\sum_{p} b_i(p)\,\rho^{h}(p),\quad
  W_i^{h}\leftarrow\gamma W_i^{h}+\sum_{p} b_i(p),
  \label{eq:dual_evidence}
\end{equation}
with decay $\gamma=0.98$ per step.
The sums are incremented only on steps that render both hypotheses, so that the two are compared under the same scene state.

\paragraph{Selection.}
Each lattice point measures how much better the metal hypothesis explains its vertices than the dielectric one, and the vertices inherit this decision.
We splat the sums $\mathbf R^{h}$ and $\mathbf W^{h}$ to the lattice and smooth them with the operator of the material updates, $\B^{-1}\Sm$ (\cref{eq:material_ls_chart}), so that vertices with similar prior features share their evidence.
The pooled sums give the mean residual $L^{h}$ of each hypothesis at each lattice point, and the relative advantage of metal sets its weight,
\begin{equation}
  w=\sigma\!\left(\frac{d-\delta_t}{\tau_t}\right),\qquad d=\frac{L^{D}-L^{M}}{L^{D}+L^{M}}.
  \label{eq:dual_selection}
\end{equation}
We slice $w$ back to the vertices with normalized weights and select metal, $z_i=1$, where the result exceeds $1/2$.
The temperature $\tau_t$ decays geometrically from 0.3 to 0.01, and the margin $\delta_t$ increases linearly from 0 to 0.02 between steps 300 and 600, so that late near-ties resolve to dielectric.
Vertices outside the lattice decide from their own sums, and vertices without evidence for both hypotheses remain dielectric.

\paragraph{Final material and runtime.}
At the end of optimization, the selected hypothesis at each vertex determines its metallic value: 0 for dielectric and 1 for metal.
The final material also uses the base color and roughness of that hypothesis.
The additional render used to fit and evaluate the alternative hypothesis increases the total optimization time to about 1.3--1.4 times that of the fixed-zero setting.

\subsection{Results}
\label{sec:metallic_results}
\Cref{tab:metallic_synthetic,tab:metallic_orb} compare the four settings.

\paragraph{Synthetic benchmarks.}
The dual hypothesis exports binary metallic by construction.
On Synthetic4Relight it selects metal on 1\% of the vertices and stays within 0.25\,dB of fixed-zero albedo PSNR, whereas learned metallic and the binarization loss lose 0.6--2.1\,dB.
Its relighting PSNR matches or exceeds that of every other setting.
On TensoIR-Synthetic it attains the highest albedo PSNR, 0.6--0.8\,dB above fixed-zero metallic.
Its relighting PSNR is 1.3\,dB above the binarization loss and on par with fixed-zero metallic, but about 0.3\,dB below learned metallic, mostly on Lego, where a continuous metallic fits better.

\paragraph{Stanford-ORB.}
On the light subset, the settings of each initialization differ by at most 0.34\,dB in novel-view and 0.15\,dB in relighting PSNR.
With the MVInverse prior, the dual hypothesis selects the steel Pitcher as metal while leaving the other six objects mostly dielectric (\cref{fig:metallic_maps}).
With the DiffusionRenderer prior, the Pitcher remains dielectric.

\paragraph{Discussion.}
Among the settings that estimate metallic, the dual hypothesis is the only one that returns binary values without the albedo loss of learned metallic or the misassignments of the binarization loss.
Its selection still depends on photometric evidence alone.
A rough metal and a glossy dielectric can fit the images almost equally well, especially in darker areas, which can make hypothesis selection ambiguous.
On TensoIR-Synthetic Ficus, for example, 25\% of the vertices are selected as metal.
The main-paper results and the following comparisons retain fixed-zero metallic.

\section{Full Stanford-ORB Evaluation}
\label{sec:orb_full}
\begin{table}[t]
\centering
\footnotesize
\setlength{\tabcolsep}{1.7pt}
\caption{\textbf{Full Stanford-ORB.} All 42 captures, scored by the official benchmark (PSNR-H on HDR images, PSNR-L on tone-mapped images).
Ours rows are three-seed means. Rows marked $\dagger$ use one run per capture. Neural-PBIR uses a specular workflow without metallic. Dual denotes the dual hypothesis.}
\label{tab:orb_full}
\resizebox{\linewidth}{!}{%
\begin{tabular}{ll ccc ccc}
\toprule
& & \multicolumn{3}{c}{Relighting} & \multicolumn{3}{c}{NVS} \\
\cmidrule(lr){3-5}\cmidrule(lr){6-8}
Method & Metallic & PSNR-H & PSNR-L & SSIM & PSNR-H & PSNR-L & SSIM \\
\midrule
Neural-PBIR$^\dagger$ & -- & 26.02 & 33.30 & .9797 & 28.93 & 36.95 & .9869 \\
DiffReg-PBIR$^\dagger$ & Learned & 27.24 & \textbf{35.01} & .9808 & 29.54 & \textbf{38.42} & .9872 \\
\midrule
Ours (MVI) & Fixed 0 & 27.19 & 34.77 & .9813 & 29.36 & 37.59 & .9874 \\
Ours (MVI) & Learned & 27.18 & 34.76 & .9811 & \textbf{29.80} & 37.98 & \textbf{.9878} \\
Ours (MVI) & Dual & 27.32 & 34.93 & .9811 & 29.77 & 37.95 & .9877 \\
\midrule
Ours (DR) & Fixed 0 & 27.31 & 34.81 & \textbf{.9815} & 29.27 & 37.41 & .9873 \\
Ours (DR) & Learned & 27.46 & 34.96 & \textbf{.9815} & 29.58 & 37.67 & .9876 \\
Ours (DR) & Dual & \textbf{27.50} & 34.98 & \textbf{.9815} & 29.58 & 37.65 & .9875 \\
\bottomrule
\end{tabular}}
\end{table}

The main paper reports the seven-scene \texttt{light} subset defined in the official Stanford-ORB benchmark code.
\Cref{tab:orb_full} extends this evaluation to all 42 captures, 14 objects under three illuminations each, with the configurations of the main paper and the metallic settings of \cref{sec:metallic}.
With fixed-zero metallic, Ours (DR) is on par with DiffReg-PBIR in relighting PSNR-H and slightly below it in novel-view PSNR-H; DiffReg-PBIR retains a higher PSNR on tone-mapped images, as on the light subset.
Neural-PBIR is below both in every metric.
Estimating metallic matters more on the full set than on the light subset.
The dual hypothesis raises relighting and novel-view PSNR-H by 0.1--0.4\,dB for both priors, placing both above DiffReg-PBIR in PSNR-H.
The largest gains are on the three captures of the metallic Pepsi can, with 2.2--2.6\,dB in relighting PSNR-H.

\section{Full Pipeline Setup}
\label{sec:pipeline_setup}

\subsection{Prior Refinement Implementation}
\label{sec:prior_refinement_details}
For the refinement in \cref{sec:prior_refinement}, we apply a joint trilateral Gaussian filter to the predicted albedo of each view, guided by the input image, the pixel position and the predicted albedo itself.
Let $\mathbf a_v$ and $\mathbf c_v$ denote the predicted albedo and the input image of view $v$, and $\bar{\mathbf a}_v$ the albedo bilinearly upsampled to the refinement resolution.
For positions $\mathbf x$ measured in prediction pixels, we define
\begin{equation}
\begin{aligned}
 \mathbf z_v(\mathbf x)&=\left[\frac{\mathbf c_v(\mathbf x)}{\tau_c},\,\frac{\mathbf x}{\tau_x},\,\frac{\bar{\mathbf a}_v(\mathbf x)}{\tau_a}\right],\\
 k_v(\mathbf x,\mathbf y)&=\exp\!\left[-\tfrac12\|\mathbf z_v(\mathbf x)-\mathbf z_v(\mathbf y)\|^2\right],
\end{aligned}
\label{eq:prior_refinement_supp}
\end{equation}
where $\tau_c$, $\tau_x$ and $\tau_a$ are the bandwidths of the input-color, spatial and predicted-albedo terms.
Each foreground pixel is replaced by the $k_v$-weighted average of the predicted albedo of its valid neighbors, computed in linear RGB.
The input-color term aligns blurred albedo boundaries with the sharper edges of the input image, and the predicted-albedo term keeps apart pixels with similar input colors but clearly different predictions.
We use $\tau_c=0.03$, $\tau_a=0.20$ in sRGB and $\tau_x=12$ prediction pixels for all scenes and both predictors.

For efficiency, we evaluate the filter on $64\times64$ tiles of prediction pixels with stride 32.
All pixels of a tile share one set of neighbors, drawn from the tile and a 32-pixel margin around it, beyond which the spatial term falls below 3\%.
This set is uniformly subsampled to at most 1024 valid pixels.
Overlapping tile outputs are blended with Hann windows.

The refined albedo is computed at up to four times the prediction resolution, with each neighbor's input color averaged over its prediction-pixel footprint, and the upsampled prediction is kept within two prediction pixels of mask boundaries.
For Stanford-ORB, each HDR view is scaled so that the 99th-percentile foreground radiance maps to 0.8 before clamping to $[0,1]$ and sRGB encoding.

\subsection{Prior Fusion}
\label{sec:prior_fusion_details}
Each training view provides per-pixel predictions of (refined) sRGB albedo, roughness and metallic, which we upsample bilinearly to the input resolution.
A pixel is valid if it lies inside the object mask and its predictions are finite and within $[0,1]$, and we erode the valid region by two pixels.
A vertex is observed in a view if it is unoccluded, projects onto a valid pixel and satisfies $\cos\theta>0.3$, where $\theta$ is the angle between its normal and the viewing direction.
An observed vertex takes the predictions at the pixel it projects onto.
The prior feature of a vertex is the $\cos^2\theta$-weighted mean of its observations, and the weighted cross-view variance sets the lattice spacing.
Vertices observed in no training view remain unsupported.

\subsection{Optimization Setup}
\label{sec:optimization_setup}
We run 1000 joint steps, each rendering two training views at $800\times800$ with 32 samples per pixel and a maximum path depth of 4.
Materials start from constants (albedo $0.3$, roughness $0.6$).
Metallic is fixed to zero; \cref{sec:metallic} evaluates settings that learn it.
The lattice variables and unobserved per-vertex offsets use Adam rates of $0.025$ for albedo and $0.075$ for roughness.
The prior-space Large Steps strength is $\lambda=20$.
The lattice spacing is the median cross-view standard deviation of the prior features over vertices observed in at least two training views (\cref{sec:setup}).
Vertex positions are optimized by Large Steps~\cite{nicolet2021large} with $\lambda=19$ at a rate of $6\times10^{-4}$, and the
environment parameters by Adam at a rate of $0.04$ with a 10-step warm-up, cosine decay to $2\%$ and gradient-norm clipping at $0.1$.
Following Neural-PBIR~\cite{neuralpbir2023}, we use 128 spherical Gaussians for the first 250 steps, then switch to a Large Steps environment texture ($\lambda=6$).
The Gaussians are initialized from the observed background on Synthetic4Relight, DTC-Synthetic and Stanford-ORB, or from a uniform initialization on TensoIR-Synthetic, where the background is unobserved.
Vertices not seen in any training view are optimized directly without an assignment to the lattice because no prior prediction is available for them.
They account for a small fraction of the mesh vertices and have a negligible effect on optimization.

\subsection{Derivative Estimator}
\label{sec:estimator}
We render with Mitsuba~3~\cite{jakob2022dr} and estimate derivatives with its projective-sampling path tracer~\cite{zhang2023projective}, which accounts for the visibility discontinuities that vertex-position gradients require.
Each step uses 32 primal samples per pixel.
The derivative estimate uses 32 samples per pixel for the continuous term and 8 each for the primary- and indirect-visibility boundary terms, without a guiding structure for boundary sampling.
Per-vertex gradients of vertex positions are clipped at the 99th percentile of their norms before the update.

We make three changes to the stock estimator.
They leave the rendered images unchanged and affect only the gradients.
First, when evaluating the radiance difference across a visibility boundary, the stock estimator clamps the interpolated shading normal to the visible side only at the occluding surface; we apply the same clamp at the surface behind it, which removes spurious differences at creases of the interpolated normals.
Second, path vertices after the primary hit follow the moving surface, and their contributions include the relative change in triangle area; the indirect-visibility boundary term accordingly measures boundary motion relative to both endpoints of the path segment.
Third, at grazing angles the BSDF derivative is evaluated with the magnitude of the local cosines floored at $10^{-6}$.
In our experiments, the stock estimator reached similar quality when optimization completed, but occasionally produced non-finite vertex-position gradients, which we did not observe with these changes.

\subsection{Update Consistency Measurement}
\label{sec:update_consistency_details}
We evaluate each variant at its step-100 checkpoint, keeping geometry, materials, illumination and optimizer state fixed.
We sample 16 independent pairs of training views, with independent Monte Carlo rendering seeds, and compute the resulting material-logit updates on training-visible vertices.
The updates use the checkpoint's bias-corrected Adam second-moment normalization, without first-moment momentum, to isolate the current gradient variability.
The lattice variant retains its material preconditioning and slicing operations.
For updates $\mathbf u_k$, let $\bar{\mathbf u}$ be their sample mean and $s^2=\sum_k\|\mathbf u_k-\bar{\mathbf u}\|^2/(K-1)$ with $K=16$.
We report noise energy $s^2$ and consistent energy $\|\bar{\mathbf u}\|^2-s^2/K$, normalized by the total mean update energy.
For the prior-loss variant, we separate the photometric and prior-loss contributions through their inner products with the mean combined update, with the same finite-sample covariance correction.

\section{Material Large Steps Preconditioning}
\label{sec:material_ls}
\subsection{Formulation}
For the lattice values $\ML$ in \cref{eq:change_variable}, we use the Large Steps change of variables~\cite{nicolet2021large}
\begin{equation}
  \ML=\B^{-1}\W,\qquad \B=\I+\lambda\Lh,
  \label{eq:material_ls_chart}
\end{equation}
where $\Lh=(\mathbf D-\mathbf A)/\bar d$ is the Laplacian of the unit-weight lattice-neighbor graph with degree matrix $\mathbf D$ and adjacency matrix $\mathbf A$, normalized by its mean degree $\bar d$.
We initialize $\W$ at zero and use $\lambda=20$.
Let $\G=\partial\Loss/\partial\MV$ be the per-vertex material gradient.
Since $\B$ is symmetric, the gradient in these coordinates is
\begin{equation}
  \frac{\partial\Loss}{\partial\W}=\B^{-1}\Sm\G.
  \label{eq:material_ls_gradient}
\end{equation}
Before pooling, we clip $\G$ to limit outliers.
Let $c$ be the 99th percentile of the nonzero row norms of $\G$.
We define $\widetilde{\G}$ by replacing each nonzero row $\mathbf g_i$ with $\widetilde{\mathbf g}_i=\mathbf g_i\min(1,c/\|\mathbf g_i\|)$, leaving zero rows unchanged.
Using this clipped gradient, we update
\begin{equation}
  \W\leftarrow\W-\eta\,\mathrm{Adam}\!\left(\B^{-1}\Sm\widetilde{\G}\right).
  \label{eq:cell_update}
\end{equation}
Slicing $\Delta\ML=\B^{-1}\Delta\W$ back to vertices yields the per-vertex material update.
The two solves smooth the pooled gradients and the resulting updates on the prior-space graph.
The fixed matrix $\B$ is shared by all material channels, so we factorize it once with a sparse Cholesky decomposition and reuse the factorization for every solve.
Since $\B$ is invertible, this preconditioning changes the optimization trajectory without changing the material fields representable by the lattice.

\subsection{Ablation}
\label{sec:material_ls_ablation}
We isolate this preconditioning by comparing $\lambda=0$ ($\B=\I$) and our default $\lambda=20$.
Both variants use the same refined predictions and post-refinement spacing, initial scene, optimizer settings and 1000-step joint schedule.
We evaluate both MVI-MVI and DR-DR initializations over all 15 scenes, with one seed per scene.

\begin{table*}[t]
\centering
\small
\setlength{\tabcolsep}{4pt}
\caption{\textbf{Effect of material Large Steps on a single lattice.} Each pair shares the same refined predictions, initial scene, lattice spacing, illumination and 1000-step schedule.
One seed per scene, means over four Synthetic4Relight, four TensoIR-Synthetic and seven Stanford-ORB scenes. ORB uses official HDR PSNR. 
Geometry and illumination retain their Large Steps parameterizations in both variants. MVI-MVI and DR-DR use MVInverse and DiffusionRenderer predictions, respectively, for both normal guidance and material priors.}
\label{tab:material_ls}
\begin{tabular}{l c cccc cccc cc}
\toprule
& & \multicolumn{4}{c}{Synthetic4Relight} & \multicolumn{4}{c}{TensoIR-Synthetic} & \multicolumn{2}{c}{Stanford-ORB} \\
\cmidrule(lr){3-6}\cmidrule(lr){7-10}\cmidrule(lr){11-12}
Initialization & Material LS & Albedo & Relit & NVS & Rough.$\downarrow$ & Albedo & Relit & NVS & Rough.$\downarrow$ & NL & NV \\
\midrule
MVI-MVI & w/o ($\lambda=0$) & 32.75 & 32.22 & \textbf{34.33} & .085 & 27.70 & 28.06 & \textbf{31.45} & .255 & 27.57 & \textbf{30.99} \\
MVI-MVI & with ($\lambda=20$) & \textbf{33.21} & \textbf{32.25} & 34.28 & \textbf{.076} & \textbf{28.53} & \textbf{28.22} & 31.23 & \textbf{.233} & \textbf{27.86} & 30.74 \\
\cmidrule{1-12}
DR-DR & w/o ($\lambda=0$) & 32.87 & \textbf{32.32} & \textbf{34.27} & .079 & 28.32 & \textbf{28.44} & \textbf{31.44} & .250 & 27.88 & \textbf{30.98} \\
DR-DR & with ($\lambda=20$) & \textbf{33.06} & 32.28 & 34.20 & \textbf{.074} & \textbf{28.70} & 28.37 & 31.31 & \textbf{.237} & \textbf{27.95} & 30.77 \\
\bottomrule
\end{tabular}
\end{table*}

\Cref{tab:material_ls} reports the effect at a fixed optimization budget.
Material Large Steps improves albedo PSNR (by up to 0.8\,dB) and roughness MAE for both initializations on both synthetic datasets, at the cost of at most 0.25\,dB in novel-view PSNR.
Its effect on relighting is small and mixed, between $-0.07$ and $+0.29$\,dB.
We adopt the Large Steps preconditioner mainly for the consistent positive effect on albedo.

\section{Alternative Material Smoothing}
\label{sec:smoothing_baselines}
We compare our prior-space reparameterization with two controls that smooth per-vertex material updates without using a material prior: mesh Large Steps~\cite{nicolet2021large} and a mesh adaptation of STAdam~\cite{chang2024spatiotemporal}.
Both controls start from the no-prior variant and retain its uniform material initialization, learning rates, gradient clipping, geometry and illumination settings, and optimization schedule.
All variants share the refined DiffusionRenderer inputs used to initialize the scene.
We evaluate the 1000-step schedule on all 15 scenes.

\paragraph{Large Steps.}
We parameterize the material logits as $\MV=\mathbf{M}_0+\mathbf B^{-1}\mathbf U$, where $\mathbf B=\mathbf I+20\mathbf L_{\mathrm{mesh}}/\bar d$ and $\bar d$ is the mean mesh degree.
Adam updates $\mathbf U$, initialized at zero.
This uses the mesh connectivity to smooth material updates, whereas ours pools and smooths updates on the fixed prior-space lattice.

\paragraph{STAdam.}
We adapt STAdam's texture optimizer to per-vertex materials.
STAdam filters Adam's moment estimates~\cite{adam} over neighboring texels before forming each update.
Since this texture-grid filter does not apply to a mesh, we filter over mesh edges with bilateral diffusion, $\mathbf F=(\mathbf I+20\mathbf L_w/\bar d)^{-1}$, where $\mathbf L_w$ is the mesh Laplacian with edge weights $w_{ij}=\exp(-\|\mathbf c_i-\mathbf c_j\|/0.02)$ and $\mathbf c_i$ is the current albedo RGB or roughness of vertex $i$.
Each update divides the filtered first moment by the square root of the filtered second moment, while the stored moments remain unfiltered.
The diffusion systems are solved by Jacobi-preconditioned conjugate gradients with relative tolerance $10^{-5}$.

\begin{table*}[t]
\centering
\small
\setlength{\tabcolsep}{4pt}
\caption{\textbf{Material smoothing on the mesh and in prior space.} All five variants use the same refined DiffusionRenderer inputs and 1000-step budget. Means over four Synthetic4Relight, four TensoIR-Synthetic and seven Stanford-ORB light scenes. ORB uses official HDR PSNR.}
\label{tab:smoothing_baselines}
\begin{tabular}{l cccc cccc cc}
\toprule
& \multicolumn{4}{c}{Synthetic4Relight} & \multicolumn{4}{c}{TensoIR-Synthetic} & \multicolumn{2}{c}{ORB light} \\
\cmidrule(lr){2-5}\cmidrule(lr){6-9}\cmidrule(lr){10-11}
Variant & Albedo & Relit & NVS & Rough.$\downarrow$ & Albedo & Relit & NVS & Rough.$\downarrow$ & NL & NV \\
\midrule
No prior & 29.91 & 31.37 & 34.14 & .128 & 26.37 & 27.60 & 31.67 & .274 & 26.92 & 30.89 \\
Prior as a loss & 31.20 & 31.92 & 34.00 & .096 & 27.55 & 28.17 & 31.44 & .292 & \textbf{27.94} & 30.60 \\
Mesh Large Steps & 30.76 & 31.64 & \textbf{34.37} & .098 & 27.12 & 27.89 & \textbf{31.98} & .268 & 27.06 & \textbf{31.09} \\
STAdam (adaptation) & 29.94 & 31.40 & 34.23 & .119 & 26.45 & 27.64 & 31.75 & .277 & 27.00 & 30.92 \\
Ours & \textbf{33.08} & \textbf{32.26} & 34.19 & \textbf{.074} & \textbf{28.72} & \textbf{28.41} & 31.40 & \textbf{.234} & 27.93 & 30.79 \\
\bottomrule
\end{tabular}
\end{table*}

\paragraph{Results.}
Mesh Large Steps improves on the no-prior variant in novel-view and relighting PSNR on all three datasets at 1000 steps, and achieves the highest novel-view PSNR among the five variants (\cref{tab:smoothing_baselines}).
Our method nevertheless achieves 0.5--0.9\,dB higher relighting PSNR on all three datasets, along with higher synthetic albedo PSNR and lower roughness MAE.
The STAdam adaptation remains close to the no-prior variant in image PSNR.
These results show that smoothing through mesh connectivity helps optimization, but does not recover the relighting and albedo quality obtained by sharing updates through the material prior.

\section{Contribution of Geometry Optimization}
\label{sec:geometry_ablation}
We isolate the contribution of vertex-position updates by holding the initial reconstructed mesh fixed while optimizing materials and illumination.
Both variants use the same refined priors, scene initialization, optimizer settings and 1000-step schedule, except that the fixed variant keeps the initially reconstructed mesh fixed throughout the joint optimization stage and its final vertex positions are verified to equal the initial positions.

\begin{table*}[t]
\centering
\small
\setlength{\tabcolsep}{3.0pt}
\caption{\textbf{Contribution of geometry optimization.} Means over four scenes per dataset.
Fixed holds the initial reconstructed mesh constant while Joint also optimizes vertex positions, with every other optimization setting identical.
Baselines use their released settings where available (\cref{sec:supp_baseline_settings}). Times are in minutes. Opt.\ is the optimization stage and Total includes setup, prior inference and mesh initialization.}
\label{tab:geometry_ablation}
\begin{tabular}{l c cccccc cccccc}
\toprule
& & \multicolumn{6}{c}{Synthetic4Relight} & \multicolumn{6}{c}{TensoIR-Synthetic} \\
\cmidrule(lr){3-8}\cmidrule(lr){9-14}
Method & Geometry & Albedo$\uparrow$ & Relit$\uparrow$ & NVS$\uparrow$ & Rough.$\downarrow$ & Opt.$\downarrow$ & Total$\downarrow$ & Albedo$\uparrow$ & Relit$\uparrow$ & NVS$\uparrow$ & Rough.$\downarrow$ & Opt.$\downarrow$ & Total$\downarrow$ \\
\midrule
Neural-PBIR~\cite{neuralpbir2023} & Fixed & 29.54 & 30.76 & 32.06 & .133 & 7.2 & \textbf{18} & 24.21 & 24.72 & 27.40 & \underline{.229} & 6.5 & \textbf{17} \\
MaterialFusion~\cite{litman2025materialfusion} & Joint & 28.31 & 29.83 & 32.66 & .110 & 122 & 122 & 25.46 & 25.93 & 28.72 & \textbf{.208} & 123.5 & 124 \\
DiffReg-PBIR~\cite{diffusion_material_regularization_eccv} & Fixed & 31.31 & 29.94 & 32.54 & .098 & 22.5 & 128 & 27.98 & 27.01 & 29.20 & .268 & 22.2 & 124 \\
\midrule
Ours (MVI) & Fixed & 31.84 & 30.77 & 32.55 & .083 & \underline{2.7} & \underline{25} & 26.60 & 25.69 & 27.82 & .260 & \underline{2.9} & \underline{24} \\
Ours (MVI) & Joint & \textbf{33.29} & \underline{32.23} & \textbf{34.26} & \textbf{.073} & 6.4 & 29 & \underline{28.51} & \underline{28.21} & \underline{31.29} & .233 & 6.4 & 27 \\
\midrule
Ours (DR) & Fixed & 32.14 & 31.33 & 33.01 & .076 & \textbf{2.5} & 107 & 27.25 & 26.52 & 28.71 & .258 & \textbf{2.5} & 103 \\
Ours (DR) & Joint & \underline{33.08} & \textbf{32.26} & \underline{34.19} & \underline{.074} & 6.1 & 111 & \textbf{28.72} & \textbf{28.41} & \textbf{31.40} & .234 & 5.9 & 107 \\
\bottomrule
\end{tabular}
\end{table*}

With geometry fixed, both of our initializations retain higher albedo and relighting PSNR than all three baselines on Synthetic4Relight (\cref{tab:geometry_ablation}).
On TensoIR-Synthetic, the fixed variant of Ours (DR) is 0.5--0.7\,dB below DiffReg-PBIR in albedo, relighting and novel-view PSNR, with a lower roughness error.
Joint geometry optimization improves novel-view, relighting and albedo PSNR for both initializations on both datasets, by 0.9--3.5\,dB.
Fixing geometry for Ours (DR) likewise lowers relighting PSNR by 0.17\,dB on the Stanford-ORB light subset, where novel-view PSNR-H drops by 0.91\,dB, and by 1.97\,dB on DTC-Synthetic, with worse SSIM and LPIPS.

\section{Albedo Oracle Fits}
\label{sec:lattice_oracle_details}
We assess the albedo capacity of the fixed learned-prior lattice used in \cref{sec:subspace_restriction}.
For each of the four Synthetic4Relight and four TensoIR-Synthetic scenes and both prior models, we use the final mesh and the lattice of the 1000-step run with a single seed.
Both oracle variants use this same mesh.
The lattice coordinates and assignments are built from the run's learned prior features.

\paragraph{Ground-truth fusion.}
We replace the per-view predicted albedo with ground-truth albedo and apply prior fusion (\cref{sec:prior_fusion_details}).
TensoIR-Synthetic supplies training-view albedo ground truth.
Since Synthetic4Relight does not supply training-view albedo ground truth, its oracle uses all 200 test-view albedo maps.
The fused per-vertex albedo defines the vertex oracle.
Vertices without a GT observation retain the optimized albedo.

\paragraph{Lattice fitting.}
Let $\mathbf a_{\mathrm{GT}}$ denote the fused linear albedo and $\mathbf{M}_0$ the initial material logits.
For each channel, we solve
\begin{equation}
 \ML^*=\arg\min_{\ML}\left\|\Sm^\top\ML-
 \left(\operatorname{logit}(\mathbf a_{\mathrm{GT}})-\mathbf{M}_0\right)\right\|^2,
 \label{eq:lattice_oracle_fit}
\end{equation}
using only lattice-supported vertices with a GT observation.
Albedo is clamped to $[10^{-3},1-10^{-3}]$ before taking logits.
The resulting lattice oracle is $\operatorname{sigmoid}(\mathbf{M}_0+\Sm^\top\ML^*)$.
Vertices outside the lattice use the same GT values as the vertex oracle where available, matching the free offsets of our representation.
We use LSQR with absolute and relative tolerances of $10^{-10}$ and at most 20000 iterations per channel.
We additionally fit in linear albedo space and clip the sliced values to $[0,1]$.

\begin{table*}[t]
\centering
\small
\setlength{\tabcolsep}{5pt}
\caption{\textbf{Albedo oracle fits with both prior models.} Four-scene means of 200-view albedo PSNR (dB), using one retained seed for oracle fits and point counts, and three seeds for both optimized representations. Mesh vertex and lattice point counts are scene means; K denotes thousands. Vertex denotes fused GT on the final mesh of ours; linear and logit denote lattice fits in the respective spaces.}
\label{tab:lattice_oracle_full}
\begin{tabular}{ll rr ccccc}
\toprule
Dataset & Prior & Mesh vertices & Lattice points & Vertex & Linear & Logit & Per-vertex opt. & Lattice opt. \\
\midrule
S4R & MVI & 296.5K & 26.4K & 36.33 & 35.40 & 35.23 & 29.73 & 33.29 \\
S4R & DR & 287.5K & 1.7K & 36.38 & 35.26 & 35.12 & 29.91 & 33.08 \\
TIR & MVI & 311.3K & 40.7K & 33.60 & 32.92 & 32.86 & 26.23 & 28.51 \\
TIR & DR & 317.1K & 1.8K & 33.87 & 32.89 & 32.86 & 26.37 & 28.72 \\
\bottomrule
\end{tabular}
\end{table*}

\paragraph{Evaluation and comparison.}
We render all 200 test views and use the paper's scene-level RGB scale alignment and gamma-2.2 albedo PSNR.
\Cref{tab:lattice_oracle} compares DiffusionRenderer runs using the same three optimization seeds for the per-vertex control and ours; each optimized result retains its own jointly optimized geometry.
The two oracle fits are evaluated on the final mesh of ours to isolate the material representation.
\Cref{tab:lattice_oracle_full} reports both predictors and both fitting spaces.
The logit-space oracle loses about 1\,dB relative to the vertex oracle, while the optimized lattice albedo remains 2--4\,dB below the logit-space oracle.
Thus the representational restriction of the lattice accounts for only part of the remaining error in these experiments.

\section{Prior Quality Sweep}
\label{sec:prior_quality_details}
\begin{figure*}[t]
\centering
\includegraphics[width=\textwidth]{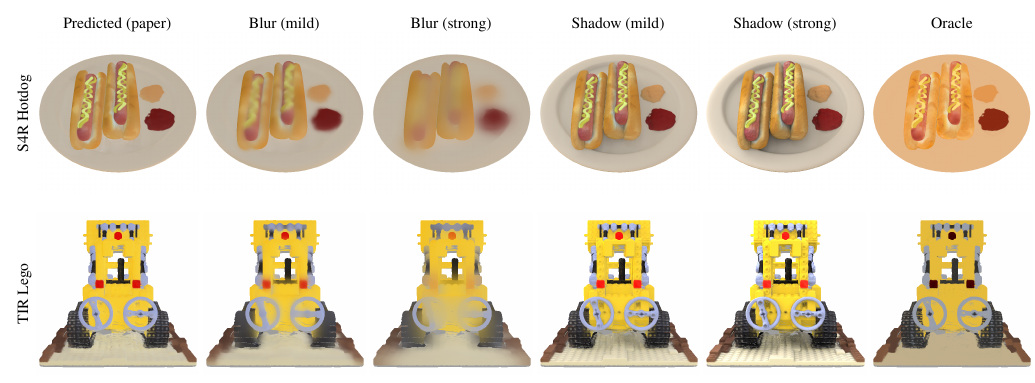}%
\caption{\textbf{Other kinds of degradation.}
Fused prior albedo on Synthetic4Relight Hotdog and TensoIR-Synthetic Lego: the predicted prior used in the paper, blur and shadow baking at the mild and strong levels, and fused ground truth (oracle).
Material confusion is shown in \cref{fig:prior_quality}.
Vertices without a prior are shown in gray.}
\label{fig:prior_alt_gallery}
\end{figure*}

\begin{figure}[t]
\centering
\includegraphics[width=\columnwidth]{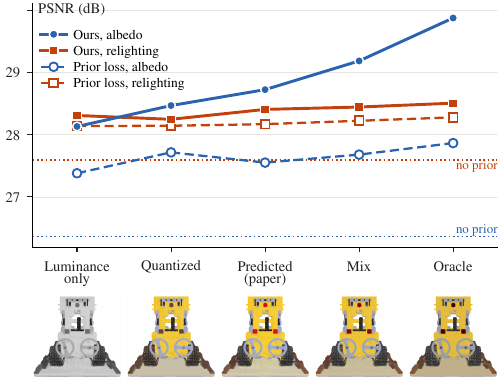}%
\caption{\textbf{Effect of prior quality on TensoIR-Synthetic.}
Material confusion, in the layout of \cref{fig:prior_quality}; prior thumbnails show Lego.
Four-scene means over three seeds.}
\label{fig:prior_quality_tir}
\end{figure}

\begin{table}[t]
\centering
\footnotesize
\setlength{\tabcolsep}{2.4pt}
\caption{\textbf{Other kinds of degradation.} Four-scene means over three seeds with DiffusionRenderer priors. Edge and loss are as in \cref{tab:prior_quality_sweep}; gap is ours minus the prior-loss control.}
\label{tab:prior_alt_degradation}
\begin{tabular}{l c ccc ccc}
\toprule
& & \multicolumn{3}{c}{Albedo PSNR} & \multicolumn{3}{c}{Relighting PSNR} \\
\cmidrule(lr){3-5}\cmidrule(lr){6-8}
Prior & Edge & Ours & Loss & Gap & Ours & Loss & Gap \\
\midrule
\multicolumn{8}{l}{\textit{Synthetic4Relight}} \\
Predicted (paper) & 0.74 & 33.08 & 31.20 & $+$1.88 & 32.26 & 31.92 & $+$0.34 \\
Shadow (mild) & 0.59 & 32.12 & 30.83 & $+$1.29 & 32.08 & 31.78 & $+$0.30 \\
Shadow (strong) & 0.52 & 32.09 & 30.74 & $+$1.35 & 32.06 & 31.81 & $+$0.25 \\
Quantized (8) & 0.53 & 32.81 & 31.44 & $+$1.37 & 32.14 & 31.95 & $+$0.19 \\
Luminance only & 0.74 & 32.33 & 31.34 & $+$0.99 & 32.08 & 31.98 & $+$0.10 \\
No prior & -- & \multicolumn{3}{c}{29.91} & \multicolumn{3}{c}{31.37} \\
\midrule
\multicolumn{8}{l}{\textit{TensoIR-Synthetic}} \\
Predicted (paper) & 0.53 & 28.72 & 27.55 & $+$1.17 & 28.41 & 28.17 & $+$0.24 \\
Shadow (mild) & 0.53 & 27.86 & 27.26 & $+$0.59 & 28.40 & 28.11 & $+$0.29 \\
Shadow (strong) & 0.47 & 27.37 & 27.09 & $+$0.28 & 28.31 & 28.04 & $+$0.27 \\
Quantized (8) & 0.43 & 28.47 & 27.72 & $+$0.75 & 28.25 & 28.14 & $+$0.10 \\
Luminance only & 0.51 & 28.13 & 27.38 & $+$0.75 & 28.31 & 28.14 & $+$0.17 \\
No prior & -- & \multicolumn{3}{c}{26.37} & \multicolumn{3}{c}{27.60} \\
\bottomrule
\end{tabular}
\end{table}

This section details the experiment in \cref{sec:prior_quality}.

\paragraph{Prior variants.}
Each variant changes only the albedo channels of the fused prior features of supported vertices.
The degraded priors are shown in \cref{fig:prior_quality} (material confusion) and \cref{fig:prior_alt_gallery} (blur and shadow baking).
\emph{Material confusion} quantizes the prior albedo of the supported vertices to eight colors with $k$-means in sRGB (mild), or keeps only its linear luminance in all three channels (strong), so that materials differing only in hue become indistinguishable.
\emph{Blur} applies $\mathbf p_v\leftarrow\tfrac12\mathbf p_v+\tfrac12\,\mathrm{mean}_{u\in\mathcal N(v)}\mathbf p_u$ to the sRGB-encoded albedo for 100 (mild) or 1000 (strong) iterations, where $\mathcal N(v)$ contains the supported mesh neighbors of $v$.
\emph{Shadow baking} multiplies the linear prior albedo by $(s_v/\tilde s)^{\gamma}$ with $\gamma=0.5$ (mild) or $\gamma=1$ (strong).
The shading $s_v$ of vertex $v$ is the luminance of the fused input images divided by the luminance of the fused ground-truth albedo, both fused with the prior-fusion procedure, and $\tilde s$ is its median over vertices; the ratio $s_v/\tilde s$ is clamped to $[0.05,4]$.
Cast shadows, occlusion and highlights of the training views are thereby baked into the prior.
\emph{Oracle} fuses ground-truth albedo onto the initial mesh with the prior-fusion procedure, as in \cref{sec:lattice_oracle_details}.
\emph{Mix} averages the predicted and oracle albedo features of each vertex in sRGB.
To summarize how well a prior separates materials, we report the correlation between $\|\mathbf p_u-\mathbf p_v\|$ and the corresponding oracle difference over mesh edges $(u,v)$ with both endpoints supported.
Prior albedo PSNR is not a suitable summary, since it is dominated by a global color offset that the lattice offsets correct; for example, both blurred priors of Synthetic4Relight Chair score within 0.2\,dB of the predicted prior.

\paragraph{Optimization and evaluation.}
All runs use the configuration of \cref{tab:prior_use}, including joint geometry optimization and prior-loss weight 1, and are evaluated on all 200 test views.
All results average the three optimization seeds of \cref{tab:prior_use}; their sample standard deviation across seeds is at most 0.08\,dB.
The predicted-prior entries of ours and of the prior-loss control and the no-prior entries are the runs of \cref{tab:prior_use} with the corresponding seeds.

\paragraph{Material confusion.}
\Cref{tab:prior_alt_degradation} lists both datasets, and \cref{fig:prior_quality_tir} plots TensoIR-Synthetic.
The prior-loss control is almost unaffected: its albedo PSNR changes by at most 0.25\,dB from the predicted prior on both datasets.
Ours responds to the prior, losing up to 0.8\,dB with the confused priors and, on TensoIR-Synthetic, gaining 1.2\,dB with the oracle prior.
Our relighting advantage accordingly shrinks from 0.2--0.3\,dB with the predicted prior to 0.1--0.2\,dB, but stays positive.
Color thus helps our method separate materials, although the edge correlation barely changes when only luminance is kept.
The oracle prior leaves TensoIR-Synthetic 2.99\,dB below its lattice oracle (\cref{tab:lattice_oracle}), so factors other than the prior also limit albedo recovery on this dataset.

\subsection{Other Kinds of Degradation}
\label{sec:prior_alt_degradation}
\begin{figure*}[t]
\centering
\begin{tabular}{@{}c@{\hspace{.02\textwidth}}c@{}}
\includegraphics[width=.47\textwidth]{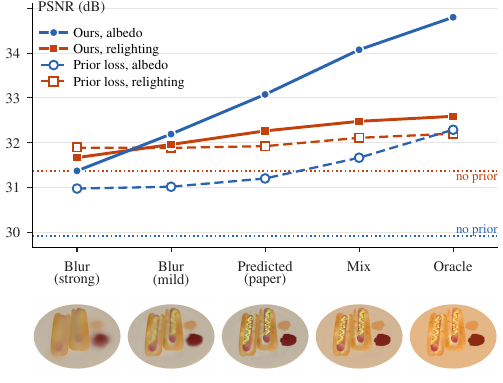} &
\includegraphics[width=.47\textwidth]{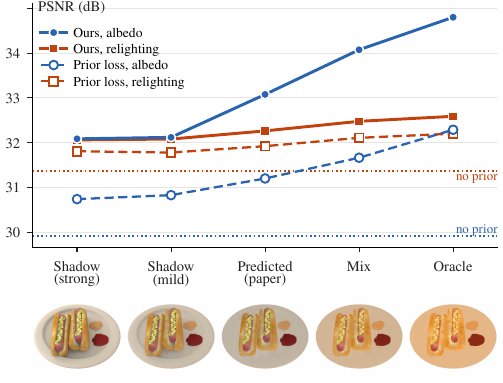} \\[-1pt]
\small (a) Synthetic4Relight, blur & \small (b) Synthetic4Relight, shadow baking \\[4pt]
\includegraphics[width=.47\textwidth]{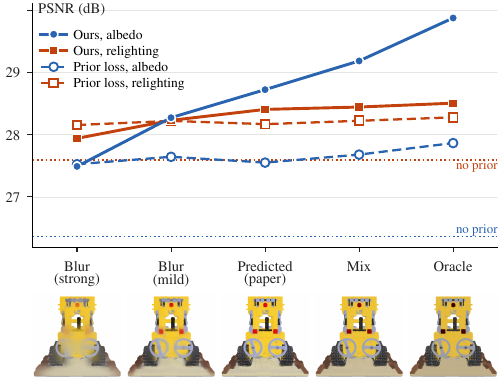} &
\includegraphics[width=.47\textwidth]{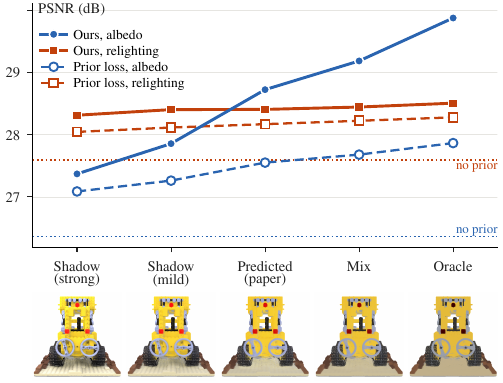} \\[-1pt]
\small (c) TensoIR-Synthetic, blur & \small (d) TensoIR-Synthetic, shadow baking \\
\end{tabular}
\caption{\textbf{Effect of prior quality under other degradations.}
Same layout as \cref{fig:prior_quality}, with the confused priors replaced by blurred priors (left) or baked shadows (right) at the strong and mild levels.
Prior thumbnails show Hotdog and Lego.
Four-scene means over three seeds.}
\label{fig:prior_quality_alt}
\end{figure*}

\begin{table*}[t]
\centering
\footnotesize
\setlength{\tabcolsep}{4pt}
\caption{\textbf{Prior quality sweep.} Four-scene means over three seeds with DiffusionRenderer priors. Only the fused prior albedo differs between rows; lattice spacing, roughness prior, geometry initialization and optimization are fixed. Edge is the correlation between prior and ground-truth albedo differences across mesh edges. Loss is the prior-loss control at weight 1, as in the main paper, and at weight 10. Gap is ours minus the prior loss at weight 1.}
\label{tab:prior_quality_sweep}
\begin{tabular}{l c cccc cccc}
\toprule
& & \multicolumn{4}{c}{Albedo PSNR} & \multicolumn{4}{c}{Relighting PSNR} \\
\cmidrule(lr){3-6}\cmidrule(lr){7-10}
Prior & Edge & Ours & Loss (1) & Loss (10) & Gap & Ours & Loss (1) & Loss (10) & Gap \\
\midrule
\multicolumn{10}{l}{\textit{Synthetic4Relight}} \\
Blur (strong) & 0.11 & 31.37 & 30.98 & 27.68 & $+$0.39 & 31.67 & 31.89 & 30.14 & $-$0.22 \\
Blur (mild) & 0.25 & 32.19 & 31.02 & 28.01 & $+$1.18 & 31.96 & 31.88 & 30.29 & $+$0.08 \\
Predicted (paper) & 0.74 & 33.08 & 31.20 & 28.28 & $+$1.88 & 32.26 & 31.92 & 30.47 & $+$0.34 \\
Mix (50\%) & 0.96 & 34.07 & 31.66 & 28.52 & $+$2.41 & 32.48 & 32.11 & 30.54 & $+$0.37 \\
Oracle & 1.00 & 34.80 & 32.29 & 28.99 & $+$2.51 & 32.59 & 32.20 & 30.63 & $+$0.39 \\
No prior & -- & \multicolumn{4}{c}{29.91} & \multicolumn{4}{c}{31.37} \\
\midrule
\multicolumn{10}{l}{\textit{TensoIR-Synthetic}} \\
Blur (strong) & 0.08 & 27.49 & 27.53 & 25.32 & $-$0.04 & 27.94 & 28.15 & 27.21 & $-$0.21 \\
Blur (mild) & 0.22 & 28.27 & 27.64 & 25.80 & $+$0.63 & 28.23 & 28.22 & 27.43 & $+$0.01 \\
Predicted (paper) & 0.53 & 28.72 & 27.55 & 26.07 & $+$1.17 & 28.41 & 28.17 & 27.54 & $+$0.24 \\
Mix (50\%) & 0.81 & 29.18 & 27.68 & 26.16 & $+$1.50 & 28.44 & 28.22 & 27.56 & $+$0.22 \\
Oracle & 1.00 & 29.87 & 27.87 & 26.31 & $+$2.01 & 28.51 & 28.28 & 27.65 & $+$0.23 \\
No prior & -- & \multicolumn{4}{c}{26.37} & \multicolumn{4}{c}{27.60} \\
\bottomrule
\end{tabular}
\end{table*}

Blur and shadow baking show the same trend as material confusion (\cref{fig:prior_quality_alt}).

\paragraph{Blur.}
From the strongly blurred to the oracle prior, our albedo PSNR rises by 3.4\,dB on Synthetic4Relight and 2.4\,dB on TensoIR-Synthetic, whereas that of the prior-loss control changes by at most 1.3\,dB (\cref{tab:prior_quality_sweep}).
Our relighting advantage likewise shrinks from 0.2--0.4\,dB with the predicted and oracle priors to about $-0.2$\,dB with the strongly blurred prior, where the control is ahead.

\paragraph{Shadow baking.}
Shadow baking mainly lowers albedo PSNR, by 0.9--1.4\,dB for ours and 0.3--0.5\,dB for the prior-loss control (\cref{tab:prior_alt_degradation}).
Our relighting advantage persists at 0.25--0.30\,dB on both datasets.
Shading mostly varies within a material and leaves material boundaries in the prior, so the grouping of vertices changes less than with blur or material confusion.
With material confusion and shadow baking, our method stays above optimization without a prior and above the prior-loss control.

\paragraph{Prior-loss weight.}
\begin{figure*}[t]
\centering
\includegraphics[width=\textwidth]{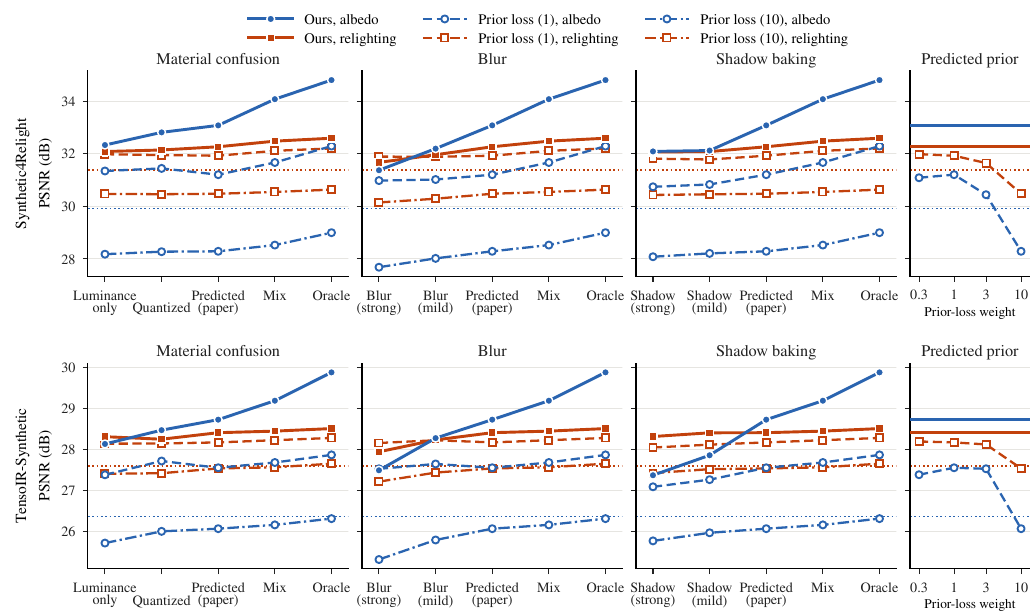}%
\caption{\textbf{Prior-loss weight.}
Albedo and relighting PSNR of our method and of the prior-loss control at weight 1, as in the main paper, and at weight 10, for each degradation (columns) on Synthetic4Relight and TensoIR-Synthetic (rows).
Right: the prior-loss control with the predicted prior at weights 0.3, 1, 3 and 10, with ours indicated as solid lines.
Dotted: optimization without a prior.
Four-scene means over three seeds with DiffusionRenderer priors.}
\label{fig:prior_loss_weight}
\end{figure*}

To test whether a stronger prior loss lets the control exploit a better prior, we repeat all three sweeps with prior-loss weight 10 (\cref{fig:prior_loss_weight}; \cref{tab:prior_quality_sweep} lists the blur sweep).
For every degradation and prior level on both datasets, the stronger loss lowers albedo PSNR below that of optimization without a prior.
Its sensitivity to the prior does not increase: from the most degraded to the oracle prior, its Synthetic4Relight albedo PSNR rises by 0.8, 1.3 and 0.9\,dB for material confusion, blur and shadow baking, whereas ours rises by 2.5, 3.4 and 2.7\,dB.
With the predicted prior, we also try weights 0.3 and 3 (right column of \cref{fig:prior_loss_weight}).
Weight 1 gives the control its highest albedo PSNR on both datasets, and no weight from 0.3 to 10 brings it within 1.1\,dB of our albedo PSNR.
We therefore keep weight 1, following DiffReg-PBIR~\cite{diffusion_material_regularization_eccv}.

\section{Baseline Settings}
\label{sec:supp_baseline_settings}
\paragraph{TensoIR baselines.}
The training and test images of TensoIR-Synthetic do not show the environment, and the object is composited onto a constant black background.
By default, Neural-PBIR and DiffReg-PBIR treat background pixels as direct observations of the environment and use them to supervise the illumination.
On TensoIR-Synthetic, this would fit the environment map to the constant background instead of the true illumination, putting both baselines at a disadvantage.
For a fairer comparison, we therefore run both baselines with the environment hidden from camera rays, and additionally disable Neural-PBIR's background-observation loss during distillation.
The released DiffReg-PBIR configurations specify the normal-guidance weight of surface reconstruction separately for each dataset and include none for TensoIR-Synthetic.
We evaluated the two values used by its released configurations, $10^{-5}$ (Synthetic4Relight) and $10^{-6}$ (Stanford-ORB), and report the better one, $10^{-6}$, which raises its relighting PSNR from 24.67 to 27.01\,dB.
This is also the weight used by our method, so both methods start from the same reconstructed mesh on this dataset.
On the other datasets, DiffReg-PBIR uses its released values.

\paragraph{Geometry.}
All baselines are run with their released code and, where available, their released configurations.
In these, Neural-PBIR and DiffReg-PBIR optimize materials and illumination on the mesh obtained from surface reconstruction without updating vertex positions.
MaterialFusion and our method optimize geometry jointly with materials and illumination.
\Cref{sec:geometry_ablation} reports our method with geometry fixed.

\section{Evaluation Protocol}
\label{sec:eval_protocols}

We use a common evaluation protocol for Synthetic4Relight and TensoIR-Synthetic, adapted from the public TensoIR implementation~\cite{tensoir}.
The main benchmark, prior-use ablation and learned-metallic study score all 200 held-out views at $800\times800$ using each method's native renderer, with the environment hidden from camera rays.
Every synthetic study fits the albedo gauge over all 200 views.

Unless stated otherwise, scores average scenes within each of three optimization seeds, then average seeds. 
Novel-view images are rendered under the learned illumination; relighting uses envmap6 and envmap12 on Synthetic4Relight, and bridge, city, fireplace, forest and night on TensoIR-Synthetic.
Stanford-ORB retains its official benchmark code and target lists throughout.

\paragraph{Shared albedo scale.}
For each reconstructed scene, we compute one RGB scale over the ground-truth foreground of all 200 test views,
\begin{equation}
 s_c=\operatorname{median}_{(i,p)\in\mathcal V_c}
 \frac{a^{\mathrm{GT}}_{i,c}(p)}{\max(\widehat a_{i,c}(p),10^{-6})},
 \quad c\in\{R,G,B\}.
\end{equation}
Here $\mathcal V_c$ contains foreground pixels from all test views with positive ground-truth and predicted channel values ($\widehat a_{i,c}>10^{-6}$).
We apply the same scale to the albedo images and to the material before rendering under each novel illumination.
The scale is fixed across views and lights.
Aligned albedo images are clipped to $[0,1]$ for scoring.
Scaled materials enter each native BRDF without an additional clamp.

\paragraph{Image metrics.}
Novel-view and relit images are evaluated on white backgrounds, retaining each method's predicted coverage.
Aligned albedo is evaluated over the ground-truth foreground composited onto white.
For every image-based PSNR, including albedo, we first compute
$\mathrm{MSE}_i=\sum_{p,c}(I_{i,c}(p)-\widehat I_{i,c}(p))^2/(3HW)$,
then average $-10\log_{10}\mathrm{MSE}_i$ over views.
Relighting also averages the novel illuminations.
Rendered RGB is scored in sRGB.
Following the public TensoIR implementation, albedo PSNR uses a $1/2.2$ power transform of linear albedo, while albedo SSIM is computed in linear space.
SSIM uses the public valid Gaussian filter of size 11 and standard deviation 1.5.

\paragraph{Other metrics and rendering.}
Roughness MAE averages foreground error per view, then views and scenes, over all four scenes of each synthetic dataset.
TensoIR normal angular error uses the public whole-image convention with a $+z$ background.
Our renderer, Neural-PBIR and DiffReg-PBIR use 128 samples per pixel for novel views and 512 for relighting; MaterialFusion averages two passes with $32^2$ light samples per pixel.
Evaluation uses fixed random seeds and leaves optimized checkpoints unchanged.
Opt.\ measures the recorded optimization loop, excluding initialization, final export and evaluation.
For ours, Total combines retained prior-inference and mesh-initialization costs with the refinement and joint-stage times.

\paragraph{DTC-Synthetic.}
We follow the released DiffReg-PBIR relighting protocol~\cite{diffusion_material_regularization_eccv} for all seven scenes.
Each reconstruction is rendered by the released Mitsuba evaluator at 1024 samples per pixel under envmap12, envmap3 and envmap6 from the same 20 orbit cameras.
PSNR, SSIM and VGG-LPIPS are evaluated on ground-truth-masked sRGB images over the full $800\times800$ frame, averaged over views, illuminations and scenes.
We report the unaligned relighting scores with no albedo-derived material rescaling applied, since the released benchmark has no material ground truth.

\section{Fine Details in the Fused Prior}
\label{sec:prior_detail}
\begin{figure}[t]
\centering
\includegraphics[width=\linewidth]{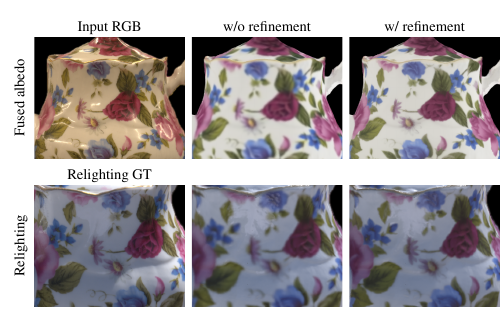}
\vspace{-10pt}
\caption{\textbf{Refining the prior prediction.} Stanford-ORB Teapot. Top: captured RGB and fused albedo before and after refinement. Bottom: ground truth and 1000-step relighting with and without prior refinement.}
\label{fig:prior_fusion_refinement}
\end{figure}

We examine how much fine detail reaches the fused prior that defines the lattice, using the printed labels of Baking and Chips and the patterned Teapot from Stanford-ORB.
\Cref{fig:prior_detail} shows a captured training image together with the per-view albedo, the fused prior and the refined fused prior of MVInverse~\cite{wu2026mvinverse} and DiffusionRenderer~\cite{liang2025diffusionrenderer}, projected onto the same camera.
\Cref{fig:prior_fusion_refinement} shows the fused albedo and the downstream relighting with and without refinement on Teapot from the official light subset, using DiffusionRenderer priors.

Fine lettering visible in the input is already attenuated in the per-view predictions, so detail is lost before fusion, and it remains weak or missing in the unrefined fused prior.
Refinement recovers part of this contrast before joint optimization; on Teapot, it sharpens the flower and leaf boundaries so that the lattice can better separate fine patterns from the surrounding material.
It cannot create contrast where the predictions are locally constant, however, and illumination or geometry edges in the input can still misguide it.
Since the fused features fix the lattice assignments, ink and surrounding regions with similar features share material-update variables, and recovering fine contrast depends on whether the features keep them apart.
This illustrates a limitation of the input prior.

\paragraph{Matched refinement comparison.}
\Cref{tab:refinement} compares the prior-refined runs with unrefined runs at the same initial inputs, configuration and seed.
Refinement changes mean scores only modestly and does not improve every metric, despite sharpening local detail.
The benefits of prior refinement are more evident in qualitative comparisons than in aggregate quantitative metrics.
\begin{table}[t]
\centering
\small
\setlength{\tabcolsep}{5pt}
\caption{\textbf{Matched refinement comparison.} Unrefined- and refined-prior runs, with the same scene inputs, optimization configuration and seed. ORB reports official HDR PSNR.}
\label{tab:refinement}
\begin{tabular}{llccc}
\toprule
Dataset / prior & Features & Albedo & Relighting & NVS \\
\midrule
S4R / MVI & Unrefined & 33.36 & 32.24 & 34.32 \\
S4R / MVI & Refined & 33.21 & 32.25 & 34.28 \\
S4R / DR & Unrefined & 33.28 & 32.37 & 34.32 \\
S4R / DR & Refined & 33.06 & 32.28 & 34.20 \\
TIR / MVI & Unrefined & 28.45 & 28.19 & 31.18 \\
TIR / MVI & Refined & 28.53 & 28.22 & 31.23 \\
TIR / DR & Unrefined & 28.73 & 28.38 & 31.31 \\
TIR / DR & Refined & 28.70 & 28.37 & 31.31 \\
ORB light / MVI & Unrefined & -- & 27.80 & 30.65 \\
ORB light / MVI & Refined & -- & 27.86 & 30.74 \\
ORB light / DR & Unrefined & -- & 27.92 & 30.80 \\
ORB light / DR & Refined & -- & 27.95 & 30.77 \\
\bottomrule
\end{tabular}
\end{table}

\begin{figure}[t]
\centering
\setlength{\tabcolsep}{.4pt}
\renewcommand{\arraystretch}{0}
\begin{tabular}{c cccc}
&&\multicolumn{3}{c}{\fontsize{7}{8}\selectfont DiffusionRenderer}\\[6pt]
&\makebox[0.236\linewidth][c]{\fontsize{6}{7}\selectfont Input RGB}&\makebox[0.236\linewidth][c]{\fontsize{6}{7}\selectfont Per-view}&\makebox[0.236\linewidth][c]{\fontsize{6}{7}\selectfont Fused}&\makebox[0.236\linewidth][c]{\fontsize{6}{7}\selectfont After refinement}\\[1.5pt]
\rotatebox{90}{\makebox[0.415\linewidth][c]{\fontsize{6}{7}\selectfont Baking}}&\includegraphics[width=0.236\linewidth]{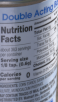}&\includegraphics[width=0.236\linewidth]{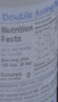}&\includegraphics[width=0.236\linewidth]{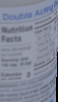}&\includegraphics[width=0.236\linewidth]{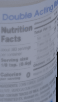}\\[.8pt]
\rotatebox{90}{\makebox[0.117\linewidth][c]{\fontsize{6}{7}\selectfont Chips}}&\includegraphics[width=0.236\linewidth]{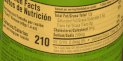}&\includegraphics[width=0.236\linewidth]{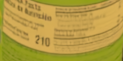}&\includegraphics[width=0.236\linewidth]{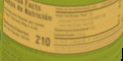}&\includegraphics[width=0.236\linewidth]{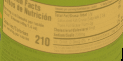}\\[.8pt]
\multicolumn{5}{c}{}\\[3pt]
&&\multicolumn{3}{c}{\fontsize{7}{8}\selectfont MVInverse}\\[6pt]
\rotatebox{90}{\makebox[0.415\linewidth][c]{\fontsize{6}{7}\selectfont Baking}}&&\includegraphics[width=0.236\linewidth]{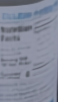}&\includegraphics[width=0.236\linewidth]{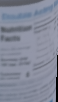}&\includegraphics[width=0.236\linewidth]{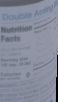}\\[.8pt]
\rotatebox{90}{\makebox[0.117\linewidth][c]{\fontsize{6}{7}\selectfont Chips}}&&\includegraphics[width=0.236\linewidth]{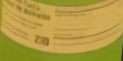}&\includegraphics[width=0.236\linewidth]{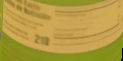}&\includegraphics[width=0.236\linewidth]{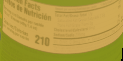}\\[.8pt]
\end{tabular}
\caption{\textbf{Printed detail before joint optimization.} Input RGB and, for each predictor, per-view albedo, unrefined fused albedo and refined fused albedo on two Stanford-ORB labels. Each row uses the same training camera and crop. Fused priors are interpolated on the same initial mesh.}
\label{fig:prior_detail}
\end{figure}

\clearpage
\onecolumn
\section{Additional Qualitative Results}
\label{sec:supp_qual}
\Crefrange{fig:supp_gallery_s4r_air_baloons}{fig:supp_gallery_orb_salt_scene004} show one test view per object: the four Synthetic4Relight and four TensoIR-Synthetic scenes, the seven scenes of the Stanford-ORB light subset, and one capture of each of the seven other Stanford-ORB objects.
Columns show the ground truth, our method with DiffusionRenderer priors and fixed-zero, learned and dual-hypothesis metallic (\cref{sec:metallic}), DiffReg-PBIR, MaterialFusion and Neural-PBIR.
Rows show novel-view synthesis, relighting, albedo, roughness and metallic.
Roughness and metallic are shown linearly from black (0) to white (1).
Our results, MaterialFusion and DiffReg-PBIR on TensoIR-Synthetic use the second of the three optimization seeds.
MaterialFusion was not run on Stanford-ORB captures outside the light subset, and its column is omitted there.
White cells are not available: metallic ground truth, roughness and albedo ground truth on Stanford-ORB, and metallic for Neural-PBIR, which has no metallic parameter.

\raggedbottom\par\vspace{10pt}
\noindent\begin{minipage}{\textwidth}\centering
\hspace{\dimexpr\baselineskip+3pt\relax}\makebox[0.0903\textheight][c]{\scriptsize GT}\makebox[0.0903\textheight][c]{\scriptsize Ours (fixed 0)}\makebox[0.0903\textheight][c]{\scriptsize Ours (learned)}\makebox[0.0903\textheight][c]{\scriptsize Ours (dual)}\makebox[0.0903\textheight][c]{\scriptsize DiffReg-PBIR}\makebox[0.0903\textheight][c]{\scriptsize MaterialFusion}\makebox[0.0903\textheight][c]{\scriptsize Neural-PBIR}\\[1pt]
{\renewcommand{\arraystretch}{0}\setlength{\tabcolsep}{0pt}\begin{tabular}[b]{c}\rotatebox{90}{\makebox[0.0889\textheight][c]{\scriptsize NVS}}\\[0.0014\textheight]\rotatebox{90}{\makebox[0.0889\textheight][c]{\scriptsize Relighting}}\\[0.0014\textheight]\rotatebox{90}{\makebox[0.0889\textheight][c]{\scriptsize Albedo}}\\[0.0014\textheight]\rotatebox{90}{\makebox[0.0889\textheight][c]{\scriptsize Roughness}}\\[0.0014\textheight]\rotatebox{90}{\makebox[0.0889\textheight][c]{\scriptsize Metallic}}\end{tabular}}\hspace{3pt}\includegraphics[height=0.450\textheight]{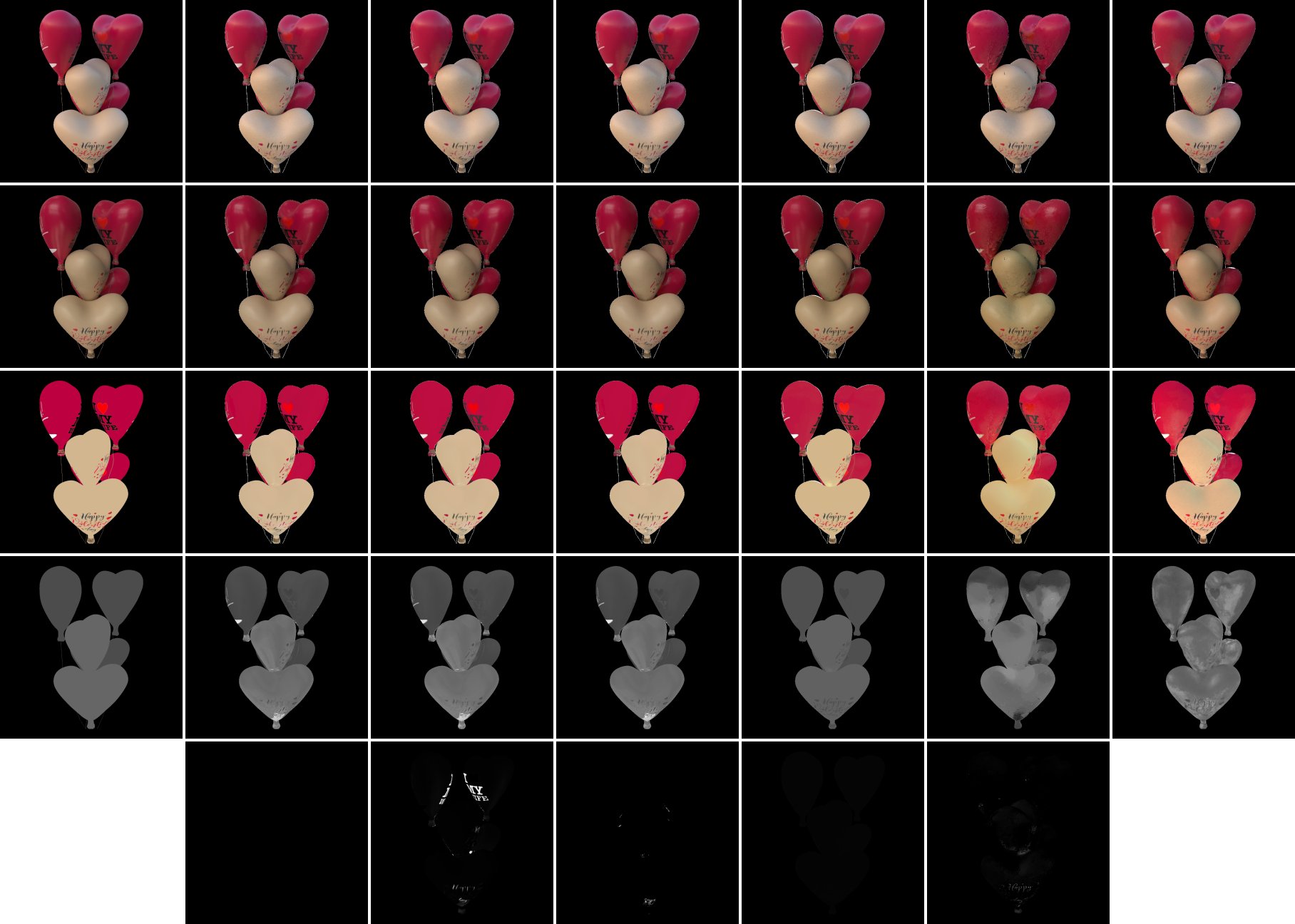}
\captionsetup{hypcap=false}
\captionof{figure}{Synthetic4Relight air\_baloons, test view 80; relighting under envmap6.}
\label{fig:supp_gallery_s4r_air_baloons}
\captionsetup{hypcap=true}
\end{minipage}\par\vspace{8pt}
\noindent\begin{minipage}{\textwidth}\centering
\hspace{\dimexpr\baselineskip+3pt\relax}\makebox[0.0903\textheight][c]{\scriptsize GT}\makebox[0.0903\textheight][c]{\scriptsize Ours (fixed 0)}\makebox[0.0903\textheight][c]{\scriptsize Ours (learned)}\makebox[0.0903\textheight][c]{\scriptsize Ours (dual)}\makebox[0.0903\textheight][c]{\scriptsize DiffReg-PBIR}\makebox[0.0903\textheight][c]{\scriptsize MaterialFusion}\makebox[0.0903\textheight][c]{\scriptsize Neural-PBIR}\\[1pt]
{\renewcommand{\arraystretch}{0}\setlength{\tabcolsep}{0pt}\begin{tabular}[b]{c}\rotatebox{90}{\makebox[0.0889\textheight][c]{\scriptsize NVS}}\\[0.0014\textheight]\rotatebox{90}{\makebox[0.0889\textheight][c]{\scriptsize Relighting}}\\[0.0014\textheight]\rotatebox{90}{\makebox[0.0889\textheight][c]{\scriptsize Albedo}}\\[0.0014\textheight]\rotatebox{90}{\makebox[0.0889\textheight][c]{\scriptsize Roughness}}\\[0.0014\textheight]\rotatebox{90}{\makebox[0.0889\textheight][c]{\scriptsize Metallic}}\end{tabular}}\hspace{3pt}\includegraphics[height=0.450\textheight]{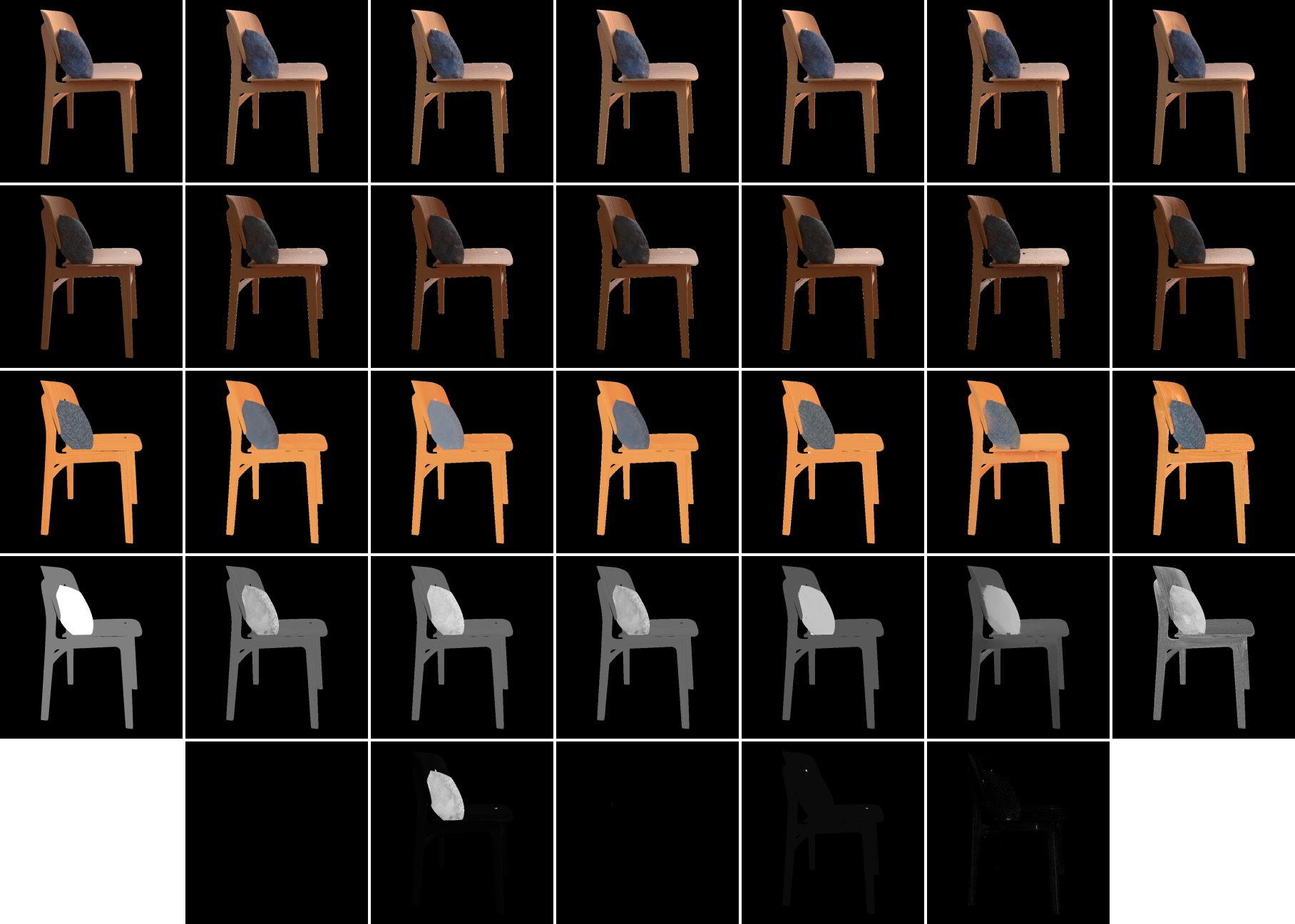}
\captionsetup{hypcap=false}
\captionof{figure}{Synthetic4Relight chair, test view 128; relighting under envmap6.}
\label{fig:supp_gallery_s4r_chair}
\captionsetup{hypcap=true}
\end{minipage}\par\vspace{8pt}
\noindent\begin{minipage}{\textwidth}\centering
\hspace{\dimexpr\baselineskip+3pt\relax}\makebox[0.0903\textheight][c]{\scriptsize GT}\makebox[0.0903\textheight][c]{\scriptsize Ours (fixed 0)}\makebox[0.0903\textheight][c]{\scriptsize Ours (learned)}\makebox[0.0903\textheight][c]{\scriptsize Ours (dual)}\makebox[0.0903\textheight][c]{\scriptsize DiffReg-PBIR}\makebox[0.0903\textheight][c]{\scriptsize MaterialFusion}\makebox[0.0903\textheight][c]{\scriptsize Neural-PBIR}\\[1pt]
{\renewcommand{\arraystretch}{0}\setlength{\tabcolsep}{0pt}\begin{tabular}[b]{c}\rotatebox{90}{\makebox[0.0889\textheight][c]{\scriptsize NVS}}\\[0.0014\textheight]\rotatebox{90}{\makebox[0.0889\textheight][c]{\scriptsize Relighting}}\\[0.0014\textheight]\rotatebox{90}{\makebox[0.0889\textheight][c]{\scriptsize Albedo}}\\[0.0014\textheight]\rotatebox{90}{\makebox[0.0889\textheight][c]{\scriptsize Roughness}}\\[0.0014\textheight]\rotatebox{90}{\makebox[0.0889\textheight][c]{\scriptsize Metallic}}\end{tabular}}\hspace{3pt}\includegraphics[height=0.450\textheight]{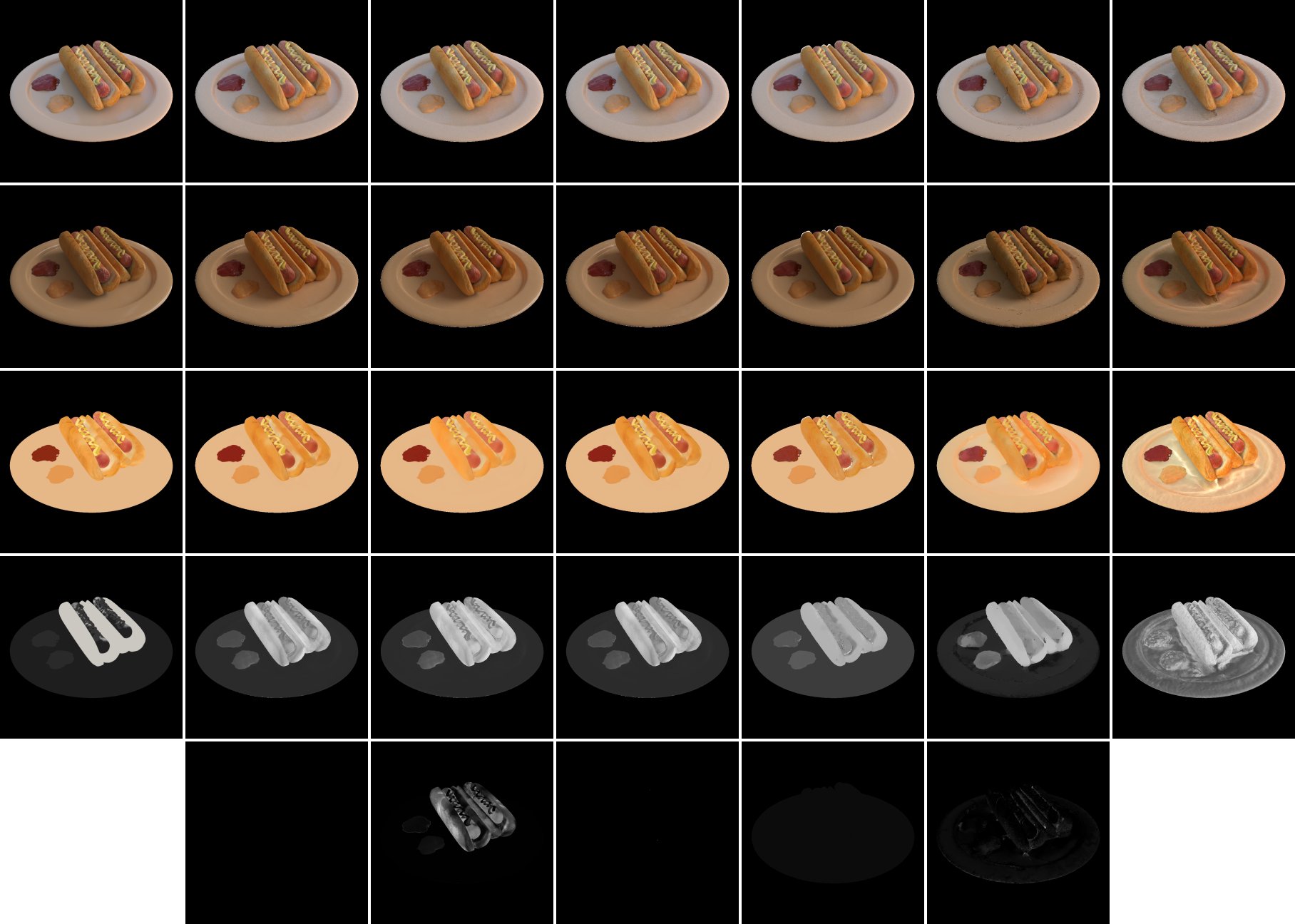}
\captionsetup{hypcap=false}
\captionof{figure}{Synthetic4Relight hotdog, test view 44; relighting under envmap6.}
\label{fig:supp_gallery_s4r_hotdog}
\captionsetup{hypcap=true}
\end{minipage}\par\vspace{8pt}
\noindent\begin{minipage}{\textwidth}\centering
\hspace{\dimexpr\baselineskip+3pt\relax}\makebox[0.0903\textheight][c]{\scriptsize GT}\makebox[0.0903\textheight][c]{\scriptsize Ours (fixed 0)}\makebox[0.0903\textheight][c]{\scriptsize Ours (learned)}\makebox[0.0903\textheight][c]{\scriptsize Ours (dual)}\makebox[0.0903\textheight][c]{\scriptsize DiffReg-PBIR}\makebox[0.0903\textheight][c]{\scriptsize MaterialFusion}\makebox[0.0903\textheight][c]{\scriptsize Neural-PBIR}\\[1pt]
{\renewcommand{\arraystretch}{0}\setlength{\tabcolsep}{0pt}\begin{tabular}[b]{c}\rotatebox{90}{\makebox[0.0889\textheight][c]{\scriptsize NVS}}\\[0.0014\textheight]\rotatebox{90}{\makebox[0.0889\textheight][c]{\scriptsize Relighting}}\\[0.0014\textheight]\rotatebox{90}{\makebox[0.0889\textheight][c]{\scriptsize Albedo}}\\[0.0014\textheight]\rotatebox{90}{\makebox[0.0889\textheight][c]{\scriptsize Roughness}}\\[0.0014\textheight]\rotatebox{90}{\makebox[0.0889\textheight][c]{\scriptsize Metallic}}\end{tabular}}\hspace{3pt}\includegraphics[height=0.450\textheight]{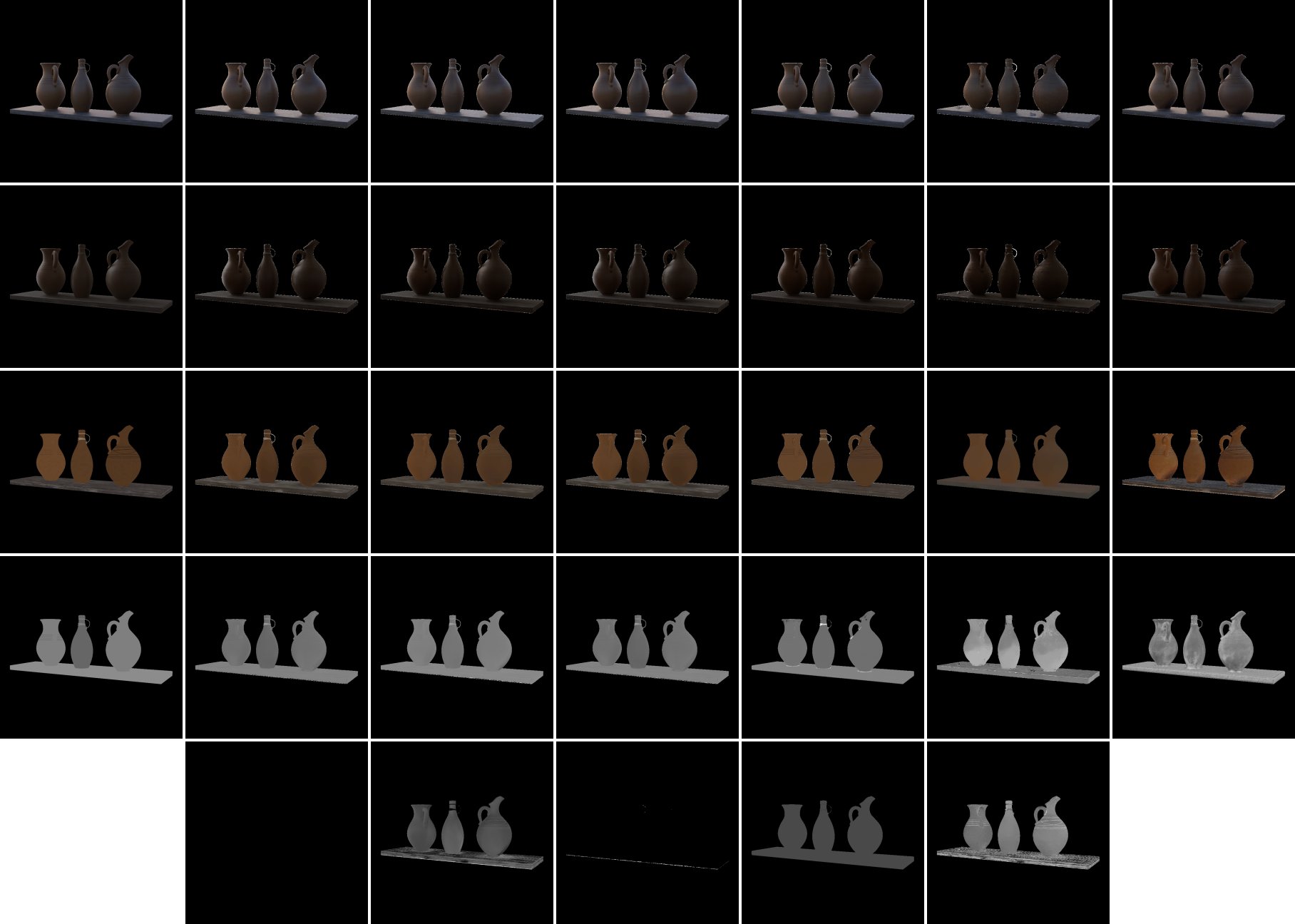}
\captionsetup{hypcap=false}
\captionof{figure}{Synthetic4Relight jugs, test view 108; relighting under envmap6.}
\label{fig:supp_gallery_s4r_jugs}
\captionsetup{hypcap=true}
\end{minipage}\par\vspace{8pt}
\noindent\begin{minipage}{\textwidth}\centering
\hspace{\dimexpr\baselineskip+3pt\relax}\makebox[0.0903\textheight][c]{\scriptsize GT}\makebox[0.0903\textheight][c]{\scriptsize Ours (fixed 0)}\makebox[0.0903\textheight][c]{\scriptsize Ours (learned)}\makebox[0.0903\textheight][c]{\scriptsize Ours (dual)}\makebox[0.0903\textheight][c]{\scriptsize DiffReg-PBIR}\makebox[0.0903\textheight][c]{\scriptsize MaterialFusion}\makebox[0.0903\textheight][c]{\scriptsize Neural-PBIR}\\[1pt]
{\renewcommand{\arraystretch}{0}\setlength{\tabcolsep}{0pt}\begin{tabular}[b]{c}\rotatebox{90}{\makebox[0.0889\textheight][c]{\scriptsize NVS}}\\[0.0014\textheight]\rotatebox{90}{\makebox[0.0889\textheight][c]{\scriptsize Relighting}}\\[0.0014\textheight]\rotatebox{90}{\makebox[0.0889\textheight][c]{\scriptsize Albedo}}\\[0.0014\textheight]\rotatebox{90}{\makebox[0.0889\textheight][c]{\scriptsize Roughness}}\\[0.0014\textheight]\rotatebox{90}{\makebox[0.0889\textheight][c]{\scriptsize Metallic}}\end{tabular}}\hspace{3pt}\includegraphics[height=0.450\textheight]{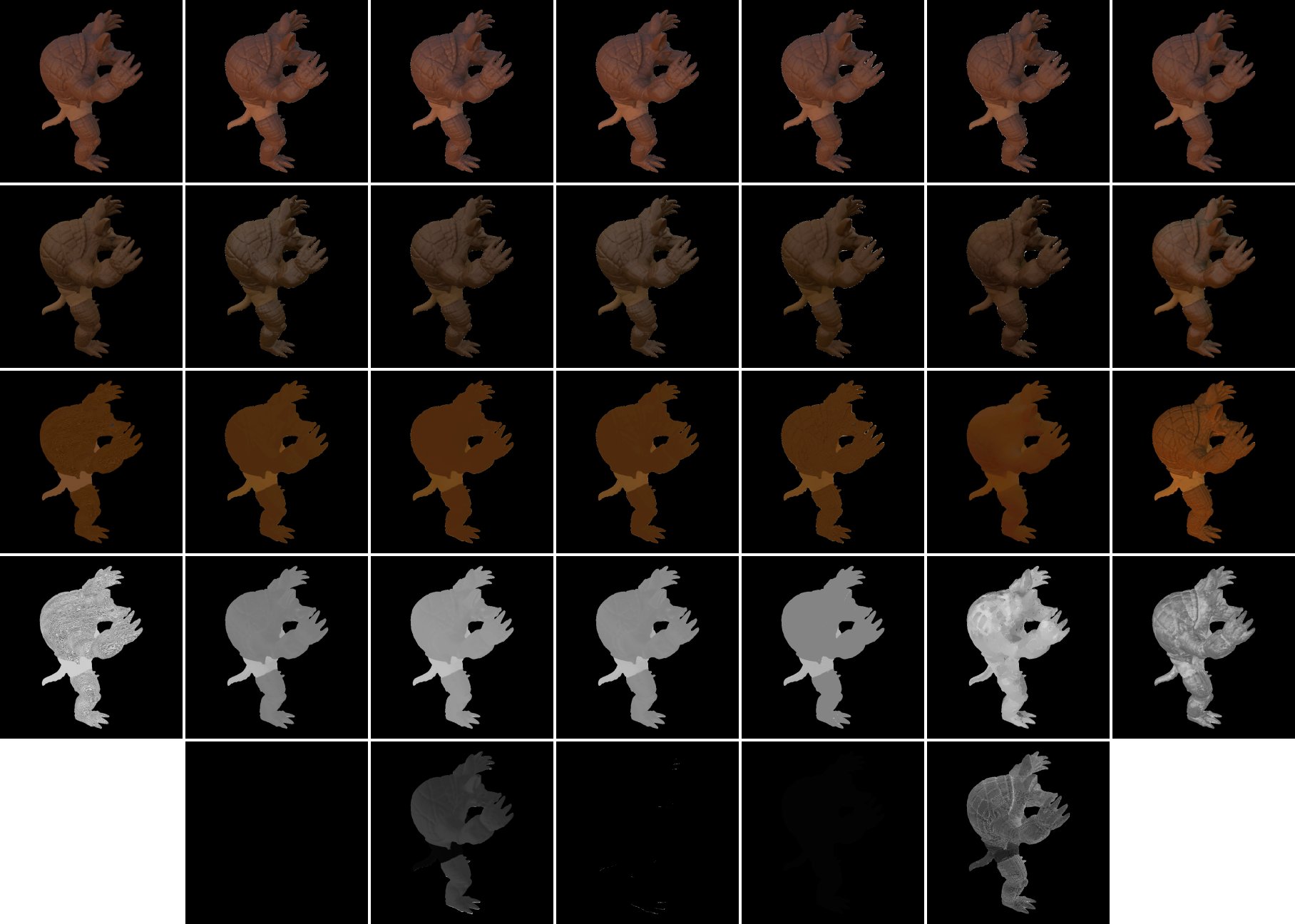}
\captionsetup{hypcap=false}
\captionof{figure}{TensoIR-Synthetic armadillo, test view 172; relighting under forest.}
\label{fig:supp_gallery_tir_armadillo}
\captionsetup{hypcap=true}
\end{minipage}\par\vspace{8pt}
\noindent\begin{minipage}{\textwidth}\centering
\hspace{\dimexpr\baselineskip+3pt\relax}\makebox[0.0903\textheight][c]{\scriptsize GT}\makebox[0.0903\textheight][c]{\scriptsize Ours (fixed 0)}\makebox[0.0903\textheight][c]{\scriptsize Ours (learned)}\makebox[0.0903\textheight][c]{\scriptsize Ours (dual)}\makebox[0.0903\textheight][c]{\scriptsize DiffReg-PBIR}\makebox[0.0903\textheight][c]{\scriptsize MaterialFusion}\makebox[0.0903\textheight][c]{\scriptsize Neural-PBIR}\\[1pt]
{\renewcommand{\arraystretch}{0}\setlength{\tabcolsep}{0pt}\begin{tabular}[b]{c}\rotatebox{90}{\makebox[0.0889\textheight][c]{\scriptsize NVS}}\\[0.0014\textheight]\rotatebox{90}{\makebox[0.0889\textheight][c]{\scriptsize Relighting}}\\[0.0014\textheight]\rotatebox{90}{\makebox[0.0889\textheight][c]{\scriptsize Albedo}}\\[0.0014\textheight]\rotatebox{90}{\makebox[0.0889\textheight][c]{\scriptsize Roughness}}\\[0.0014\textheight]\rotatebox{90}{\makebox[0.0889\textheight][c]{\scriptsize Metallic}}\end{tabular}}\hspace{3pt}\includegraphics[height=0.450\textheight]{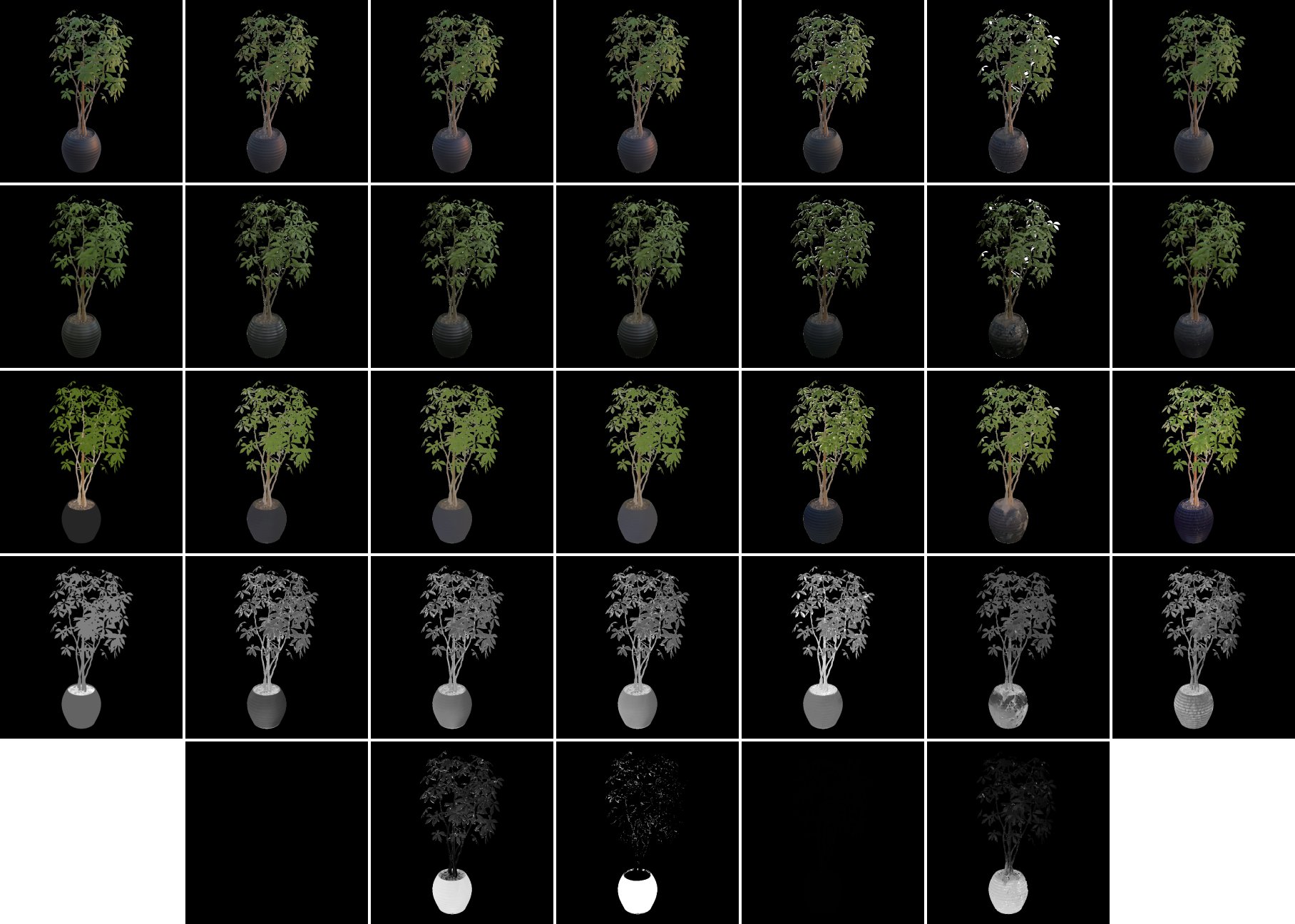}
\captionsetup{hypcap=false}
\captionof{figure}{TensoIR-Synthetic ficus, test view 140; relighting under forest.}
\label{fig:supp_gallery_tir_ficus}
\captionsetup{hypcap=true}
\end{minipage}\par\vspace{8pt}
\noindent\begin{minipage}{\textwidth}\centering
\hspace{\dimexpr\baselineskip+3pt\relax}\makebox[0.0903\textheight][c]{\scriptsize GT}\makebox[0.0903\textheight][c]{\scriptsize Ours (fixed 0)}\makebox[0.0903\textheight][c]{\scriptsize Ours (learned)}\makebox[0.0903\textheight][c]{\scriptsize Ours (dual)}\makebox[0.0903\textheight][c]{\scriptsize DiffReg-PBIR}\makebox[0.0903\textheight][c]{\scriptsize MaterialFusion}\makebox[0.0903\textheight][c]{\scriptsize Neural-PBIR}\\[1pt]
{\renewcommand{\arraystretch}{0}\setlength{\tabcolsep}{0pt}\begin{tabular}[b]{c}\rotatebox{90}{\makebox[0.0889\textheight][c]{\scriptsize NVS}}\\[0.0014\textheight]\rotatebox{90}{\makebox[0.0889\textheight][c]{\scriptsize Relighting}}\\[0.0014\textheight]\rotatebox{90}{\makebox[0.0889\textheight][c]{\scriptsize Albedo}}\\[0.0014\textheight]\rotatebox{90}{\makebox[0.0889\textheight][c]{\scriptsize Roughness}}\\[0.0014\textheight]\rotatebox{90}{\makebox[0.0889\textheight][c]{\scriptsize Metallic}}\end{tabular}}\hspace{3pt}\includegraphics[height=0.450\textheight]{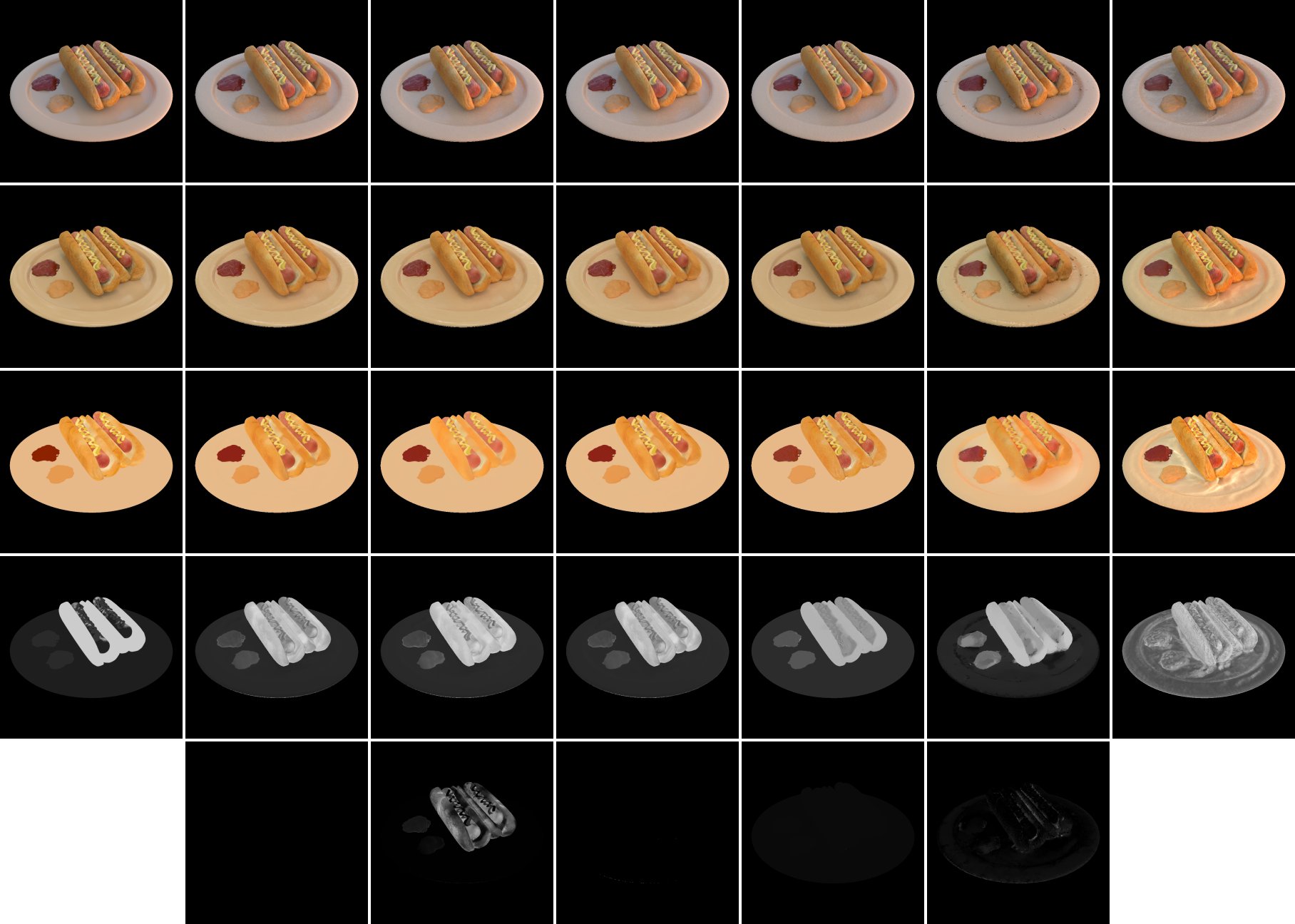}
\captionsetup{hypcap=false}
\captionof{figure}{TensoIR-Synthetic hotdog, test view 44; relighting under forest.}
\label{fig:supp_gallery_tir_hotdog}
\captionsetup{hypcap=true}
\end{minipage}\par\vspace{8pt}
\noindent\begin{minipage}{\textwidth}\centering
\hspace{\dimexpr\baselineskip+3pt\relax}\makebox[0.0903\textheight][c]{\scriptsize GT}\makebox[0.0903\textheight][c]{\scriptsize Ours (fixed 0)}\makebox[0.0903\textheight][c]{\scriptsize Ours (learned)}\makebox[0.0903\textheight][c]{\scriptsize Ours (dual)}\makebox[0.0903\textheight][c]{\scriptsize DiffReg-PBIR}\makebox[0.0903\textheight][c]{\scriptsize MaterialFusion}\makebox[0.0903\textheight][c]{\scriptsize Neural-PBIR}\\[1pt]
{\renewcommand{\arraystretch}{0}\setlength{\tabcolsep}{0pt}\begin{tabular}[b]{c}\rotatebox{90}{\makebox[0.0889\textheight][c]{\scriptsize NVS}}\\[0.0014\textheight]\rotatebox{90}{\makebox[0.0889\textheight][c]{\scriptsize Relighting}}\\[0.0014\textheight]\rotatebox{90}{\makebox[0.0889\textheight][c]{\scriptsize Albedo}}\\[0.0014\textheight]\rotatebox{90}{\makebox[0.0889\textheight][c]{\scriptsize Roughness}}\\[0.0014\textheight]\rotatebox{90}{\makebox[0.0889\textheight][c]{\scriptsize Metallic}}\end{tabular}}\hspace{3pt}\includegraphics[height=0.450\textheight]{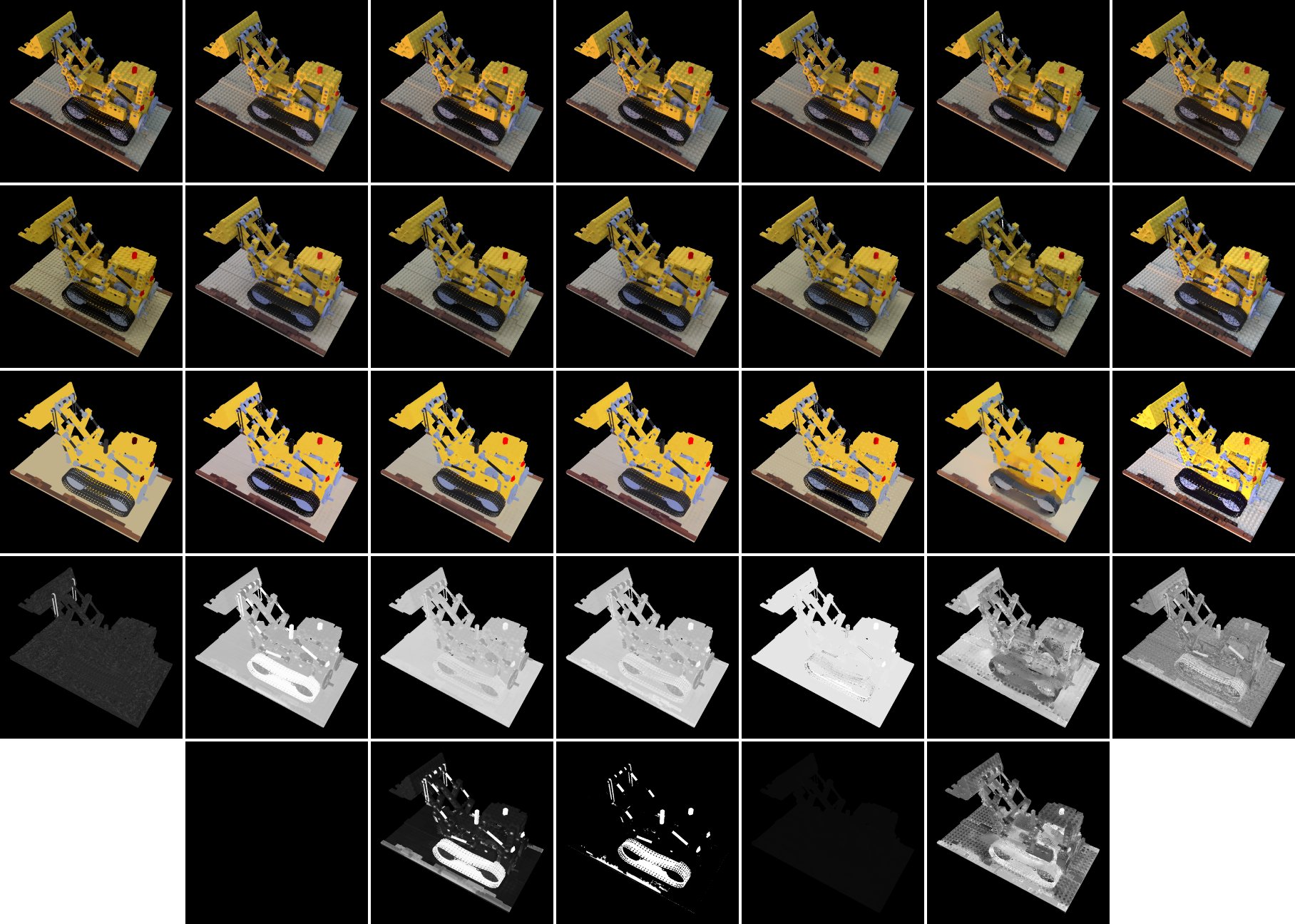}
\captionsetup{hypcap=false}
\captionof{figure}{TensoIR-Synthetic lego, test view 184; relighting under forest.}
\label{fig:supp_gallery_tir_lego}
\captionsetup{hypcap=true}
\end{minipage}\par\vspace{8pt}
\noindent\begin{minipage}{\textwidth}\centering
\hspace{\dimexpr\baselineskip+3pt\relax}\makebox[0.0903\textheight][c]{\scriptsize GT}\makebox[0.0903\textheight][c]{\scriptsize Ours (fixed 0)}\makebox[0.0903\textheight][c]{\scriptsize Ours (learned)}\makebox[0.0903\textheight][c]{\scriptsize Ours (dual)}\makebox[0.0903\textheight][c]{\scriptsize DiffReg-PBIR}\makebox[0.0903\textheight][c]{\scriptsize MaterialFusion}\makebox[0.0903\textheight][c]{\scriptsize Neural-PBIR}\\[1pt]
{\renewcommand{\arraystretch}{0}\setlength{\tabcolsep}{0pt}\begin{tabular}[b]{c}\rotatebox{90}{\makebox[0.0889\textheight][c]{\scriptsize NVS}}\\[0.0014\textheight]\rotatebox{90}{\makebox[0.0889\textheight][c]{\scriptsize Relighting}}\\[0.0014\textheight]\rotatebox{90}{\makebox[0.0889\textheight][c]{\scriptsize Albedo}}\\[0.0014\textheight]\rotatebox{90}{\makebox[0.0889\textheight][c]{\scriptsize Roughness}}\\[0.0014\textheight]\rotatebox{90}{\makebox[0.0889\textheight][c]{\scriptsize Metallic}}\end{tabular}}\hspace{3pt}\includegraphics[height=0.450\textheight]{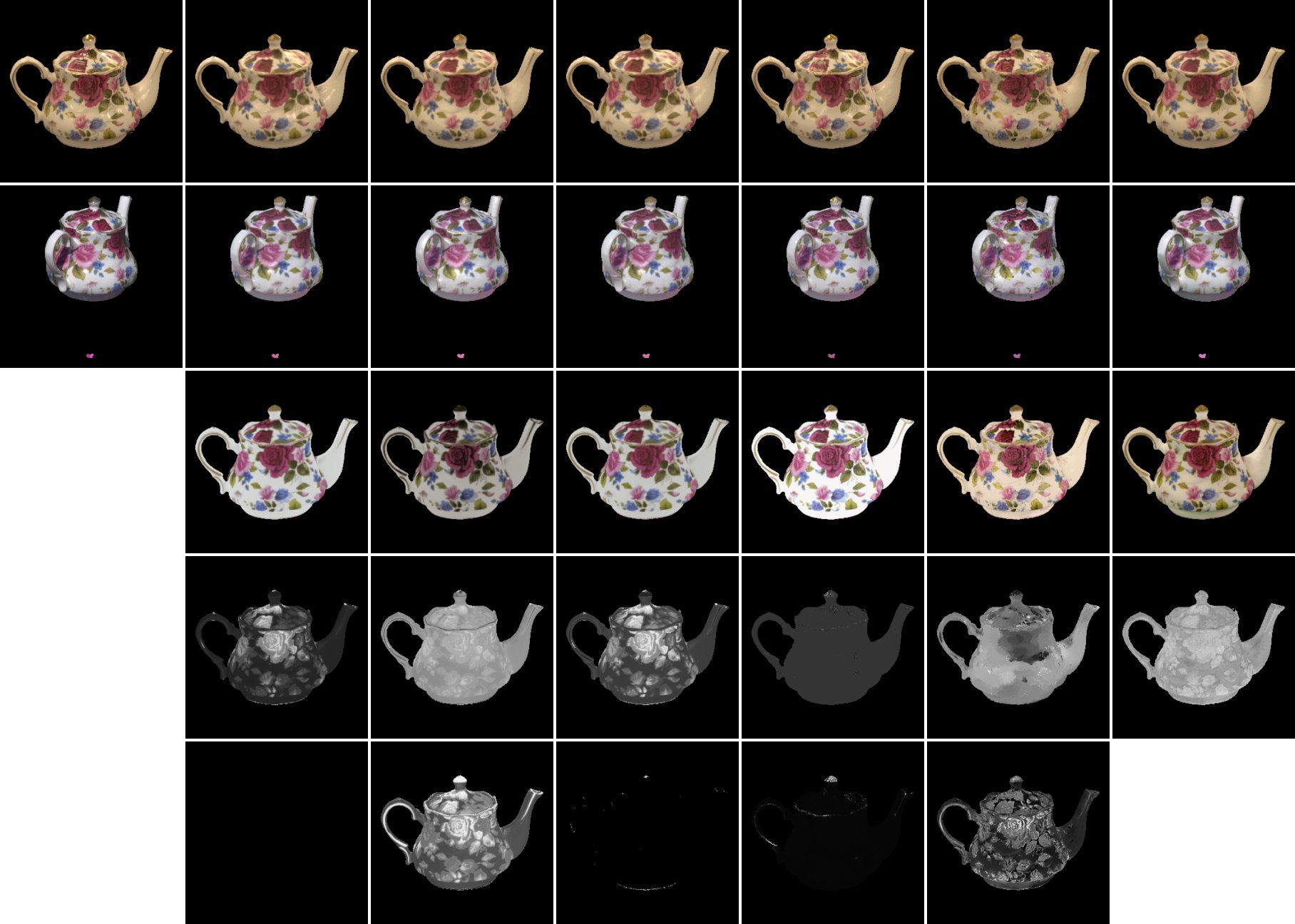}
\captionsetup{hypcap=false}
\captionof{figure}{Stanford-ORB teapot\_scene001 (light subset), camera 0063; relighting under the illumination of teapot\_scene002.}
\label{fig:supp_gallery_orb_teapot_scene001}
\captionsetup{hypcap=true}
\end{minipage}\par\vspace{8pt}
\noindent\begin{minipage}{\textwidth}\centering
\hspace{\dimexpr\baselineskip+3pt\relax}\makebox[0.0903\textheight][c]{\scriptsize GT}\makebox[0.0903\textheight][c]{\scriptsize Ours (fixed 0)}\makebox[0.0903\textheight][c]{\scriptsize Ours (learned)}\makebox[0.0903\textheight][c]{\scriptsize Ours (dual)}\makebox[0.0903\textheight][c]{\scriptsize DiffReg-PBIR}\makebox[0.0903\textheight][c]{\scriptsize MaterialFusion}\makebox[0.0903\textheight][c]{\scriptsize Neural-PBIR}\\[1pt]
{\renewcommand{\arraystretch}{0}\setlength{\tabcolsep}{0pt}\begin{tabular}[b]{c}\rotatebox{90}{\makebox[0.0889\textheight][c]{\scriptsize NVS}}\\[0.0014\textheight]\rotatebox{90}{\makebox[0.0889\textheight][c]{\scriptsize Relighting}}\\[0.0014\textheight]\rotatebox{90}{\makebox[0.0889\textheight][c]{\scriptsize Albedo}}\\[0.0014\textheight]\rotatebox{90}{\makebox[0.0889\textheight][c]{\scriptsize Roughness}}\\[0.0014\textheight]\rotatebox{90}{\makebox[0.0889\textheight][c]{\scriptsize Metallic}}\end{tabular}}\hspace{3pt}\includegraphics[height=0.450\textheight]{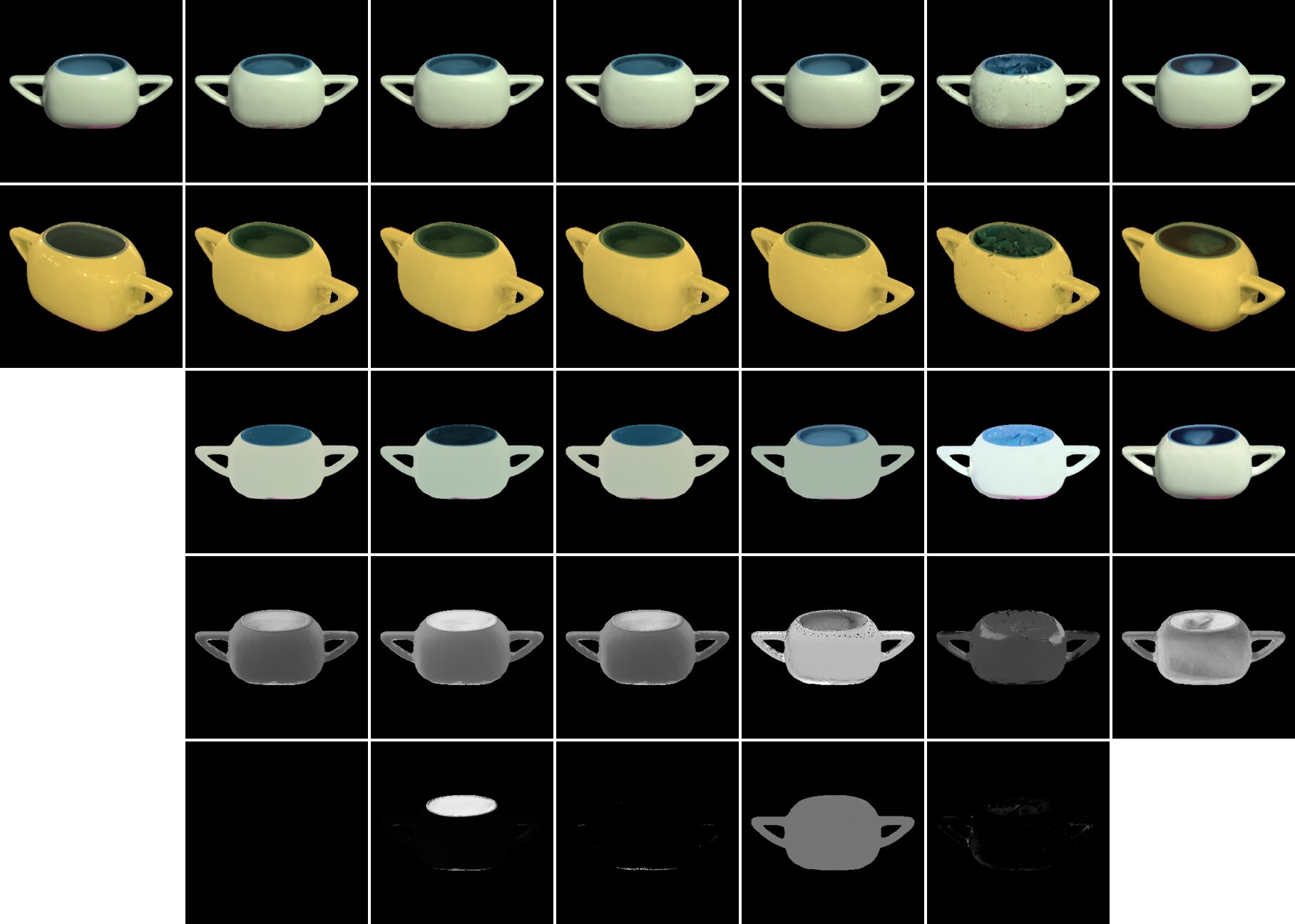}
\captionsetup{hypcap=false}
\captionof{figure}{Stanford-ORB grogu\_scene002 (light subset), camera 0060; relighting under the illumination of grogu\_scene003.}
\label{fig:supp_gallery_orb_grogu_scene002}
\captionsetup{hypcap=true}
\end{minipage}\par\vspace{8pt}
\noindent\begin{minipage}{\textwidth}\centering
\hspace{\dimexpr\baselineskip+3pt\relax}\makebox[0.0903\textheight][c]{\scriptsize GT}\makebox[0.0903\textheight][c]{\scriptsize Ours (fixed 0)}\makebox[0.0903\textheight][c]{\scriptsize Ours (learned)}\makebox[0.0903\textheight][c]{\scriptsize Ours (dual)}\makebox[0.0903\textheight][c]{\scriptsize DiffReg-PBIR}\makebox[0.0903\textheight][c]{\scriptsize MaterialFusion}\makebox[0.0903\textheight][c]{\scriptsize Neural-PBIR}\\[1pt]
{\renewcommand{\arraystretch}{0}\setlength{\tabcolsep}{0pt}\begin{tabular}[b]{c}\rotatebox{90}{\makebox[0.0889\textheight][c]{\scriptsize NVS}}\\[0.0014\textheight]\rotatebox{90}{\makebox[0.0889\textheight][c]{\scriptsize Relighting}}\\[0.0014\textheight]\rotatebox{90}{\makebox[0.0889\textheight][c]{\scriptsize Albedo}}\\[0.0014\textheight]\rotatebox{90}{\makebox[0.0889\textheight][c]{\scriptsize Roughness}}\\[0.0014\textheight]\rotatebox{90}{\makebox[0.0889\textheight][c]{\scriptsize Metallic}}\end{tabular}}\hspace{3pt}\includegraphics[height=0.450\textheight]{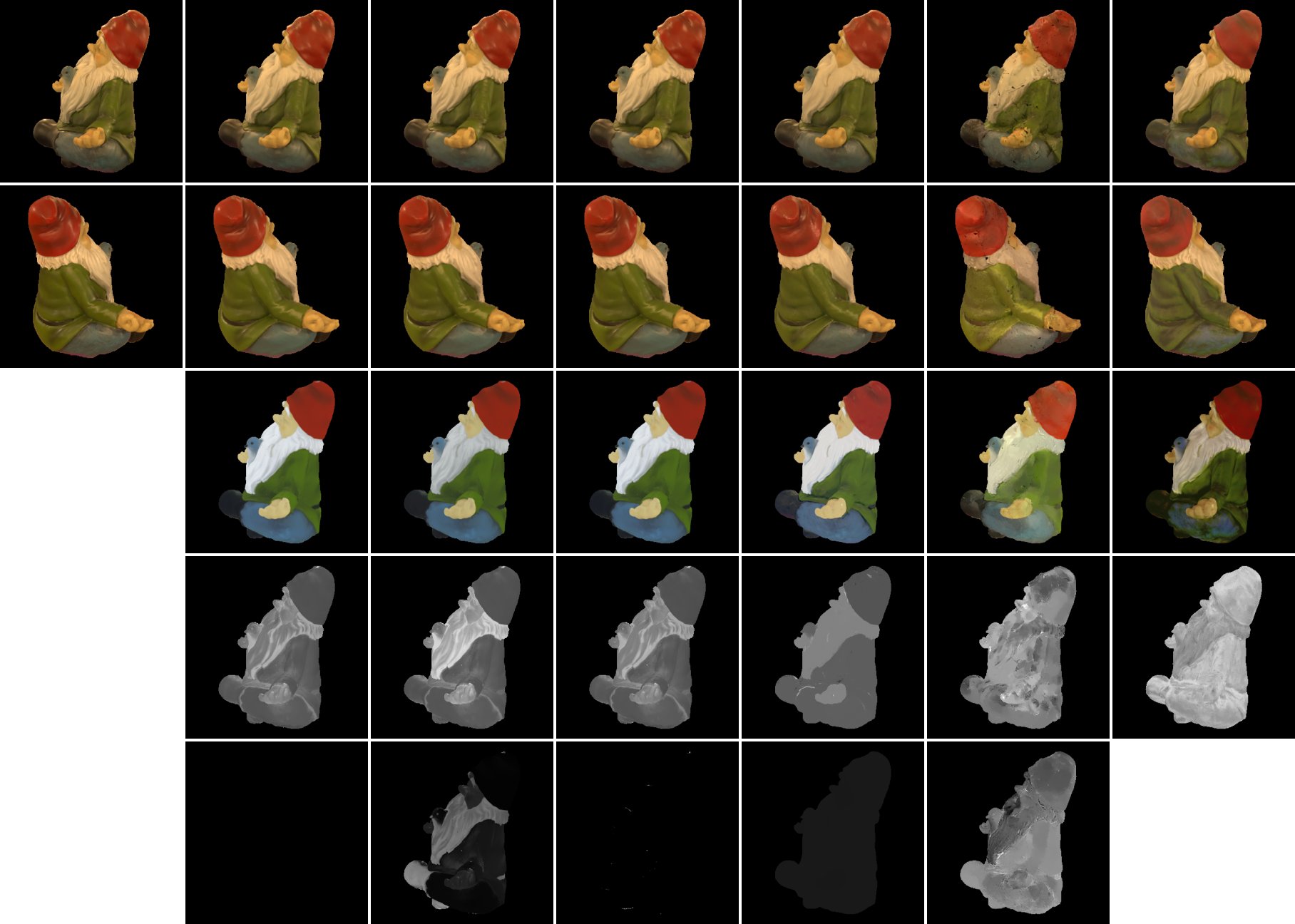}
\captionsetup{hypcap=false}
\captionof{figure}{Stanford-ORB gnome\_scene003 (light subset), camera 0061; relighting under the illumination of gnome\_scene005.}
\label{fig:supp_gallery_orb_gnome_scene003}
\captionsetup{hypcap=true}
\end{minipage}\par\vspace{8pt}
\noindent\begin{minipage}{\textwidth}\centering
\hspace{\dimexpr\baselineskip+3pt\relax}\makebox[0.0903\textheight][c]{\scriptsize GT}\makebox[0.0903\textheight][c]{\scriptsize Ours (fixed 0)}\makebox[0.0903\textheight][c]{\scriptsize Ours (learned)}\makebox[0.0903\textheight][c]{\scriptsize Ours (dual)}\makebox[0.0903\textheight][c]{\scriptsize DiffReg-PBIR}\makebox[0.0903\textheight][c]{\scriptsize MaterialFusion}\makebox[0.0903\textheight][c]{\scriptsize Neural-PBIR}\\[1pt]
{\renewcommand{\arraystretch}{0}\setlength{\tabcolsep}{0pt}\begin{tabular}[b]{c}\rotatebox{90}{\makebox[0.0889\textheight][c]{\scriptsize NVS}}\\[0.0014\textheight]\rotatebox{90}{\makebox[0.0889\textheight][c]{\scriptsize Relighting}}\\[0.0014\textheight]\rotatebox{90}{\makebox[0.0889\textheight][c]{\scriptsize Albedo}}\\[0.0014\textheight]\rotatebox{90}{\makebox[0.0889\textheight][c]{\scriptsize Roughness}}\\[0.0014\textheight]\rotatebox{90}{\makebox[0.0889\textheight][c]{\scriptsize Metallic}}\end{tabular}}\hspace{3pt}\includegraphics[height=0.450\textheight]{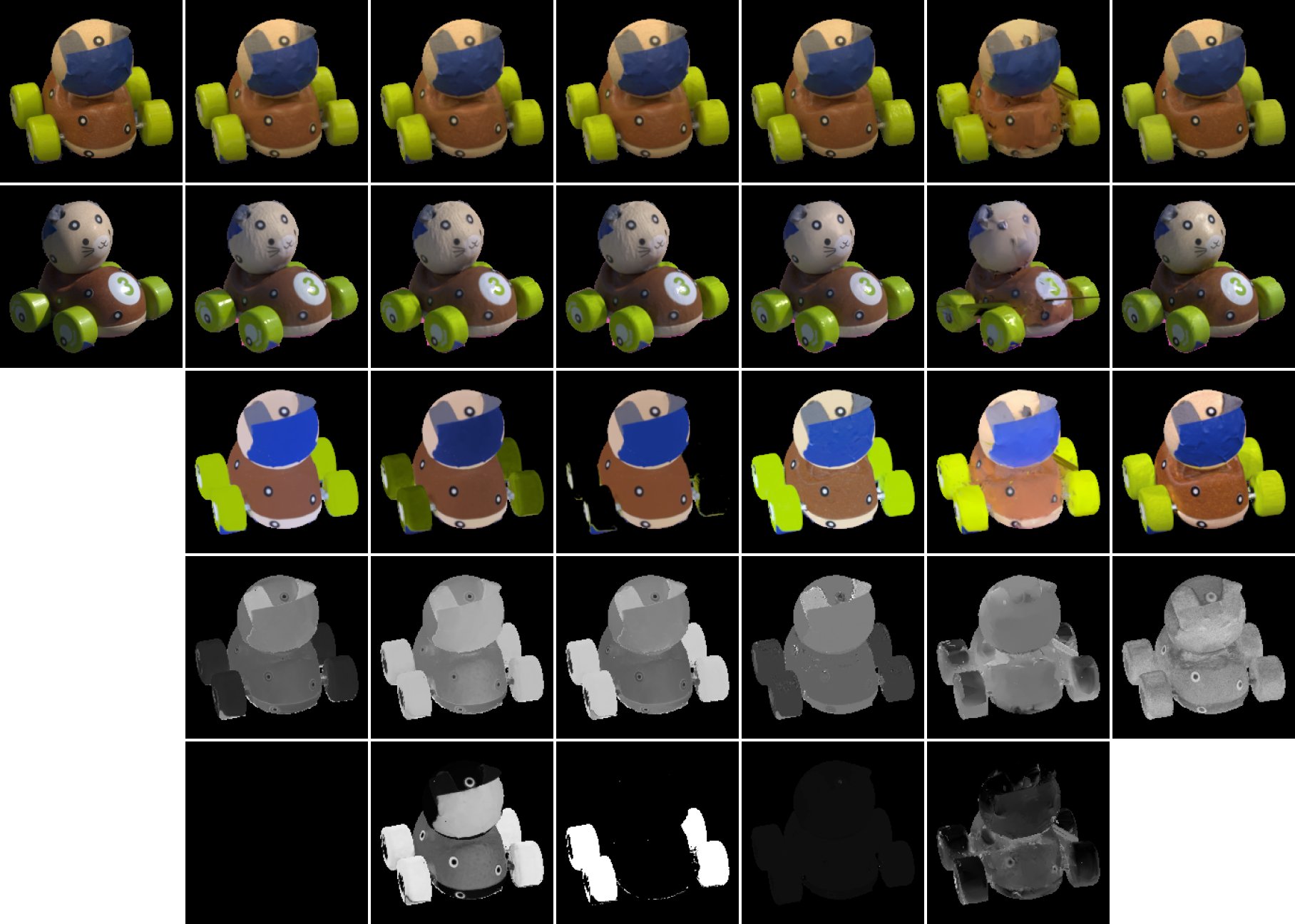}
\captionsetup{hypcap=false}
\captionof{figure}{Stanford-ORB car\_scene004 (light subset), camera 0064; relighting under the illumination of car\_scene002.}
\label{fig:supp_gallery_orb_car_scene004}
\captionsetup{hypcap=true}
\end{minipage}\par\vspace{8pt}
\noindent\begin{minipage}{\textwidth}\centering
\hspace{\dimexpr\baselineskip+3pt\relax}\makebox[0.0903\textheight][c]{\scriptsize GT}\makebox[0.0903\textheight][c]{\scriptsize Ours (fixed 0)}\makebox[0.0903\textheight][c]{\scriptsize Ours (learned)}\makebox[0.0903\textheight][c]{\scriptsize Ours (dual)}\makebox[0.0903\textheight][c]{\scriptsize DiffReg-PBIR}\makebox[0.0903\textheight][c]{\scriptsize MaterialFusion}\makebox[0.0903\textheight][c]{\scriptsize Neural-PBIR}\\[1pt]
{\renewcommand{\arraystretch}{0}\setlength{\tabcolsep}{0pt}\begin{tabular}[b]{c}\rotatebox{90}{\makebox[0.0889\textheight][c]{\scriptsize NVS}}\\[0.0014\textheight]\rotatebox{90}{\makebox[0.0889\textheight][c]{\scriptsize Relighting}}\\[0.0014\textheight]\rotatebox{90}{\makebox[0.0889\textheight][c]{\scriptsize Albedo}}\\[0.0014\textheight]\rotatebox{90}{\makebox[0.0889\textheight][c]{\scriptsize Roughness}}\\[0.0014\textheight]\rotatebox{90}{\makebox[0.0889\textheight][c]{\scriptsize Metallic}}\end{tabular}}\hspace{3pt}\includegraphics[height=0.450\textheight]{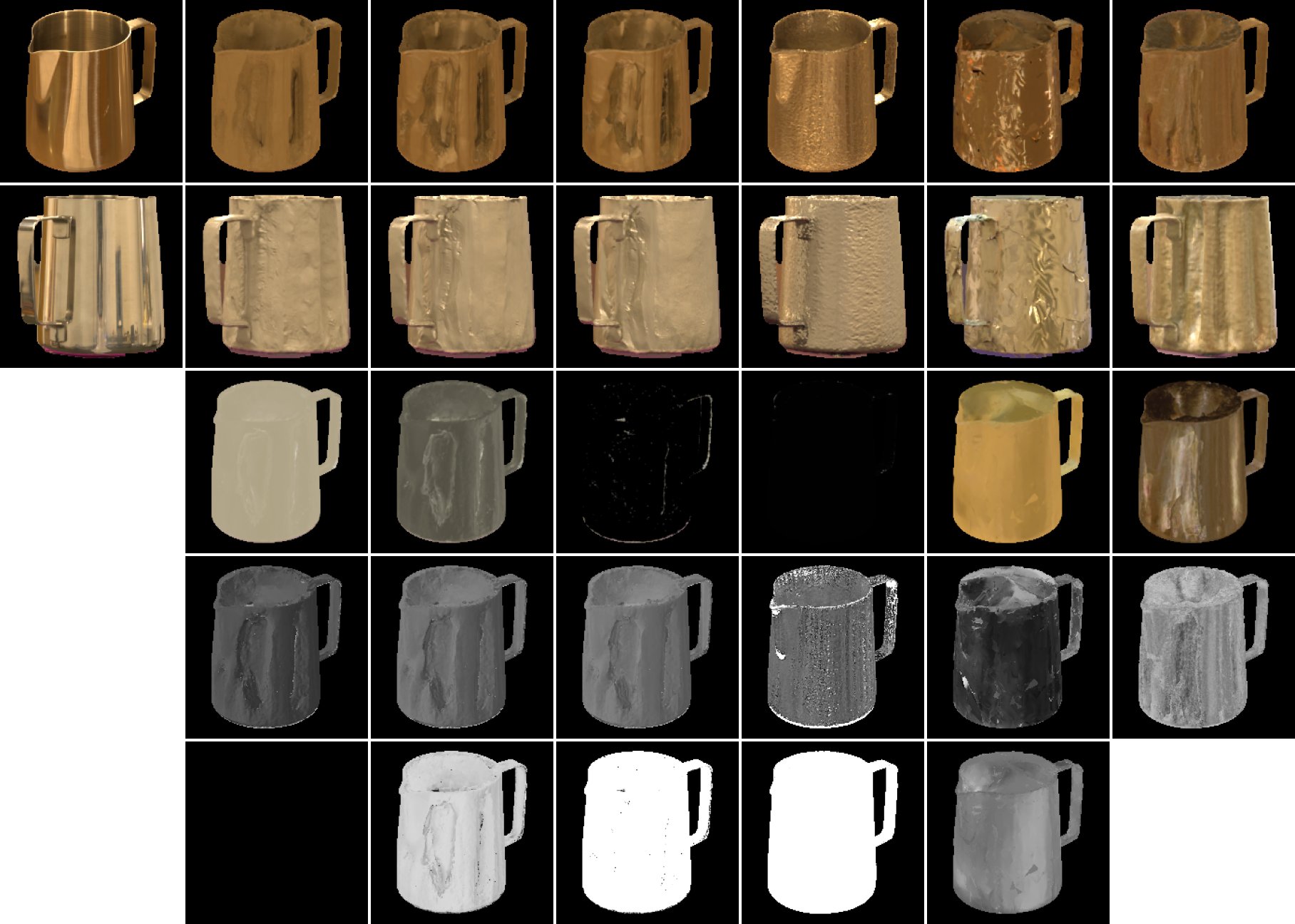}
\captionsetup{hypcap=false}
\captionof{figure}{Stanford-ORB pitcher\_scene005 (light subset), camera 0061; relighting under the illumination of pitcher\_scene001.}
\label{fig:supp_gallery_orb_pitcher_scene005}
\captionsetup{hypcap=true}
\end{minipage}\par\vspace{8pt}
\noindent\begin{minipage}{\textwidth}\centering
\hspace{\dimexpr\baselineskip+3pt\relax}\makebox[0.0903\textheight][c]{\scriptsize GT}\makebox[0.0903\textheight][c]{\scriptsize Ours (fixed 0)}\makebox[0.0903\textheight][c]{\scriptsize Ours (learned)}\makebox[0.0903\textheight][c]{\scriptsize Ours (dual)}\makebox[0.0903\textheight][c]{\scriptsize DiffReg-PBIR}\makebox[0.0903\textheight][c]{\scriptsize MaterialFusion}\makebox[0.0903\textheight][c]{\scriptsize Neural-PBIR}\\[1pt]
{\renewcommand{\arraystretch}{0}\setlength{\tabcolsep}{0pt}\begin{tabular}[b]{c}\rotatebox{90}{\makebox[0.0889\textheight][c]{\scriptsize NVS}}\\[0.0014\textheight]\rotatebox{90}{\makebox[0.0889\textheight][c]{\scriptsize Relighting}}\\[0.0014\textheight]\rotatebox{90}{\makebox[0.0889\textheight][c]{\scriptsize Albedo}}\\[0.0014\textheight]\rotatebox{90}{\makebox[0.0889\textheight][c]{\scriptsize Roughness}}\\[0.0014\textheight]\rotatebox{90}{\makebox[0.0889\textheight][c]{\scriptsize Metallic}}\end{tabular}}\hspace{3pt}\includegraphics[height=0.450\textheight]{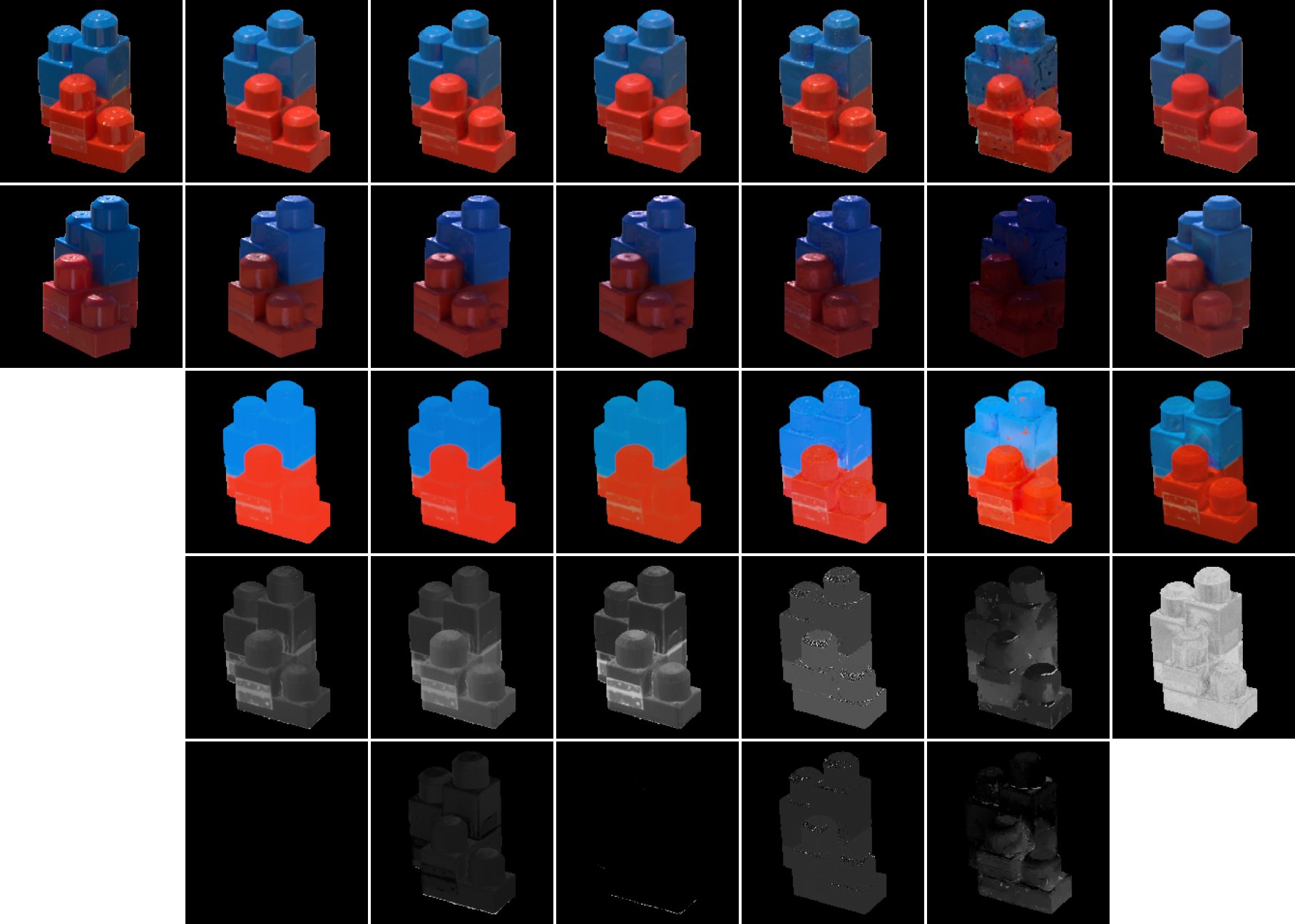}
\captionsetup{hypcap=false}
\captionof{figure}{Stanford-ORB blocks\_scene006 (light subset), camera 0065; relighting under the illumination of blocks\_scene002.}
\label{fig:supp_gallery_orb_blocks_scene006}
\captionsetup{hypcap=true}
\end{minipage}\par\vspace{8pt}
\noindent\begin{minipage}{\textwidth}\centering
\hspace{\dimexpr\baselineskip+3pt\relax}\makebox[0.0903\textheight][c]{\scriptsize GT}\makebox[0.0903\textheight][c]{\scriptsize Ours (fixed 0)}\makebox[0.0903\textheight][c]{\scriptsize Ours (learned)}\makebox[0.0903\textheight][c]{\scriptsize Ours (dual)}\makebox[0.0903\textheight][c]{\scriptsize DiffReg-PBIR}\makebox[0.0903\textheight][c]{\scriptsize MaterialFusion}\makebox[0.0903\textheight][c]{\scriptsize Neural-PBIR}\\[1pt]
{\renewcommand{\arraystretch}{0}\setlength{\tabcolsep}{0pt}\begin{tabular}[b]{c}\rotatebox{90}{\makebox[0.0889\textheight][c]{\scriptsize NVS}}\\[0.0014\textheight]\rotatebox{90}{\makebox[0.0889\textheight][c]{\scriptsize Relighting}}\\[0.0014\textheight]\rotatebox{90}{\makebox[0.0889\textheight][c]{\scriptsize Albedo}}\\[0.0014\textheight]\rotatebox{90}{\makebox[0.0889\textheight][c]{\scriptsize Roughness}}\\[0.0014\textheight]\rotatebox{90}{\makebox[0.0889\textheight][c]{\scriptsize Metallic}}\end{tabular}}\hspace{3pt}\includegraphics[height=0.450\textheight]{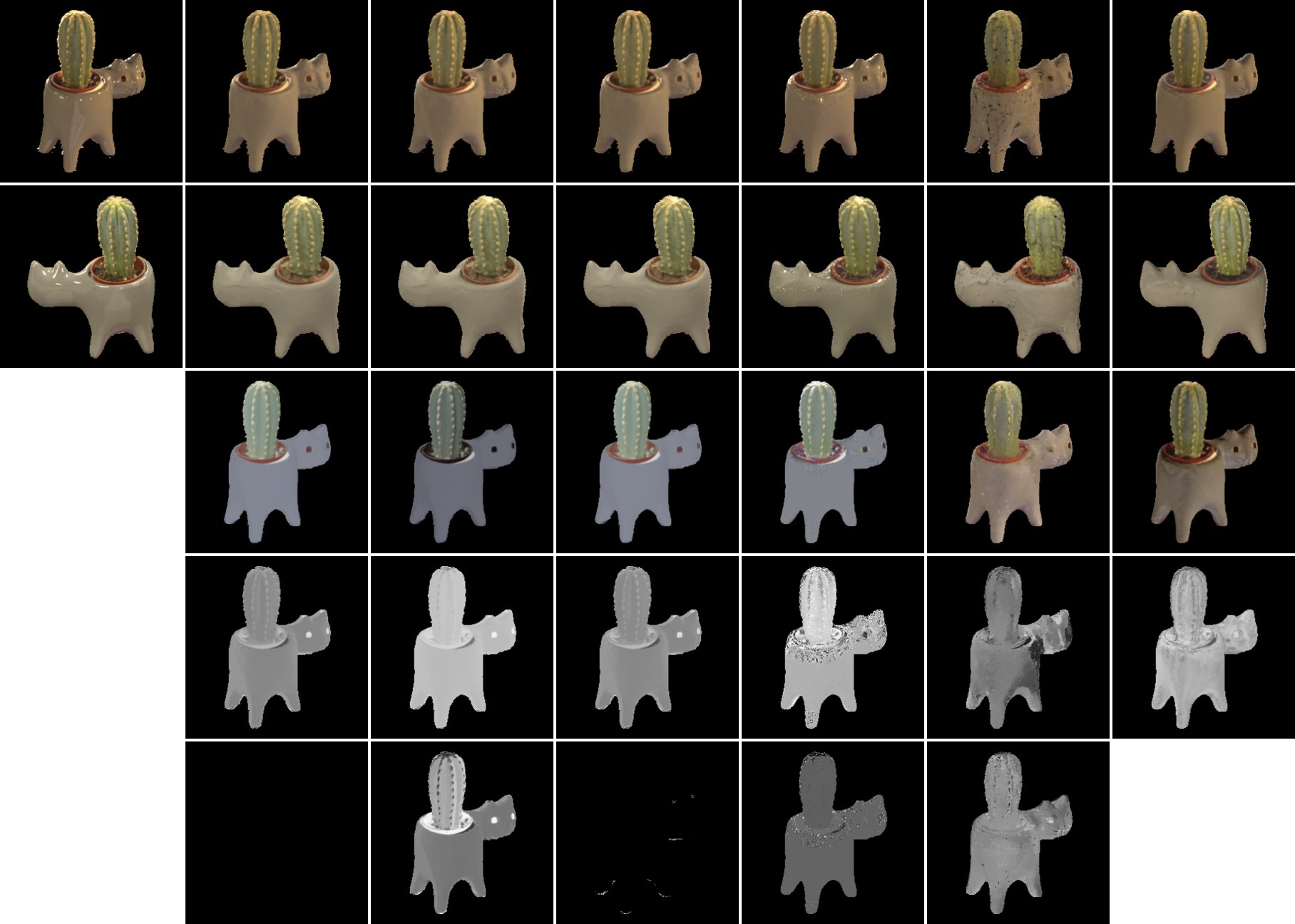}
\captionsetup{hypcap=false}
\captionof{figure}{Stanford-ORB cactus\_scene007 (light subset), camera 0062; relighting under the illumination of cactus\_scene001.}
\label{fig:supp_gallery_orb_cactus_scene007}
\captionsetup{hypcap=true}
\end{minipage}\par\vspace{8pt}
\noindent\begin{minipage}{\textwidth}\centering
\hspace{\dimexpr\baselineskip+3pt\relax}\makebox[0.0903\textheight][c]{\scriptsize GT}\makebox[0.0903\textheight][c]{\scriptsize Ours (fixed 0)}\makebox[0.0903\textheight][c]{\scriptsize Ours (learned)}\makebox[0.0903\textheight][c]{\scriptsize Ours (dual)}\makebox[0.0903\textheight][c]{\scriptsize DiffReg-PBIR}\makebox[0.0903\textheight][c]{\scriptsize Neural-PBIR}\\[1pt]
{\renewcommand{\arraystretch}{0}\setlength{\tabcolsep}{0pt}\begin{tabular}[b]{c}\rotatebox{90}{\makebox[0.0889\textheight][c]{\scriptsize NVS}}\\[0.0014\textheight]\rotatebox{90}{\makebox[0.0889\textheight][c]{\scriptsize Relighting}}\\[0.0014\textheight]\rotatebox{90}{\makebox[0.0889\textheight][c]{\scriptsize Albedo}}\\[0.0014\textheight]\rotatebox{90}{\makebox[0.0889\textheight][c]{\scriptsize Roughness}}\\[0.0014\textheight]\rotatebox{90}{\makebox[0.0889\textheight][c]{\scriptsize Metallic}}\end{tabular}}\hspace{3pt}\includegraphics[height=0.450\textheight]{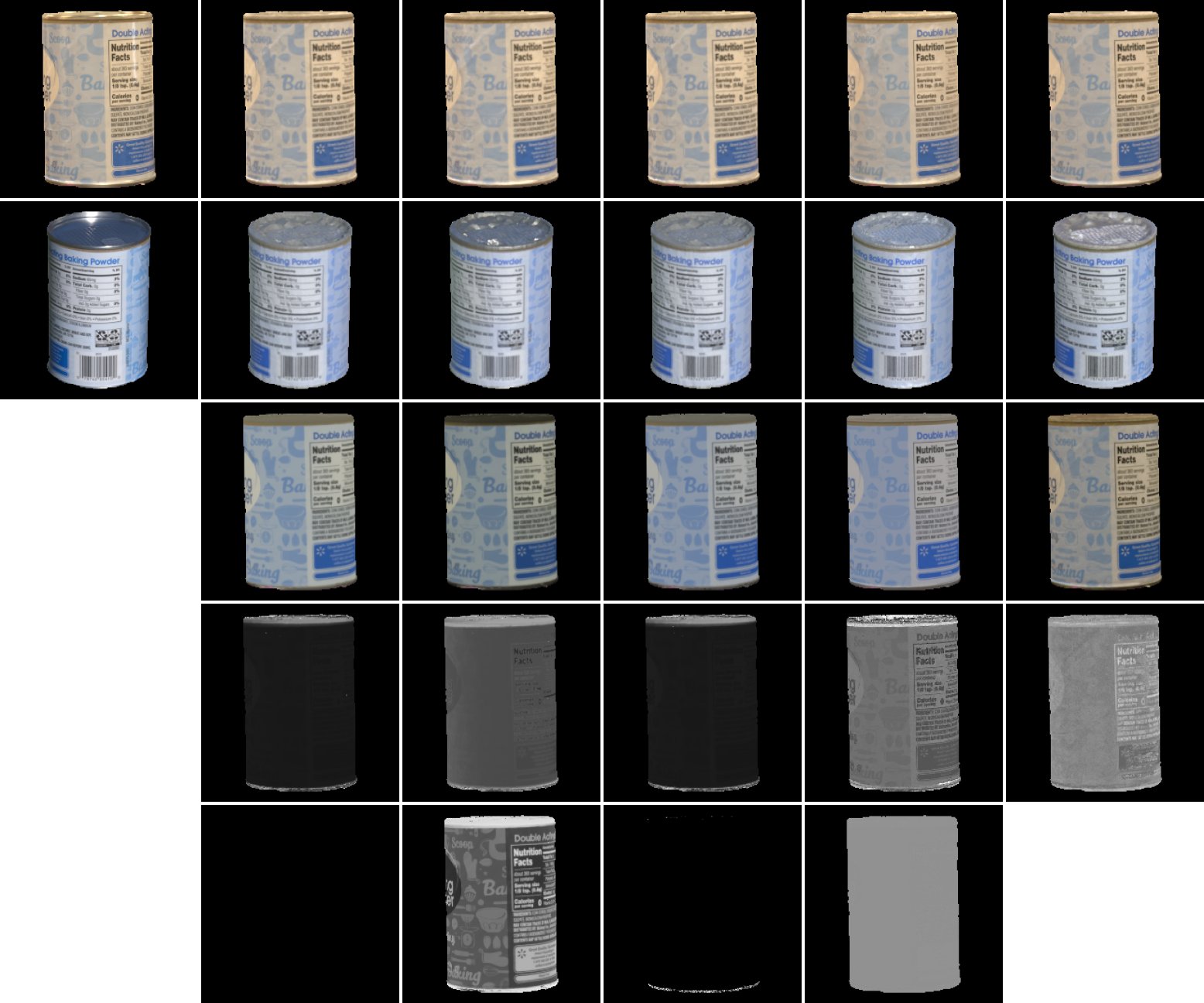}
\captionsetup{hypcap=false}
\captionof{figure}{Stanford-ORB baking\_scene001, camera 0063; relighting under the illumination of baking\_scene002.}
\label{fig:supp_gallery_orb_baking_scene001}
\captionsetup{hypcap=true}
\end{minipage}\par\vspace{8pt}
\noindent\begin{minipage}{\textwidth}\centering
\hspace{\dimexpr\baselineskip+3pt\relax}\makebox[0.0903\textheight][c]{\scriptsize GT}\makebox[0.0903\textheight][c]{\scriptsize Ours (fixed 0)}\makebox[0.0903\textheight][c]{\scriptsize Ours (learned)}\makebox[0.0903\textheight][c]{\scriptsize Ours (dual)}\makebox[0.0903\textheight][c]{\scriptsize DiffReg-PBIR}\makebox[0.0903\textheight][c]{\scriptsize Neural-PBIR}\\[1pt]
{\renewcommand{\arraystretch}{0}\setlength{\tabcolsep}{0pt}\begin{tabular}[b]{c}\rotatebox{90}{\makebox[0.0889\textheight][c]{\scriptsize NVS}}\\[0.0014\textheight]\rotatebox{90}{\makebox[0.0889\textheight][c]{\scriptsize Relighting}}\\[0.0014\textheight]\rotatebox{90}{\makebox[0.0889\textheight][c]{\scriptsize Albedo}}\\[0.0014\textheight]\rotatebox{90}{\makebox[0.0889\textheight][c]{\scriptsize Roughness}}\\[0.0014\textheight]\rotatebox{90}{\makebox[0.0889\textheight][c]{\scriptsize Metallic}}\end{tabular}}\hspace{3pt}\includegraphics[height=0.450\textheight]{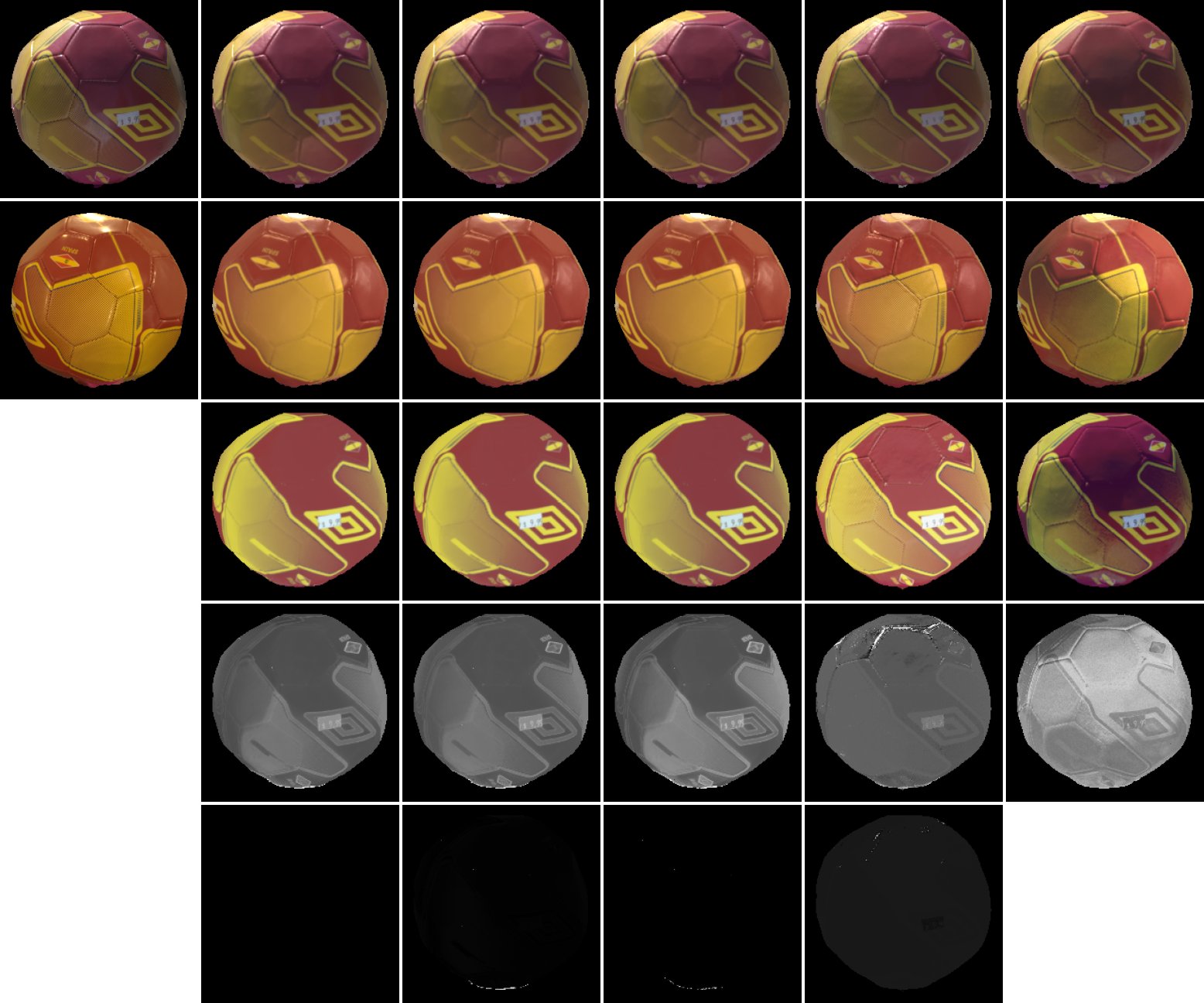}
\captionsetup{hypcap=false}
\captionof{figure}{Stanford-ORB ball\_scene002, camera 0060; relighting under the illumination of ball\_scene003.}
\label{fig:supp_gallery_orb_ball_scene002}
\captionsetup{hypcap=true}
\end{minipage}\par\vspace{8pt}
\noindent\begin{minipage}{\textwidth}\centering
\hspace{\dimexpr\baselineskip+3pt\relax}\makebox[0.0903\textheight][c]{\scriptsize GT}\makebox[0.0903\textheight][c]{\scriptsize Ours (fixed 0)}\makebox[0.0903\textheight][c]{\scriptsize Ours (learned)}\makebox[0.0903\textheight][c]{\scriptsize Ours (dual)}\makebox[0.0903\textheight][c]{\scriptsize DiffReg-PBIR}\makebox[0.0903\textheight][c]{\scriptsize Neural-PBIR}\\[1pt]
{\renewcommand{\arraystretch}{0}\setlength{\tabcolsep}{0pt}\begin{tabular}[b]{c}\rotatebox{90}{\makebox[0.0889\textheight][c]{\scriptsize NVS}}\\[0.0014\textheight]\rotatebox{90}{\makebox[0.0889\textheight][c]{\scriptsize Relighting}}\\[0.0014\textheight]\rotatebox{90}{\makebox[0.0889\textheight][c]{\scriptsize Albedo}}\\[0.0014\textheight]\rotatebox{90}{\makebox[0.0889\textheight][c]{\scriptsize Roughness}}\\[0.0014\textheight]\rotatebox{90}{\makebox[0.0889\textheight][c]{\scriptsize Metallic}}\end{tabular}}\hspace{3pt}\includegraphics[height=0.450\textheight]{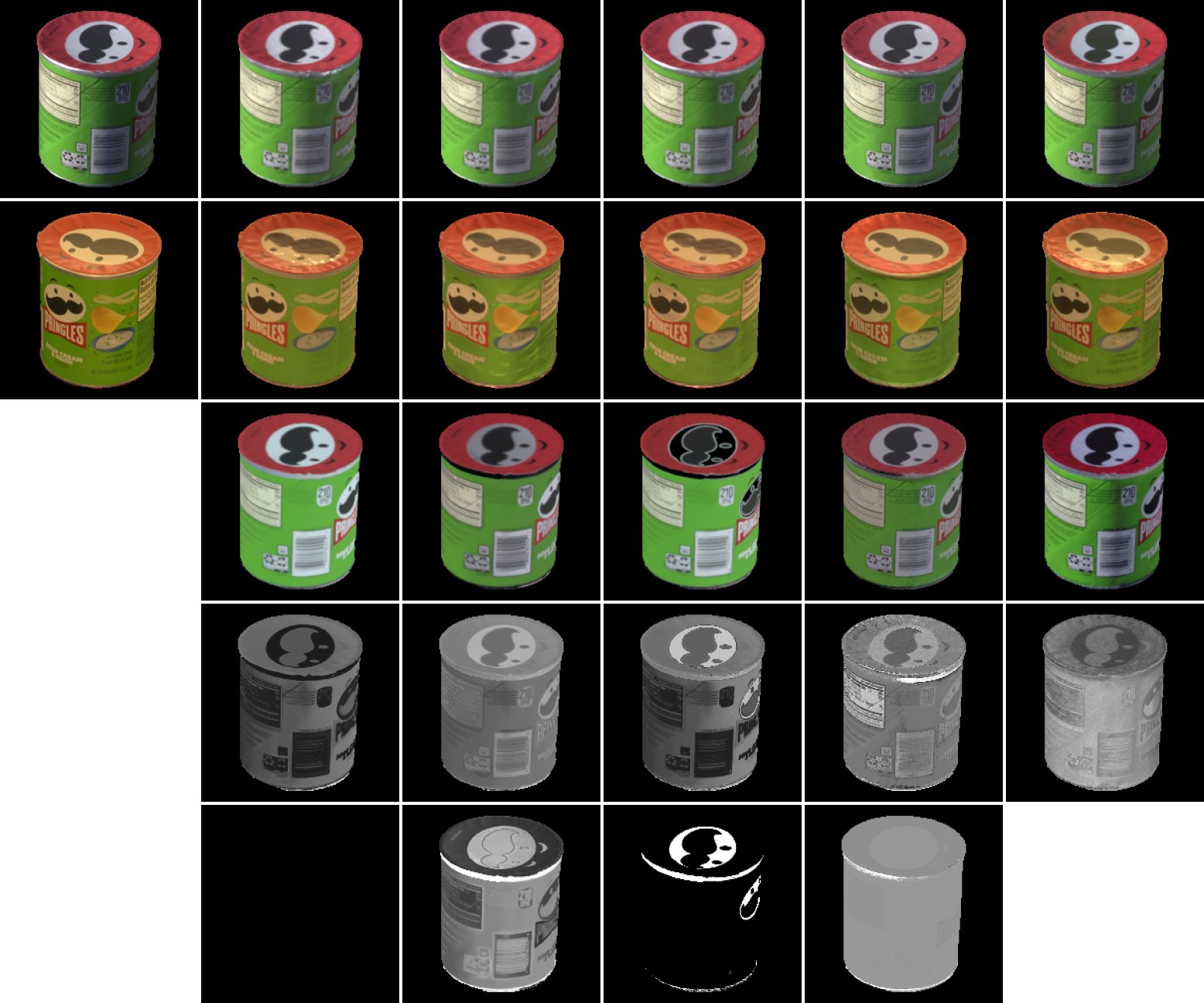}
\captionsetup{hypcap=false}
\captionof{figure}{Stanford-ORB chips\_scene002, camera 0062; relighting under the illumination of chips\_scene003.}
\label{fig:supp_gallery_orb_chips_scene002}
\captionsetup{hypcap=true}
\end{minipage}\par\vspace{8pt}
\noindent\begin{minipage}{\textwidth}\centering
\hspace{\dimexpr\baselineskip+3pt\relax}\makebox[0.0903\textheight][c]{\scriptsize GT}\makebox[0.0903\textheight][c]{\scriptsize Ours (fixed 0)}\makebox[0.0903\textheight][c]{\scriptsize Ours (learned)}\makebox[0.0903\textheight][c]{\scriptsize Ours (dual)}\makebox[0.0903\textheight][c]{\scriptsize DiffReg-PBIR}\makebox[0.0903\textheight][c]{\scriptsize Neural-PBIR}\\[1pt]
{\renewcommand{\arraystretch}{0}\setlength{\tabcolsep}{0pt}\begin{tabular}[b]{c}\rotatebox{90}{\makebox[0.0889\textheight][c]{\scriptsize NVS}}\\[0.0014\textheight]\rotatebox{90}{\makebox[0.0889\textheight][c]{\scriptsize Relighting}}\\[0.0014\textheight]\rotatebox{90}{\makebox[0.0889\textheight][c]{\scriptsize Albedo}}\\[0.0014\textheight]\rotatebox{90}{\makebox[0.0889\textheight][c]{\scriptsize Roughness}}\\[0.0014\textheight]\rotatebox{90}{\makebox[0.0889\textheight][c]{\scriptsize Metallic}}\end{tabular}}\hspace{3pt}\includegraphics[height=0.450\textheight]{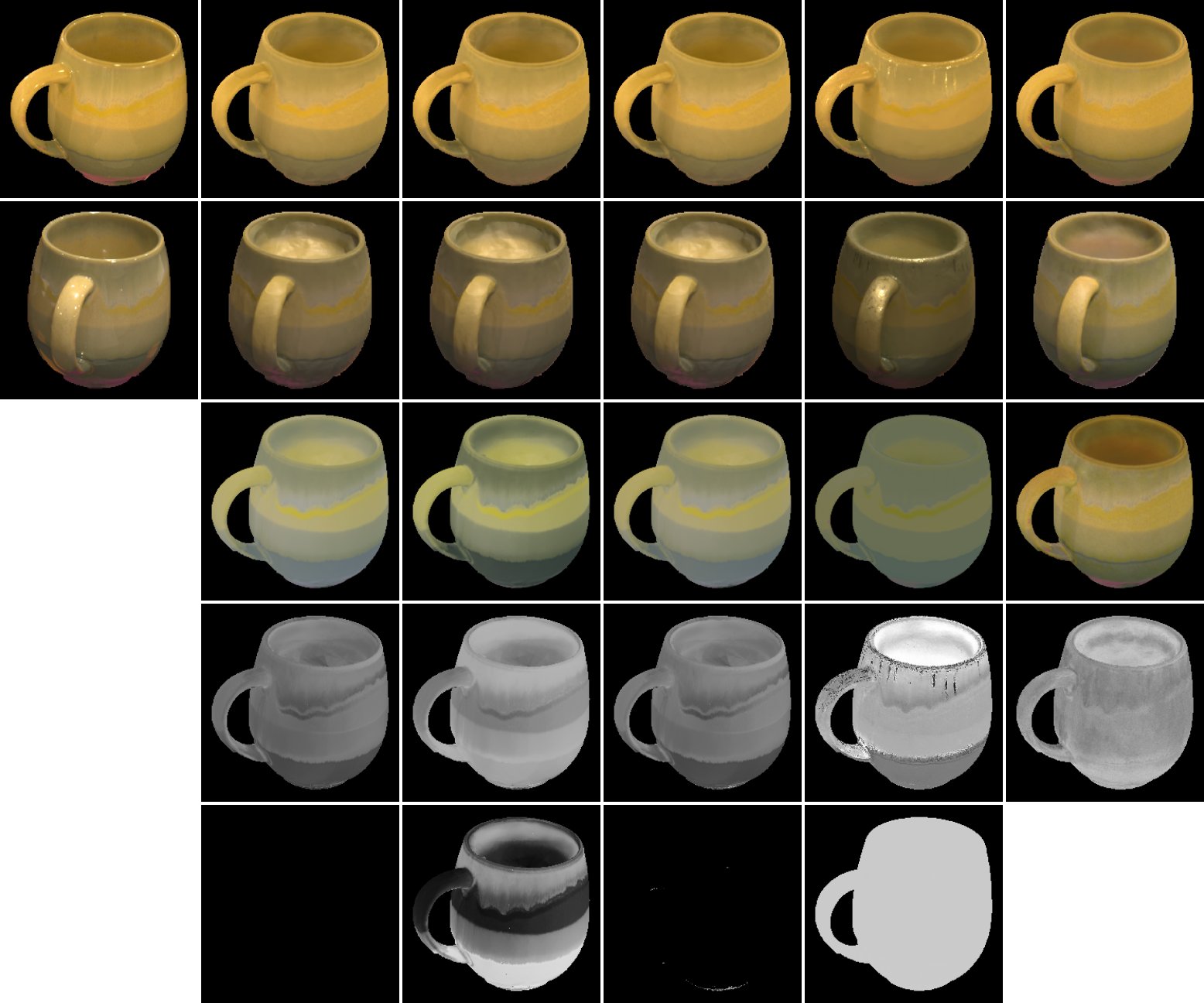}
\captionsetup{hypcap=false}
\captionof{figure}{Stanford-ORB cup\_scene003, camera 0060; relighting under the illumination of cup\_scene006.}
\label{fig:supp_gallery_orb_cup_scene003}
\captionsetup{hypcap=true}
\end{minipage}\par\vspace{8pt}
\noindent\begin{minipage}{\textwidth}\centering
\hspace{\dimexpr\baselineskip+3pt\relax}\makebox[0.0903\textheight][c]{\scriptsize GT}\makebox[0.0903\textheight][c]{\scriptsize Ours (fixed 0)}\makebox[0.0903\textheight][c]{\scriptsize Ours (learned)}\makebox[0.0903\textheight][c]{\scriptsize Ours (dual)}\makebox[0.0903\textheight][c]{\scriptsize DiffReg-PBIR}\makebox[0.0903\textheight][c]{\scriptsize Neural-PBIR}\\[1pt]
{\renewcommand{\arraystretch}{0}\setlength{\tabcolsep}{0pt}\begin{tabular}[b]{c}\rotatebox{90}{\makebox[0.0889\textheight][c]{\scriptsize NVS}}\\[0.0014\textheight]\rotatebox{90}{\makebox[0.0889\textheight][c]{\scriptsize Relighting}}\\[0.0014\textheight]\rotatebox{90}{\makebox[0.0889\textheight][c]{\scriptsize Albedo}}\\[0.0014\textheight]\rotatebox{90}{\makebox[0.0889\textheight][c]{\scriptsize Roughness}}\\[0.0014\textheight]\rotatebox{90}{\makebox[0.0889\textheight][c]{\scriptsize Metallic}}\end{tabular}}\hspace{3pt}\includegraphics[height=0.450\textheight]{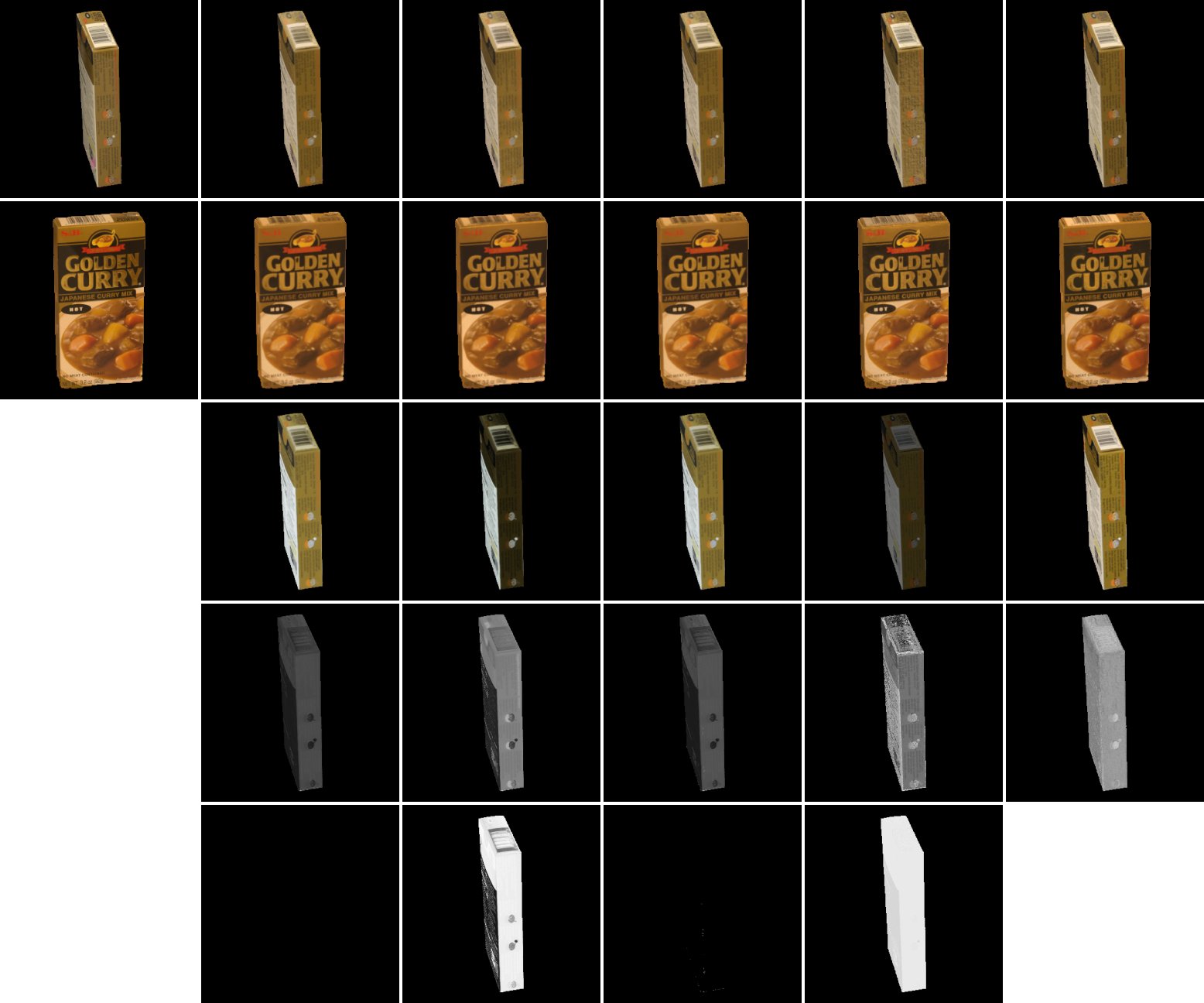}
\captionsetup{hypcap=false}
\captionof{figure}{Stanford-ORB curry\_scene001, camera 0062; relighting under the illumination of curry\_scene005.}
\label{fig:supp_gallery_orb_curry_scene001}
\captionsetup{hypcap=true}
\end{minipage}\par\vspace{8pt}
\noindent\begin{minipage}{\textwidth}\centering
\hspace{\dimexpr\baselineskip+3pt\relax}\makebox[0.0903\textheight][c]{\scriptsize GT}\makebox[0.0903\textheight][c]{\scriptsize Ours (fixed 0)}\makebox[0.0903\textheight][c]{\scriptsize Ours (learned)}\makebox[0.0903\textheight][c]{\scriptsize Ours (dual)}\makebox[0.0903\textheight][c]{\scriptsize DiffReg-PBIR}\makebox[0.0903\textheight][c]{\scriptsize Neural-PBIR}\\[1pt]
{\renewcommand{\arraystretch}{0}\setlength{\tabcolsep}{0pt}\begin{tabular}[b]{c}\rotatebox{90}{\makebox[0.0889\textheight][c]{\scriptsize NVS}}\\[0.0014\textheight]\rotatebox{90}{\makebox[0.0889\textheight][c]{\scriptsize Relighting}}\\[0.0014\textheight]\rotatebox{90}{\makebox[0.0889\textheight][c]{\scriptsize Albedo}}\\[0.0014\textheight]\rotatebox{90}{\makebox[0.0889\textheight][c]{\scriptsize Roughness}}\\[0.0014\textheight]\rotatebox{90}{\makebox[0.0889\textheight][c]{\scriptsize Metallic}}\end{tabular}}\hspace{3pt}\includegraphics[height=0.450\textheight]{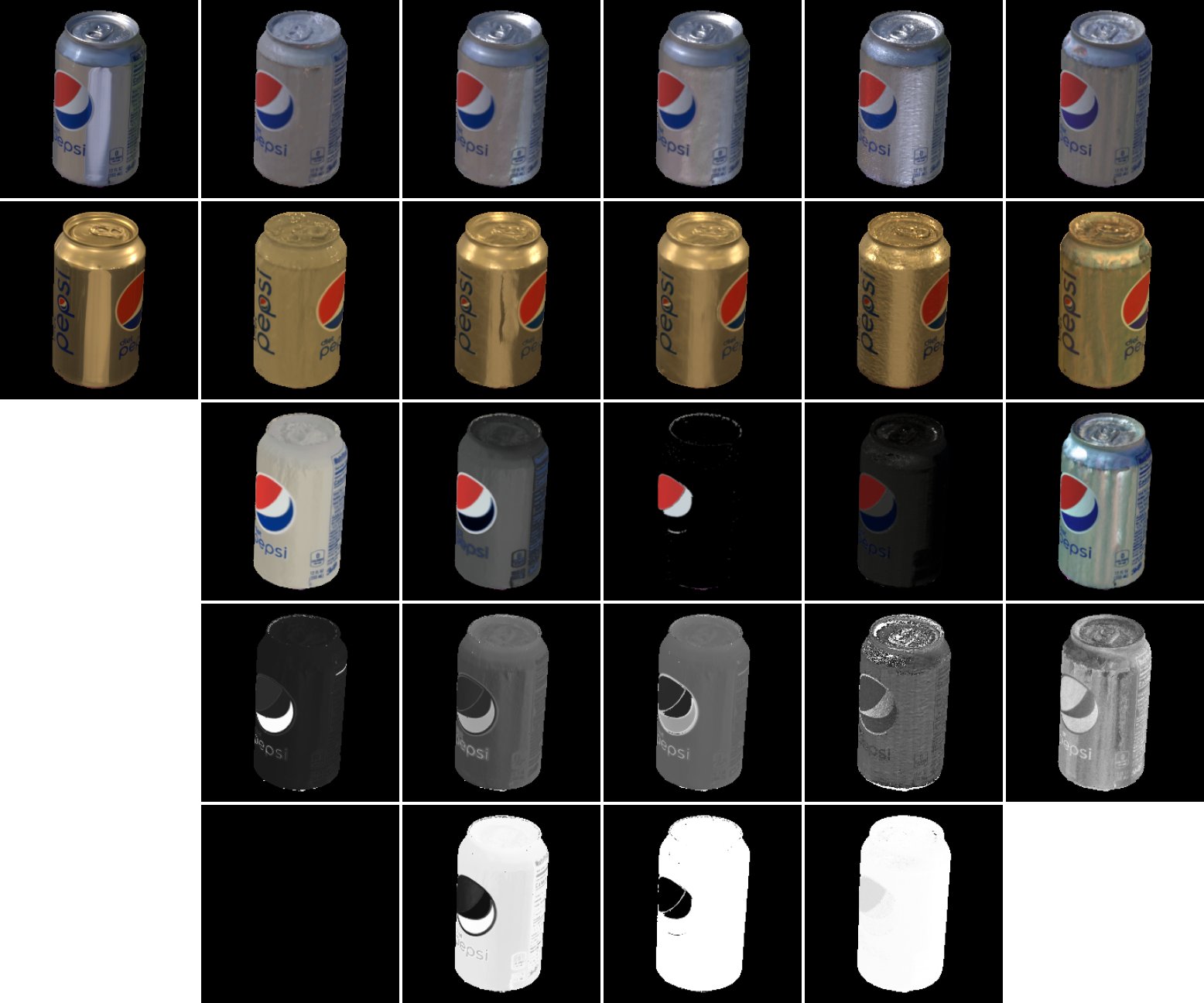}
\captionsetup{hypcap=false}
\captionof{figure}{Stanford-ORB pepsi\_scene002, camera 0061; relighting under the illumination of pepsi\_scene003.}
\label{fig:supp_gallery_orb_pepsi_scene002}
\captionsetup{hypcap=true}
\end{minipage}\par\vspace{8pt}
\noindent\begin{minipage}{\textwidth}\centering
\hspace{\dimexpr\baselineskip+3pt\relax}\makebox[0.0903\textheight][c]{\scriptsize GT}\makebox[0.0903\textheight][c]{\scriptsize Ours (fixed 0)}\makebox[0.0903\textheight][c]{\scriptsize Ours (learned)}\makebox[0.0903\textheight][c]{\scriptsize Ours (dual)}\makebox[0.0903\textheight][c]{\scriptsize DiffReg-PBIR}\makebox[0.0903\textheight][c]{\scriptsize Neural-PBIR}\\[1pt]
{\renewcommand{\arraystretch}{0}\setlength{\tabcolsep}{0pt}\begin{tabular}[b]{c}\rotatebox{90}{\makebox[0.0889\textheight][c]{\scriptsize NVS}}\\[0.0014\textheight]\rotatebox{90}{\makebox[0.0889\textheight][c]{\scriptsize Relighting}}\\[0.0014\textheight]\rotatebox{90}{\makebox[0.0889\textheight][c]{\scriptsize Albedo}}\\[0.0014\textheight]\rotatebox{90}{\makebox[0.0889\textheight][c]{\scriptsize Roughness}}\\[0.0014\textheight]\rotatebox{90}{\makebox[0.0889\textheight][c]{\scriptsize Metallic}}\end{tabular}}\hspace{3pt}\includegraphics[height=0.450\textheight]{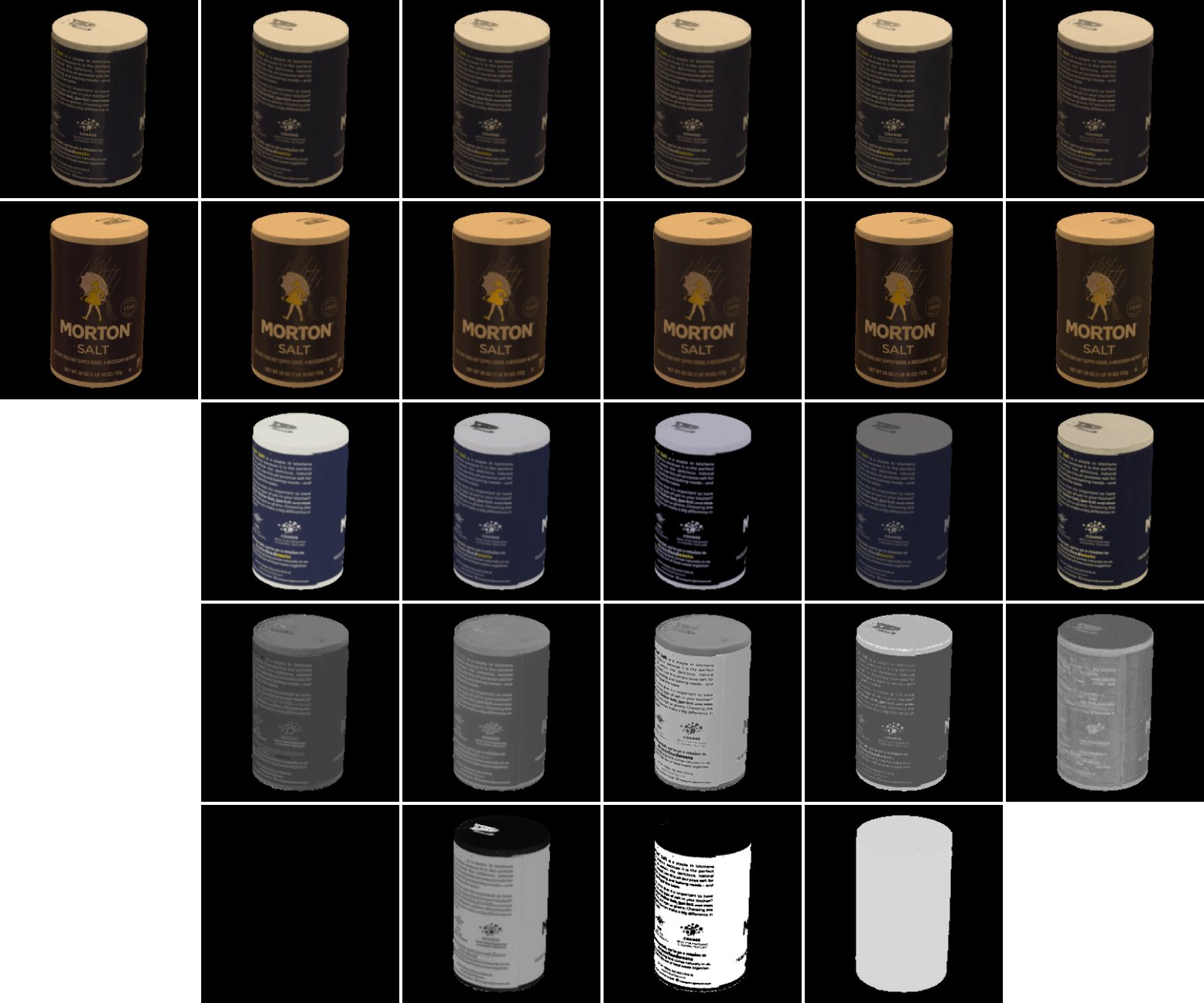}
\captionsetup{hypcap=false}
\captionof{figure}{Stanford-ORB salt\_scene004, camera 0066; relighting under the illumination of salt\_scene005.}
\label{fig:supp_gallery_orb_salt_scene004}
\captionsetup{hypcap=true}
\end{minipage}\par\vspace{8pt}

\end{document}